\documentclass{article}
\usepackage{ijcai22}

\usepackage{times}
\usepackage{soul}
\usepackage{url}
\usepackage[hidelinks]{hyperref}
\usepackage[utf8]{inputenc}
\usepackage[small]{caption}
\usepackage{graphicx}
\usepackage{amsmath}
\usepackage{amsthm}
\usepackage{booktabs}
\usepackage{algorithm}
\usepackage{color}
\usepackage{multirow}
\usepackage[noend]{algpseudocode}
\usepackage{amssymb}
\usepackage{enumerate}
\usepackage{subfigure}
\usepackage{pifont}
\usepackage[switch]{lineno}
\usepackage{lipsum}

\newtheorem{lemma}{Lemma}
\newtheorem{definition}{Definition}
\newtheorem{assumption}{Assumption}
\newtheorem{theorem}{Theorem}
\makeatletter
\def\thanks#1{\protected@xdef\@thanks{\@thanks
        \protect\footnotetext{#1}}}
\makeatother

\title{Meta-Tsallis-Entropy Minimization: A New Self-Training Approach\\
 for Domain Adaptation on Text Classification}

\author{
Menglong Lu$^{1}$\and
Zhen Huang$^{1}$\and
Zhiliang Tian$^{1}$\and
Yunxiang Zhao$^{2}$\and \\
Xuanyu Fei$^{3}$ \And
Dongsheng Li$^1$
\affiliations
$^1$National Key Laboratory of Parallel and Distributed Computing, National University of Defense Technology, China\\
$^2$Beijing Institute of Biotechnology, China \\
$^3$Suiren Information,China
\emails
\{lumenglong, huangzhen, tianzhiliang, dsli\}@nudt.edu.cn,
\{zhaoyx1993, xuanyufelix\}@163.com
}
\begin{document}
\maketitle

\begin{abstract}
Text classification is a fundamental task for natural language processing, and adapting text classification models across domains has broad applications. 
Self-training generates pseudo-examples from the model's predictions and iteratively train on the pseudo-examples, i.e., mininizes the loss on the source domain and the Gibbs entropy on the target domain. However, Gibbs entropy is sensitive to prediction errors, and thus, self-training tends to fail when the domain shift is large. In this paper, we propose Meta-Tsallis Entropy minimization (MTEM), which applies meta-learning algorithm to optimize the instance adaptive Tsallis entropy on the target domain. To reduce the computation cost of MTEM, we propose an approximation technique to approximate the Second-order derivation involved in the meta-learning. To efficiently generate pseudo labels, we propose an annealing sampling mechanism for exploring the model's prediction probability. Theoretically, we prove the convergence of the meta-learning algorithm in MTEM and analyze the effectiveness of MTEM in achieving domain adaptation. Experimentally, MTEM improves the adaptation performance of BERT with an average of 4 percent on the benchmark dataset.
\end{abstract}

\section{Introduction}
Text classification plays a crucial role in language understanding and anomaly detection for social media text. With the recent advance of deep learningf~\cite{kipf2016semi,devlin2019bert}, text classification has experienced remarkable progress. Despite the success, existing text classification approaches are vulnerable to domain shift. When transferred to a new domain, a well-performed model undergoes severe performance deterioration. To address such deterioration, domain adaptation, which aims to adapt a model trained on one domain to a new domain, has attracted much attention~\cite{du2020adversarial,lu2022sifter}.

A direct way to achieve domain adaptation is to build a training set that approximates the distribution of the target domain. For this purpose, self-training~\cite{zou2019confidence,liu2021cycle} uses the unlabeled data from the target domain to bootstrap the model. 
In specific, self-training first uses the model's prediction to generate pseudo-labels and then uses the pseudo-labeled data to re-train the model. 
In this process, self-training forces the model to increase its confidence in the confident class, which is a Gibbs entropy minimization process in essence~\cite{lee2013pseudo}. 

\begin{figure}[t]
  \centering
   \includegraphics[width=3.0in]{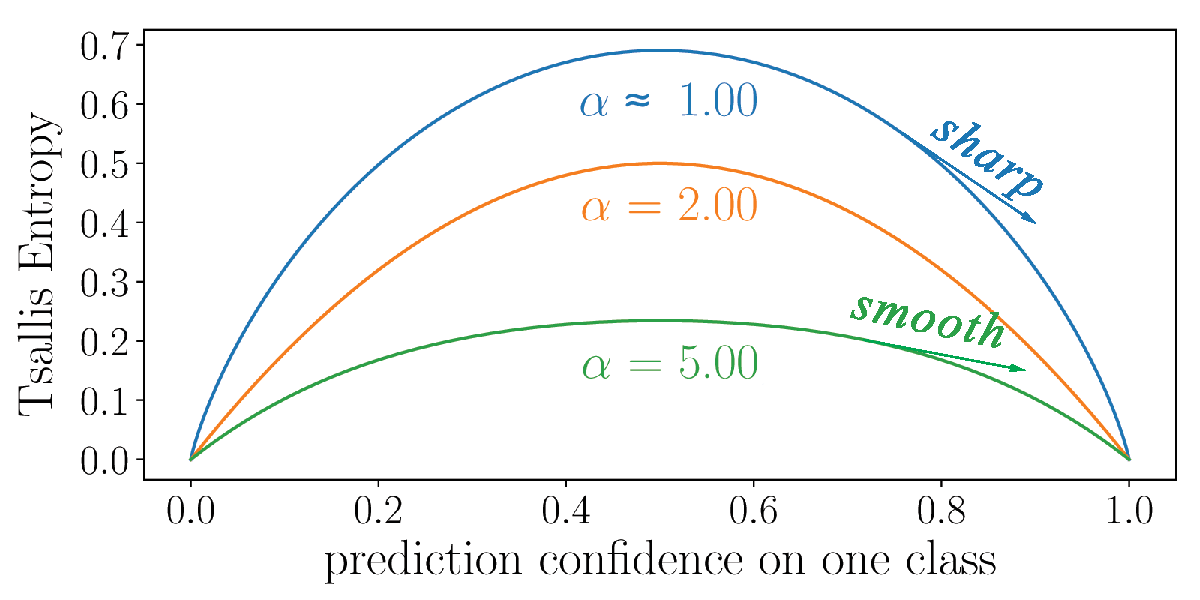}
  \caption{Tsallis entropy curve with respect to different entropy index (i.e., $\alpha$ below the curve).}. 
  \label{fig:tsallis}
\end{figure}

However, Gibbs entropy minimization is sensitive to prediction errors~\cite{mukherjee2020uncertainty}. To handle the intractable label noise (i.e., prediction errors), data selection strategies are designed to select reliable pseudo labels~\cite{mcclosky2006reranking,reichart2007self,rotman2019deep}. Among them, many qualified achievements~\cite{roychowdhury2019automatic,shin2020two} are grounded on prior knowledge about the tasks (e.g., the temporal consistency on video~\cite{roychowdhury2019automatic}), and thus hard to be applied in text classification tasks. 
Since the Gibbs entropy minimization process in self-training is to minimize the model's uncertainty on the new domain, \cite{liu2021cycle} recently proposes to replace the Gibbs entropy with the Tsallis entropy, which is another effective metric for measuring uncertainty.

Tsallis entropy is a generalization of Gibbs entropy, referring to a set of entropy types controlled by the entropy index. Fig.~\ref{fig:tsallis} shows the change of Tsallis entropy with different entropy indexes for binary problems. When the entropy index is small (the resultant entropy curve is sharp), the entropy minimization process tends to increase one dimension to 1.0 sharply, thus only being suitable for the scenario where pseudo labels are reliable. Otherwise, Tsallis entropy with a larger entropy index (smoother curve) is more suitable for scenarios with large label noise, e.g., domain adaptation scenarios with a large domain shift. 
Researchers~\cite{liu2021cycle} tried to use the Tsallis entropy to improve self-training, but the proposed objective only involves a unified entropy index for all unlabeled data in the target domain.
As illustrated in~\cite{kumar2010self,kumar2020understanding}, different instances in the target domain have different degrees of shifts from the source domain. Thus, a unified entropy index cannot fully exploit the different pseudo instances in the target domain.

In this paper, we propose Meta-Tsallis-Entropy Minimization (MTEM) that uses an instance adaptive Tsallis entropy minimization process to minimize the model's prediction uncertainty on the target domain. Since the best entropy indexes changes along with the training, manually selecting an appropriate entropy index for each unlabeled data is intractable. Thus, we employ meta-learning to adaptively learn a suitable entropy index for each unlabeled data.
The meta-learning process iterates over the \textit{inner loop} on the target domain and \textit{outer loop} on the source domain. In this process, the parameters optimized on the target domain also achieves a low loss on the source domain, which forces the model to obtain task informations on the target domain. 
However, the proposed MTEM still faces two challenges.

Firstly, the meta-learning algorithm in MTEM involves a Second-order derivation (i.e., the gradient of the entropy index), which requires much computation cost, especially when the model is large. Hence, it is hard to apply MTEM for prevailing big pre-trained language models. To this end, we propose to approximate the Second-order derivation via a Taylor expansion, which reduces the computation cost substantially.

Secondly, minimizing Tsallis entropy requires the guidance of pseudo labels (see $\S$~\ref{subsec:tsa_define} and $\S$~\ref{subsec:stda}). Previous self-training approaches generate pseudo labels by selecting the prediction with the largest probability (i.e., greedy selection), which tends to collapse when the model's prediction is unreliable~\cite{zou2019confidence}. To this end, we propose to sample pseudo labels from the model's predicted distribution instead of a greedy selection. Further, we propose an annealing sampling mechanism to improve the sampling efficiency.

To summarize, our contributions are in three folds\footnote{{As the rest paper involves many mathematic symbols, we provide a symbol list (Tab.~7 in Appendix~A) for reading convenience.}}:

\noindent (i) We propose Meta-Tsallis-Entropy Minimization (MTEM) for domain adaptation on text classification. MTEM involves an approximation technique to accelerate the computation, and an annealing sampling mechanism to improve the sampling efficiency.

\noindent (ii) We provide theoretical analysis for the MTEM, including its effectiveness in achieving domain adaptation and the convergence of the involved meta-learning process.

\noindent (iii) Experiments on two benchmark datasets demonstrate the effectiveness of the MTEM. Specifically, MTEM improves BERT on cross-domain sentiment classification tasks with an average of 4 percent, and improves BiGCN on cross-domain rumor detection task with an average of 21 percent.
\section{Preliminary}
\subsection{Domain Adaptation on Text Classification}
\label{subsec:Problem_Setting}
Text classification is a task that aims to map a text to a specific label space. On a correct classification case, the process is expressed as $y_i = \arg\max\limits_{k} f_{[k]}(x_i;\mathbf{\theta})$, where $x_i \in \mathcal{X}$ is an input text, $y_i  \in \lbrace 0, 1 \rbrace^{K}$ is the corresponding one-hot label with $K$ classes, and $f$ is a model with parameters $\theta$, $f(x_i;\mathbf{\theta})$ is the prediction probability. Domain adaptation is to adapt a text classification model trained on the source domain (denoted as $\mathbb{D}_{S}$) to the target domain (denoted as $\mathbb{D}_{T}$). On the source domain, we have a set of labeled instances, i.e., $D_S = \lbrace (x_i, y_i) \rbrace_{i=1}^{N}$, which satisfies that $D_{S} \subseteq \mathbb{D}_{S}$. On the target domain, unlabeled text in the target domain is available, which we denote as $D_T^{u} = \lbrace (x_m) \rbrace_{m=1}^{U}$. 

\subsection{Tsallis Entropy}
\label{subsec:tsa_define}
In information theory, Tsallis entropy refers to a set of entropy types, where the entropy index is used to identify a specific entropy. Formally, Tsallis entropy with $\alpha$ denoting the entropy index is written as Eq.~\eqref{eq:tsallis}, 

\begin{small}
\begin{eqnarray}
e_\mathbf{\alpha}(p_i) &=& \frac{1}{\mathbf{\alpha} - 1}(1 - \sum_{j=1}^K p^{\mathbf{\alpha}}_{i[j]}) \label{eq:tsallis}
\end{eqnarray}
\end{small}

\noindent where $p_i$ is the prediction probability. When $\alpha >1$, $e_\mathbf{\alpha}$ is a concave function~\cite{plastino1999tsallis}. When $\alpha \rightarrow 1$, $e_\mathbf{\alpha}$ recovers the Gibbs entropy, as shown in Eq.~\eqref{eq:gibbs_entropy_main}~\footnote{{The second equation is obtained by L'Hôpital's rule.}}:

\begin{small}
\begin{eqnarray}
\quad e_{\mathbf{\alpha}\rightarrow 1}(p_i) = \frac{\lim_{\mathbf{\alpha} \rightarrow 1}1 - \sum_{j=1}^K p^{\mathbf{\alpha}}_{i[j]}}{\lim_{\mathbf{\alpha} \rightarrow 1}\mathbf{\alpha} - 1} = \sum_{j=1}^K - p_{i[j]}log(p_{i[j]}) \label{eq:gibbs_entropy_main} 
\end{eqnarray}
\end{small}

More intuitively, Fig.~\ref{fig:tsallis} exhibits the impact of the entropy index on the curves of the Tsallis entropy type. Specifically, a larger entropy index makes the curve more smooth, while a smaller entropy index exerts a more sharp curve.

Extending from the unsupervised Tsallis entropy, the corresponding \textit{Tsallis loss} $\ell_{\alpha}(p_i, y_i)$ is expressed as Eq.~\eqref{eq:tsallis_loss}. When $\alpha \rightarrow 1$, the corresponding supervised loss is the widely used cross-entropy loss (see Appendix~B.1).

\begin{small}
\begin{eqnarray}
\ell_\mathbf{\alpha}(p_i, y_i) &=& \frac{1}{\mathbf{\alpha} - 1}(1 - \sum_{j=1}^K y_{i[j]}\cdot p^{\mathbf{\alpha}-1}_{i[j]}) \label{eq:tsallis_loss}
\end{eqnarray}
\end{small}



\subsection{Self-Training for Domain Adaptation}
\label{subsec:stda}
Self-training aims to achieve domain adaptation by optimizing the model's parameters with respect to the supervised loss on the source domain and the unsupervised loss (prediction uncertainty) on the target domain, as shown in Eq.~\eqref{eq:ST}.
\begin{small}
\begin{eqnarray}
\min\limits_{\mathbf{\theta}}\mathcal{L}_{ST}(\mathbf{\theta}|D_S, D_T^u) &=& \mathcal{L}_S(\mathbf{\theta}|D_S) + \lambda \cdot \mathcal{L}_T(\mathbf{\theta}|D_T^u) \label{eq:ST}
\end{eqnarray}
\end{small}

\noindent where $\mathcal{L}_S$ is a supervised loss, $\mathcal{L}_T$ is an unsupervised loss, and $\lambda$ is a coefficient to balance $\mathcal{L}_S$ and $\mathcal{L}_T$. Due to the simplicity, Gibbs entropy is widely used to measure the prediction uncertainty on the target domain~\cite{zou2019confidence,zou2018unsupervised}, which is expressed as below: 

\begin{small}
\begin{eqnarray}
\mathcal{L}_T(\mathbf{\theta}|D_T^u) = \frac{1}{|D_T^u|} \sum\limits_{x_i \in D_T^u} -f(x_i;\mathbf{\theta}) \cdot log(f(x_i;\mathbf{\theta})) \label{eq:ST_DU}
\end{eqnarray}
\end{small}
However, as $\mathcal{L}_T(\mathbf{\theta}|D_T^u)$ in Eq.~\eqref{eq:ST_DU} is a concave function, minimizing $\mathcal{L}_T(\mathbf{\theta}|D_T^u)$ is hard to converge because the gradients on the minimal are larger than 0~\cite{benson1995concave}. For this purpose, self-training uses pseudo labels to guide the entropy minimization process, i.e., replacing Eq.~\eqref{eq:ST_DU} with Eq~\eqref{eq:ST_DU_new} where $\tilde{y}_i = \arg\max\limits_{k} f_{[k]}(x_i;\mathbf{\theta})$ is the pseudo label.

\begin{small}
\begin{eqnarray}
\mathcal{L}_T(\mathbf{\theta}|D_T^u) = \frac{1}{|D_T^u|} \sum\limits_{x_k \in D_T^u} -\tilde{y}_i^T \cdot log(f(x_i;\mathbf{\theta})) \label{eq:ST_DU_new}
\end{eqnarray}
\end{small}


\begin{figure*}[t]
    \includegraphics[width=6.3in]{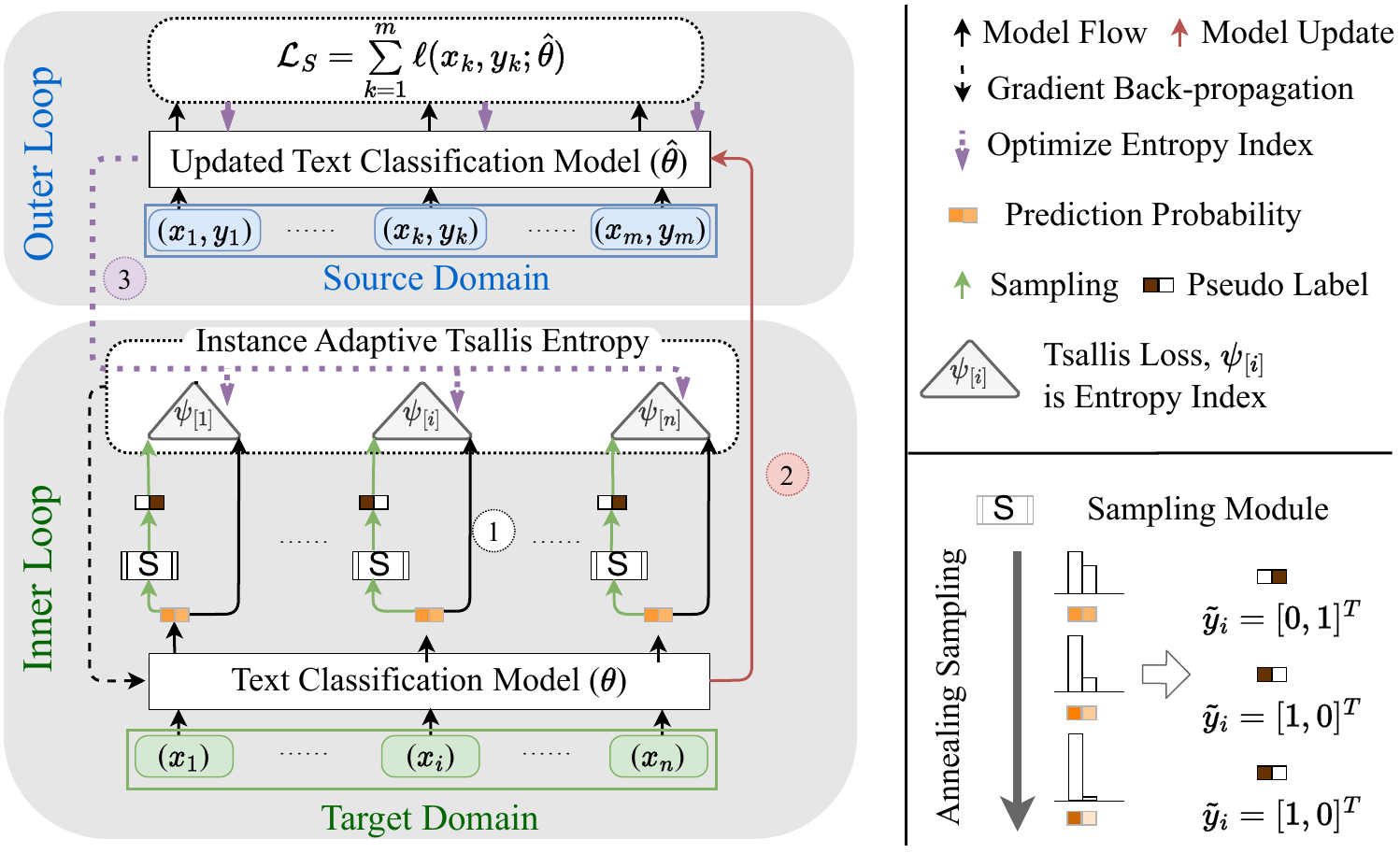}
    \label{subfig:MetaTsallis}
  \centering
  \caption{Meta-Tsallis-Entropy Minimization for domain adaptation on text classification:~\ding{172} generate pseudo labels with annealing sampling module, and then update the model with the instance adaptive Tsallis entropy; \ding{173} validate the model on the source domain; \ding{174} update the entropy indexes with respect to the validation performance.
  }
  \label{fig:overview}
\end{figure*}
\section{Meta-Tsallis-Entropy Minimization}
\label{subsec:BiHPO}
MTEM inherits the basic framework of self-training, i.e., (i) simultaneously minimizing the supervised loss on the source domain and the unsupervised loss (prediction uncertainty) on the target domain; (ii) generating the pseudo labels to guide the entropy minimization process. The improvements of the MTEM are in three folds. Firstly, we propose an instance adaptive Tsallis entropy to measure the prediction uncertainty ($\S$~\ref{subsec:Instance_Adaptive_Tsallis_Entropy}). Secondly, we propose to use a meta-learning algorithm to minimize the joint loss ($\S$~\ref{subsec:meta-learning}), which involves an approximation technique to reduce computation cost ($\S$~\ref{subsec:Subsec_Taylor}). Thirdly, we propose to generate pseudo labels with an annealing sampling ($\S$~\ref{subsec:annealing_sampling}). Fig.~\ref{fig:overview} exhibits the overview of the MTEM, and Algorithm~\ref{algo:MTEM} presents the core process.

\begin{algorithm}[tb]
\small
\caption{Meta-Tsallis-Entropy Minimization}
\label{algo:MTEM}
\begin{algorithmic}[1]
\Require labeled source dataset $D_S$, unlabeled target dataset $D_T^{u}$, initial entropy-index on $D_T^{u}$, i.e., $\psi_1 = [\psi_{1[i]}|_{i=1}^{|D_T^u|}]_{t}$
\For{t = 1 $\to T_\textnormal{max}$}
\State Sampling training batch $\mathcal{B}=\lbrace x_j \rbrace$ from $D^u_{T}$
\State Sampling validation batch $\mathcal{V}$ from $D_{S}$
\State Set $\kappa_{t}$ with Eq.\eqref{eq:sigmoid_annealing}
\State Sampling $\tilde{y}_{j} \sim p(\bullet|\mathbf{\theta}, x_j, \kappa_t)$ for each instance in $\mathcal{B}$
\State $\hat{\mathbf{\theta}}(\psi_{t}) = \mathbf{\theta}_{t} - \eta_t\cdot\frac{\partial \mathcal{L}_T(\mathbf{\theta}, \mathbf{\psi}_{t}| \mathcal{B})}{\partial \mathbf{\theta}}|_{\mathbf{\theta} = \mathbf{\theta}_t} $ \Comment{\textcolor[RGB]{0, 102, 0}{Inner-Loop}}
\State $\mathbf{\psi}_{t+1} =\mathbf{\psi}_{t} - \beta_t\cdot \frac{\partial \mathcal{L}_S(\hat{\mathbf{\theta}}(\psi_{t})| \mathcal{V})}{\partial \mathbf{\psi}}|_{\psi =\psi_t}$ \Comment{\textcolor[RGB]{0, 102, 204}{Outer-Loop}}
\State $\mathbf{\theta}_{t+1} = \mathbf{\theta}_{t} - \eta_t\cdot\frac{\partial \mathcal{L}_T(\mathbf{\theta}, \mathbf{\psi}^{t+1}| \mathcal{B})}{\partial \mathbf{\theta}}|_{\mathbf{\theta} = \mathbf{\theta}_t} $ 
\EndFor
\end{algorithmic}
\end{algorithm}
\subsection{Instance Adaptive Tsallis Entropy}
\label{subsec:Instance_Adaptive_Tsallis_Entropy}
The instance adaptive Tsallis Entropy, i.e., the unsupervised loss on the target domain, is as below:

\begin{small}
\begin{eqnarray}
\mathcal{L}_T(\theta, \psi|D_T^u) &=& \frac{1}{|D_T^u|} \sum\limits_{x_k \in D_T^u} e_{\psi_{[k]}}(f(x_k;\mathbf{\theta})) \label{eq:object_old} 
\end{eqnarray}
\end{small}

\noindent where $\psi_{[k]}$ indicates the entropy index for unlabeled data $x_{k}$, $e_{\psi_{[k]}}$ is the resultant Tsallis entropy.

Such an instance Tsallis entropy minimization is more effective in exploiting the model's prediction. In general, the prediction correctness is different on different instances. Thus, entropy index should be different on different instances too. For the instances with wrong prediction, we can increase the entropy index to make Tsallis entropy more smooth, then the model is updated more cauciously. Otherwise, for instances with correct prediction, we can set a small entropy index to update the model more aggressively.

However, as we are not aware of the label, setting the appropriate entropy indexes for each unlabeled data is intractable. Furthermore, as prediction errors can be corrected during the model training, the best entropy indexes change along with the updates of the model. To handle the above issues, we propose to use meta-learning to determine the entropy indexes automatically. 

\subsection{Meta-Learning}
\label{subsec:meta-learning}
Meta-learning can help MTEM to find the appropriate entropy indexes due to the following reasons. Firstly, parameters optimized on a well-determined instance adaptive Tsallis entropy (entropy indexes are determined appropriately) should be more generalizable, which corresponds to the characteristics of meta-learning, i.e., training a model that can be fast adapted to a new task~\cite{finn2017model}. Secondly, the meta-learning process updates the entropy indexes dynamically, thus maintaining the consistency between the model’s parameters and the entropy indexes along with the whole training process. In specific, the meta-learning algorithm in MTEM iterates over the \textit{Inner-Loop} on the target data and the \textit{Outer-Loop} on the source domain.

In the \textit{Inner-Loop}, we fix the entropy indexes to optimize the model's parameters with respect to the instance-adaptive Tsallis entropy on the target domain. In specific, we sample a batch of unlabeled data $\mathcal{B}$ from $D_T^u$ and update the model with respect to their instance adaptive Tsallis entropy ($\mathcal{L}_T(\theta, \psi|\mathcal{B})$), as shown in Eq.~\eqref{eq:virtual_update}:
\begin{small}
\begin{eqnarray}
\hat{\mathbf{\theta}}_{t+1}(\psi_t) &=& \mathbf{\theta}_t - \eta\cdot\frac{\partial \mathcal{L}_T(\mathbf{\theta}, \mathbf{\psi}_t|\mathcal{B})}{\partial \mathbf{\theta}}|_{\mathbf{\theta} = \mathbf{\theta}_t}, \label{eq:virtual_update}
\end{eqnarray}
\end{small}

\noindent However, as introduced in $\S$~\ref{subsec:tsa_define} and $\S$~\ref{subsec:stda}, $e_{\psi}$ in Eq.~\eqref{eq:object_old} is a concave function hard to be minimized. Following the way in self-training, we use the equation in Eq.~\eqref{eq:loss_new}~\footnote{{Deduction is provided in Appendix~B.2}} to transform a concave function to a convex function, 

\begin{small}
\begin{eqnarray}
e_\psi(f(x_i;\mathbf{\theta}_t))&=& \mathop{\mathbb{E}}\limits_{\tilde{y}_i \sim f(x_i;\mathbf{\theta}_t)}\ell_{\psi_{[i]}}(f(x_i;\mathbf{\theta}_t), \tilde{y}_i) \label{eq:loss_new}
\end{eqnarray}
\end{small}

\noindent where $\tilde{y}_i \in \lbrace 0, 1\rbrace^K$ is a one-hot pseudo label sampled from the model's prediction probability (i.e., $\tilde{y}_i \sim f(x_i;\mathbf{\theta}_t)$). Then, the objective in the Inner-Loop is as Eq.~\eqref{eq:object_new}:

\begin{small}
\begin{eqnarray}
\min\limits_{\mathbf{\theta}}\mathcal{L}_T(\mathbf{\theta}, \mathbf{\psi}| D_T^u) &=& \mathop{\mathbb{E}}\limits_{\tilde{y}_i \sim f(x_i;\mathbf{\theta})}\frac{1}{|D_T^u|}\sum\limits_{x_i \in D_T^u} \ell_{\psi_{[i]}}(f(x_i;\mathbf{\theta}), \tilde{y_i}) \nonumber \\ \label{eq:object_new}
\end{eqnarray}
\end{small}

In the \textit{Outer-Loop}, we validate the model updated (i.e., $\hat{\mathbf{\theta}}_{t+1}(\psi_t)$ in Eq.~\eqref{eq:virtual_update}) with labeled data from the source domain. Since different entropy indexes $\psi$ leads to different $\hat{\mathbf{\theta}}_{t+1}(\psi_t)$, we adjust $\psi$ to find the better $\hat{\mathbf{\theta}}_{t+1}(\psi^t)$ that can be fast adapted to the validation set. For this purpose, we optimize the entropy indexes $\psi$ to minimize the validation loss. In each meta-validation step, we sample a valid batch of labeled data from the source domain, i.e., $\mathcal{V} \sim D_{S}$, and use the validation loss $\mathcal{L}_{S}(\hat{\mathbf{\theta}}_{t+1}(\psi^t)|\mathcal{V})$ to evaluate the model, then update entropy indexes $\psi$ with $\bigtriangledown_{\psi}\mathcal{L}_{S}(\hat{\mathbf{\theta}}_{t+1}(\psi^t)|\mathcal{V})$. With the updated entropy indexes $\psi_{t+1}$, we return to update the model's parameters, as shown in line 8 of Algorithm~\ref{algo:MTEM}.



\subsection{Taylor Approximation Technique}
\label{subsec:Subsec_Taylor}
The first challenge in the above meta-learning algorithm is the computation cost carried out in the $\bigtriangledown_{\psi}\mathcal{L}_{S}(\hat{\mathbf{\theta}}_{t+1}(\psi^t)|\mathcal{V})$. Formally, the computation of $\bigtriangledown_{\psi}\mathcal{L}_{S}(\hat{\mathbf{\theta}}_{t+1}(\psi^t)|\mathcal{V})$ is as:

\begin{small}
\begin{eqnarray}
\frac{\partial \mathcal{L}_S(\hat{\mathbf{\theta}}_{t+1}(\psi^t)|\mathcal{V})}{\partial \mathbf{\psi}} &=& \frac{\partial \mathcal{L}_S(\hat{\mathbf{\theta}}_{t+1}(\psi_t)|\mathcal{V})}{\partial \hat{\mathbf{\theta}}_{t+1}(\psi_t)}\cdot \frac{\partial \hat{\mathbf{\theta}}_{t+1}(\psi_t)}{\partial \mathbf{\psi}} \nonumber \\
&=& -\eta \bigtriangledown_{\hat{\theta}}\mathcal{L}_S(\hat{\mathbf{\theta}}_{t+1}(\psi_t)|\mathcal{V}) \nonumber \\
& & \quad \cdot \frac{\partial^2 \mathcal{L}_T(\mathbf{\theta}, \mathbf{\psi}_t| \mathcal{B})}{\partial \mathbf{\theta}\partial \mathbf{\psi}}  \label{eq:meta_update} 
\end{eqnarray}
\end{small}

\noindent where the second equation is obtained by substituting $\hat{\mathbf{\theta}}_{t+1}(\psi^t)$ with Eq.~\eqref{eq:virtual_update}. 
Since $\frac{\partial^2 \mathcal{L}_T(\mathbf{\theta}, \mathbf{\psi}^t| \mathcal{B})}{\partial \mathbf{\theta}\partial \mathbf{\psi}}$ in Eq.~\eqref{eq:meta_update} is a hessian matrix (Second-order derivation), the computation in Eq.~\eqref{eq:meta_update} is intractable. 
Although deep learning codebase, i.e. Pytorch and TensorFlow, provide tools for computing the Second-order derivation, the computation cost is quadratic to the model’s parameters, which is thus unacceptable for recent big pre-trained language models (e.g., BERT). 

Inspired by the research in~\cite{liu2018darts,chen2021wind}, we propose an approximation technique for computing Eq.~\eqref{eq:meta_update}.
In specific, we employ the Taylor Expansion to rewrite the term $\frac{\partial\mathcal{L}_S(\hat{\theta}(\mathbf{\psi}))}{\partial\hat{\theta}}\frac{\partial^2 \mathcal{L}_T(\mathbf{\psi})}{\partial\theta \partial\mathbf{\psi}}$ in Eq.~\eqref{eq:meta_update} with Eq.~\eqref{eq:taylor_exp}. 

\begin{small}
\begin{eqnarray}
&&\bigtriangledown_{\hat{\theta}}\mathcal{L}_S(\hat{\mathbf{\theta}}_{t+1}(\psi_t)|\mathcal{V})\cdot \frac{\partial^2 \mathcal{L}_T(\mathbf{\theta})}{\partial \mathbf{\theta}\partial \mathbf{\psi}} \nonumber \\
&&= \frac{\bigtriangledown_{\mathbf{\psi}}\mathcal{L}_T(\theta^+) - \bigtriangledown_{\mathbf{\psi}}\mathcal{L}_T(\theta^-)}{2*\epsilon} \label{eq:taylor_exp} 
\end{eqnarray}
\end{small}

\noindent where $\epsilon$ is a small scalar, $\mathbf{\theta}^{+}$ and $\mathbf{\theta}^{-}$ are defined as below: 

\begin{small}
\begin{eqnarray}
\theta^+ &=& \theta + \epsilon\cdot\bigtriangledown_{\hat{\theta}}\mathcal{L}_S(\hat{\mathbf{\theta}}_{t+1}(\psi_t)|\mathcal{V}), \nonumber \\
\theta^- &=& \theta - \epsilon\cdot\bigtriangledown_{\hat{\theta}}\mathcal{L}_S(\hat{\mathbf{\theta}}_{t+1}(\psi_t)|\mathcal{V}) 
\end{eqnarray}
\end{small}

\noindent where $\mathcal{L}_T(\mathbf{\theta})$ is the abbreviation of $\mathcal{L}_T(\mathbf{\theta}, \mathbf{\psi}_t| \mathcal{B})$.
As demonstrated in~\cite{liu2018darts}, Eq.~\eqref{eq:taylor_exp} would be accurate enough for approximation when $\epsilon$ is small.
However, computing $\bigtriangledown_{\psi}\mathcal{L}_{T}$ in Eq.~\eqref{eq:taylor_exp} still requires much computation cost as it involves a forward operation (i.e., $\mathcal{L}_{T}$) and a backward operation (i.e., $\bigtriangledown_{\psi}\mathcal{L}_{T}$). To this end, we derive the explicit form of  $\bigtriangledown_{\psi}\mathcal{L}_{T}$ as Eq.~\eqref{eq:psi_grad} (details are in Appendix~B.3).  

\begin{small}
\begin{eqnarray}
\bigtriangledown_{\psi_{[i]}}\mathcal{L}_T(\mathbf{\theta})
&=& \frac{1}{\psi_{[i]} - 1}\times[l_{1}(x_i, \tilde{y}_i) - l_{\psi_{[i]}}(x_i, \tilde{y}_i)]  \nonumber \\
& & \quad - l_{1}(x_i, \tilde{y}_i) \times l_{\psi_{[i]}}(x_i, \tilde{y}_i)\label{eq:psi_grad} 
\end{eqnarray}
\end{small}

$l_{1}(x_i, \tilde{y}_i)$ and $l_{\psi_{[i]}}$ in Eq.~\eqref{eq:psi_grad} can be computed without gradients, thus preventing the time-consuming back-propagation process. Therefore, computing $\bigtriangledown_{\psi}\mathcal{L}_{T}$ with the above explicit form can further reduce the computation cost. 

\subsection{Annealing Sampling}
\label{subsec:annealing_sampling}
In domain adaptation, the naive sampling mechanism in the Inner-Loop can suffer from the lowly efficient sampling problem. When the domain shift is large, the model usually performs worse in the target domain than in the source domain. As a result, the model's prediction confidence (i.e., the sampling probability) on the true class is small. Considering an extreme binary classification case, where the instance's ground truth label is $[0, 1]_{t}$ but the model's prediction is $[0.99, 0.01]_{t}$, the probability of sampling the ground truth label is 0.01. In this case, most of the training cost is wasted on the pseudo instances with error labels. 

To improve the sampling efficiency, we propose an annealing sampling mechanism. 
With a temperature parameter $\kappa$, we control the sharpness of the model's prediction probability (sampling probability) by
$p(\bullet;\mathbf{\theta}, x_i, \kappa) = softmax(\frac{score}{\kappa})$, 
 where $p$ is the sampling probability and $score$ is the model's original prediction score. 
In the earlier training phase, the model's prediction is not that reliable, so we set a high-temperature parameter $\kappa$ to smooth the model's prediction distribution. With this setting, different class labels are sampled with roughly equal probability, which guarantees the possibility of sampling the correct pseudo label. Along with the convergence of the training process, the model's prediction is more and more reliable, thus the temperature scheduler will decrease the model's temperature. We design a temperature scheduler as Eq.~\eqref{eq:sigmoid_annealing}:

\begin{small}
\begin{equation}
\label{eq:sigmoid_annealing}
\kappa_t = \kappa_{max} - (\kappa_{max} - \kappa_{min})\sigma(s - 2s\times\frac{t}{T_{\textnormal{max}}})
\end{equation}
\end{small}

where $\sigma$ denotes the \textit{sigmoid} function\footnote{{$\sigma(x) = \frac{1}{1+e^{x}}$, which approaches to 0 when $x < -5.0$ and saturates to 1.0 when $x>5.0$}}, $\kappa_{max}$ and $\kappa_{min}$ are the expected maximum temperature and minimum temperature, $s$ is a manual set positive scalar. $t$ is the index of the current training iteration, $T_\textnormal{max}$ is the maximum of the training iterations. Thus, $\frac{t}{T_{\textnormal{max}}}$ increases from 0.0 to 1.0, and the input $s - 2s\times\frac{t}{T_{\textnormal{max}}}$ decreases from $s$ to $-s$. In our implementation, $s$ is large value that satisfies $\sigma(s) \approx 1.0$ and $\sigma(-s) \approx 0.0$, which gurantees that $\kappa_{t}$ will decrease from $\kappa_{max}$ to $\kappa_{min}$.


\section{Theoretical Analysis}
Proofs of Lemma~\ref{lemma:1}, Theorem~\ref{theo:1}, Theorem~\ref{theo:2}, and Theorem~\ref{theo:3} are detailed in Appendix~A.
\begin{lemma}
\label{lemma:1}
 Suppose the operations in the base model is Lipschitz smooth, then $\ell_{\mathbf{\psi}_{[i]}}(f(x_i, \mathbf{\theta}), \tilde{y}_i)$ is Lipschitz smooth with respect to $\mathbf{\theta}$ for $\forall \psi_{[i]} > 1$ and $\forall x_{i} \in D_{S}\bigcup D^u_{T}$, i.e., there exists a finite constant $\rho_{1}$ and a finite constant $L_{1}$ that satisfy: 

\begin{small}
\vspace{-10pt}
\begin{eqnarray}
&&||\frac{\partial \ell_{\psi_{[i]}}(f(x_i, \mathbf{\theta}), \tilde{y}_i)}{\partial \mathbf{\theta}}||_2 \leq \rho_1, \nonumber \\
&&||\frac{\partial^2 l_{\psi_{[i]}}(f(x_i, \mathbf{\theta}), \tilde{y}_i)}{\partial \mathbf{\theta}^2}||_2 \leq L_1  \nonumber
\end{eqnarray}
\vspace{-10pt}
\end{small}

Also, for $\forall \psi_{[i]} > 1$ and $\forall x_{i} \in D^u_{T}$, $\ell_{\mathbf{\psi}_{[i]}}(f(x_i, \mathbf{\theta}), \tilde{y}_i)$ is Lipschitz smooth with respect to $\psi_{[i]}$, i.e., there exists a finite constant $\rho_{2}$ and a finite constant $L_{2}$ that satisfy:
\vspace{-5pt}
\begin{small}
\begin{eqnarray}
&&||\frac{\partial \ell_{\psi_{[i]}}(f(x_i, \mathbf{\theta}), \tilde{y}_i)}{\partial \mathbf{\psi}_{[i]}}||_2 \leq \rho_2, \nonumber \\
&&||\frac{\partial^2 \ell_{\psi_{[i]}}(f(x_i, \mathbf{\theta}), \tilde{y}_i)}{\partial \mathbf{\psi}_{[i]}^2}||_2 \leq L_2 \nonumber
\end{eqnarray}
\end{small}
\end{lemma}




\begin{assumption}
\label{assmp:2}
The learning rate $\eta_t$ (line 10 of Algorithm~\ref{algo:MTEM}) satisfies $\eta_t = min\{ 1,\frac{k_1}{t}\}$ for some $k_1 > 0$, where $\frac{k_1}{t} < 1$.  In addition, The learning rate $\beta_t$ (line 8 of Algorithm~\ref{algo:MTEM}) is a monotone descent sequence and $\beta_t = min\{\frac{1}{L},{\frac{k_2}{\sqrt[3]{t^2}}}\}$ for some $k_2 > 0$, where $L=\max\lbrace L_1, L_2 \rbrace$ and ${\frac{\sqrt[3]{t^2}}{k_2} \geq L}$.
\end{assumption}
Based on the Assumption~\ref{assmp:2} and Lemma~\ref{lemma:1}, we deduce Theorem~\ref{theo:1} and Theorem~\ref{theo:2}. 
Theorem~\ref{theo:1} demonstrates that, by adjusting $\psi$, the model trained on the target domain can be generalized to the source domain immediately. In other words, by adjusting $\psi$, the learning process on the target domain (i.e., Eq.~\eqref{eq:object_new}) has learned the domain agnostic features. At the same time, Theorem~\ref{theo:2} guarantees the convergence of the learning process on the target domain.


\begin{theorem}
\label{theo:1}
The training process in MTEM can achieve \begin{small} $\mathbb{E}[\|\nabla_{\psi} \mathcal{L}_{S}(\hat{\mathbf{\theta}}_{t}(\mathbf{\psi}_{t})|D_S)\|_2^2] \leq \epsilon $ in $\mathcal{O}(\frac{1}{\epsilon^3})$\end{small} steps:

\begin{small}
$$
   \min _{0 \leq t \leq T} \mathbb{E}[\|\nabla_{\mathbf{\psi}} \mathcal{L}_{S}(\hat{\mathbf{\theta}}_{t}(\mathbf{\psi}_{t})|D_S)\|_2^2] \leq \mathcal{O}(\frac{C}{\sqrt[3]{T}}) 
$$
\end{small}

\noindent where $C$ is an independent constant.
\end{theorem}

\begin{theorem}
\label{theo:2}
With the training process in MTEM, the instance adaptive Tsallis entropy is guaranteed to be converged on unlabeled data. Formally, 
\begin{small}
\begin{eqnarray}
\lim _{t \rightarrow \infty} \mathbb{E}[\|\nabla_{\mathbf{\theta}} \mathcal{L}_T(\mathbf{\theta}_{t}, \mathbf{\psi}_{t+1}| D_T^u)\|_2^2]=0
\end{eqnarray}
\end{small}
\end{theorem}

We use hypothesis $h:\mathcal{X} \rightarrow \Delta^{K-1}$ to analyze the effectiveness of MTEM in achieving domain adaptation. Formally, $h_{\mathbf{\theta}}(x_i) = \arg\max\limits_{k} f_{[k]}(x_i;\mathbf{\theta})$. We let $R_D(h)$ denote the model's robustness to the perturbations on dataset $D$. 
We let $\hat{\mathcal{R}}(\mathcal{H}|_D)$ denote the Rademacher complexity~\cite{gnecco2008approximation} of function class $\mathcal{H}$ ($h \in \mathcal{H}$) on dataset $D$. Radmacher complexity evaluates the ability of the worst hypothesis $h\in\mathcal{H}$ in fitting random labels. 
If there exists a $h \in \mathcal{H}$ that fits most random labels on $D$, then $\hat{\mathcal{R}}(\mathcal{H}|_D)$ is large. With the above definitions, we deduce Theorem~\ref{theo:3}.
\begin{theorem}
\label{theo:3}
Suppose $D_{S}$ and $D_{T}^{u}$ satisfy \textbf{$(q, c)-$ constant expansion}~\cite{DBLP:conf/iclr/WeiSCM21} for some constant $q, c \in (0, 1)$. With the probability at least $1 - \delta$ over the drawing of $D_T^{u}$ from $\mathbb{D}_{T}$, the error rates of the model $h_{\mathbf{\theta}}$ ($h\in \mathcal{H}$) on the target domain (i.e., $\epsilon_{\mathbb{D}_T}(h_{\mathbf{\theta}})$) is bounded by:

\begin{small}
\begin{eqnarray}
\epsilon_{\mathbb{D}_T}(h_{\mathbf{\theta}}) &\leq& \mathcal{L}_S({\mathbf{\theta}}|D_S) + \mathcal{L}_T(\mathbf{\theta}, \psi|D_T^u) + 2q + 2K\cdot\hat{\mathcal{R}}(\mathcal{H}|_{D_S}) \nonumber\\
&&+ 4K\cdot\hat{\mathcal{R}}(\tilde{\mathcal{H}}\times\mathcal{H}|_{D_T^u}) +\frac{R_{D_S \cup D_T^u}(h)}{\min \lbrace c, q\rbrace} +\zeta \label{eq:theo3_normal}
\end{eqnarray}
\end{small}

\noindent where \begin{small}$\zeta = \mathcal{O}(\sqrt{\frac{-log(\delta)}{|D_S|}} + \sqrt{\frac{-log(\delta)}{|D_T^u|}})$\end{small} is a low-order term. $\tilde{\mathcal{H}}\times\mathcal{H}$ refers to the function class $\lbrace x \rightarrow h(x)_{[h'(x)]}: h, h' \in \mathcal{H} \rbrace$.
\end{theorem}
With Theorem~\ref{theo:3}, we demonstrate that:
\begin{enumerate}
\item Theorem~\ref{theo:1} and Theorem~\ref{theo:2} prove that MTEM can simultaneously optimize the $\psi$ and $\mathbf{\theta}$ to minimize $\mathcal{L}_S(h|D_S) + \mathcal{L}_T(h, \psi|D_T^u)$, i.e., the first two term in Eq.~\eqref{eq:theo3_normal}. 

\item With the bi-level optimization process, the learning process on $D_T^u$ is regularized by supervised loss on the source domain. As $D_S$ does not overlap with $D_T^u$, fitting the random labels on $D_T^u$ cannot carry out the decrease of the supervised loss on the source domain (i.e., $\mathcal{L}_S({\mathbf{\theta}}|D_S)$). Thus, $h\in \mathcal{H}$ fits less noise on $D_{T}^{u}$, reducing $\hat{\mathcal{R}}(\tilde{\mathcal{H}}\times\mathcal{H}|_{D_T^u})$. At the same time, as $D_S$ is unseen in the training process, it is also hard to fit the random label on $D_S$, thereby reducing $\hat{\mathcal{R}}(\mathcal{H}|_{D_S})$.

\item Instance adaptive Tsallis-entropy is an unsupervised loss. As accessing unlabeled data is easier than accessing the labeled data, MTEM provides the possibility of sampling a larger unlabeled data to make $\zeta$ smaller.

\item $R_{D_S \cup D_T^u}(h)$ is a term that can be minimized in the training process technically, e.g., adversarial training~\cite{jiang2020bidirectional} or SAM (Sharpness-Aware-Minimization) optimizer~\cite{foret2020sharpness}.
\end{enumerate}

\section{Experiments}
\label{sec:experiments}

\subsection{Experiment Settings}
\label{subsec:exp_settings}
\paragraph{Datasets.}
On the rumor detection task, we conduct experiments with the dataset TWITTER~\cite{zubiaga2016learning}, which contains five domains: ``Cha.”, ``Ger.'', ``Fer.”, ``Ott.”, and ``Syd.”. 
On the sentiment classification task, we conduct experiments with the dataset Amazon~\cite{blitzer2007biographies}, which contains four domains: books, dvd, electronics, and kitchen. 
Preprocess and statistics on the TWITTER dataset and the Amazon dataset can be found in Appendix~D. 

\paragraph{Comparing Methods.} We compare MTEM with previous domain adaptation approaches on both \textit{semi-supervised}\footnote{{A small set of labeled data in the target domain can be accessed, named \textit{in-domain} dataset.}} and \textit{unsupervised} domain adaptation scenarios. Under the unsupervised domain adaptation, we compare MTEM with Out~\cite{chen2021wind}, DANN~\cite{ganin2016domain}, FixMatch~\cite{sohn2020fixmatch},  and CST~\cite{liu2021cycle}. Under the semi-supervised domain adaptation, MTEM\footnote{{For semi-supervised domain adaptation, we insert the labeled target data into $D_{S}$ to run MTEM.}} is compared with In+Out~\cite{chen2021wind}, MME~\cite{saito2019semi}, BiAT~\cite{jiang2020bidirectional}, and Wind~\cite{chen2021wind}. Out and In+Out are two straightforward ways for realizing unsupervised and semi-supervised domain adaptation, where Out means the base model is trained on the out-of-domain data (i.e., $D_{S}$) and In+Out means the base model is trained on both the in-domain and the out-of-domain data. DANN realizes domain adaptation by min-max the domain classification loss. CST and FixMatch are self-training approaches that generates pseudo instances to augment domain adaptation. Although CST also involves Tsallis entropy, the entropy-index is a mannually set hyper-parameters and is not instance-adaptative. WIND is a meta-reweigting based domain adaptation approach that learns-to-learn suitable instance weights of different labeled samples in the source domain. More details about the baseline methods can be found in the references.
\begin{table*}[tbh]
\small
\centering
\begin{tabular}{c|cclcc|ccccc}
\hline
\multirow{2}{*}{\begin{tabular}[c]{@{}c@{}}Base Model\\ (BiGCN)\end{tabular}} & \multicolumn{5}{c|}{Unsupervised domain adaptation}                    & \multicolumn{5}{c}{Semi-Supervised domain adaptation}                \\ \cline{2-11} 
& Out   & DANN  & FixMatch & \multicolumn{1}{c|}{CST}   & MTEM         & In+Out & MME   & BiAT  & \multicolumn{1}{c|}{Wind}  & MTEM         \\ \hline
Cha.   & 0.561 & 0.501 & 0.614    & \multicolumn{1}{c|}{0.573} & \textbf{0.627} & 0.586  & 0.601 & 0.547 & \multicolumn{1}{c|}{0.552} & \textbf{0.637} \\ \hline
Fer. & 0.190 & 0.387 & 0.473    & \multicolumn{1}{c|}{0.446} & \textbf{0.549} & 0.200  & 0.381 & 0.256 & \multicolumn{1}{c|}{0.291} & \textbf{0.635} \\ \hline
Ott. & 0.575 & 0.544 & 0.672    & \multicolumn{1}{c|}{0.649} & \textbf{0.728} & 0.599  & 0.612 & 0.614 & \multicolumn{1}{c|}{0.633} & \textbf{0.817} \\ \hline
Syd. & 0.438 & 0.461 & 0.694    & \multicolumn{1}{c|}{0.653} & \textbf{0.729} & 0.424  & 0.677 & 0.661 & \multicolumn{1}{c|}{0.628} & \textbf{0.750} \\ \hline
Mean & 0.441 & 0.473 & 0.613    & \multicolumn{1}{c|}{0.598} & \textbf{0.658} & 0.452  & 0.567 & 0.520 & \multicolumn{1}{c|}{0.526} & \textbf{0.709} \\ \hline
\end{tabular}
\caption{F1 scores on the TWITTER dataset}
\label{tab:TWITTERResults}
\end{table*}

\begin{table*}[tbh]
\small
\centering
\begin{tabular}{c|ccccc|ccccc}
\hline
\multirow{2}{*}{\begin{tabular}[c]{@{}c@{}}Base Model\\ (BERT)\end{tabular}} & \multicolumn{5}{l|}{Unsupervised Domain Adaptation}                                                                                      & \multicolumn{5}{l}{Semi-Supervised Domain Adaptation}  \\ \cline{2-11} & \multicolumn{1}{l}{Out} & \multicolumn{1}{l}{DANN} & \multicolumn{1}{l}{CST} & \multicolumn{1}{l|}{FixMatch} & \multicolumn{1}{l|}{MTEM} & \multicolumn{1}{l}{In+Out} & \multicolumn{1}{l}{MME} & \multicolumn{1}{l}{BiAT} & \multicolumn{1}{l|}{WIND}  & \multicolumn{1}{l}{MTEM} \\ \hline
books                                                                        & 0.902                   & 0.912                    & 0.912                   & \multicolumn{1}{c|}{0.906}    & \textbf{0.939}            & 0.912                      & 0.923                   & 0.922                    & \multicolumn{1}{c|}{0.917} & \textbf{0.946}           \\ \hline
dvd                                                                          & 0.902                   & 0.909                    & 0.923                   & \multicolumn{1}{c|}{0.907}    & \textbf{0.937}            & 0.908                      & 0.924                   & 0.903                    & \multicolumn{1}{c|}{0.911} & \textbf{0.947}           \\ \hline
electronics                                                                  & 0.894                   & 0.934                    & 0.923                   & \multicolumn{1}{c|}{0.913}    & \textbf{0.935}            & 0.926                      & 0.927                   & 0.930                    & \multicolumn{1}{c|}{0.931} & \textbf{0.945}           \\ \hline
kitchen                                                                      & 0.895                   & 0.934                    & 0.924                   & \multicolumn{1}{c|}{0.922}    & \textbf{0.937}            & 0.931                      & 0.931                   & 0.933                    & \multicolumn{1}{c|}{0.940} & \textbf{0.942}           \\ \hline
Mean                                                                         & 0.898                   & 0.922                    & 0.920                   & \multicolumn{1}{c|}{0.912}    & \textbf{0.937}            & 0.919                      & 0.926                   & 0.922                    & \multicolumn{1}{c|}{0.925} & \textbf{0.945}           \\ \hline
\end{tabular}
\caption{Accuracy scores on the Amazon dataset}
\label{tab:AmazonResults}
\end{table*}
\paragraph{Implementation Details.} 
The base model on the Amazon dataset is BERT~\cite{devlin2019bert} and the base model on the TWITTER dataset is BiGCN~\cite{bian2020rumor}. Domain adaptation experiments are conducted on every domain on the benchmark datasets. For every domain on the benchmark dataset, we seperately take them as the target domain and merges the rest domains  as the source domain. For example, when the ``books” domain in the Amazon dataset is taken as the target domain, the ``dvd”, ``electronics” and ``kitchen” domains are merged as the source domain.
All unlabeled data from the target domain are involved in the training process, meanwhile the labeled data in the target domain are used for evaluation (with a ratio of 7:3). Since the TWITTER dataset does not contain extra unlabeled data, we take 70\% of the labeled data on the target domain as the unlabeled data for training the model and preserve the rest ones for evaluation. The experiments on TWITTER are conducted on ``Cha.”, ``Fer.”, ``Ott.”, and ``Syd.”\footnote{{The labeled data in ``Ger.” domain is too scare to provide extra unlabeled data.}}. For the symbols in Algorithm~\ref{algo:MTEM}, we set $\eta_t$ and $\beta_{t}$ with respect to Assumption~\ref{assmp:2}. 


\subsection{General Results}
\label{subsec:general_rst}
We use all baseline approaches (including MTEM) to adapt BiGCN across domains on TWITTER,  and to adapt BERT across domains on Amazon. We validate the effectiveness of the proposed MTEM on both unsupervised and semi-supervised domain adaptation scenarios. For semi-supervised domain adaptation scenario, 100 labeled instances in the target domain are randomly selected as the in-domain dataset.  As the rumor detection task mainly concerns the classification performance in the `rumor' category, we use the F1 score to evaluate the performance on TWITTER. On the sentiment classification task, different classes are equally important. Thus, we use the accuracy score to evaluate different models, which is also convenient for comparison with previous studies.
Experiment results are listed in Table~\ref{tab:TWITTERResults}, Table~\ref{tab:AmazonResults}.

The results in Table~\ref{tab:TWITTERResults} and Table~\ref{tab:AmazonResults} demonstrate the effectiveness of the proposed MTEM algorithm. In particular, MTEM outperforms all baseline approaches on all benchmark datasets. Compared with the self-training approaches, i.e., FixMatch and CST, MTEM maintains the superiority of an average of nearly 2 percent on the Amazon dataset and an average of 4 percent on the TWITTER set. Thus, regularizing the self-training process with an instance adaptative is beneficial. Moreover, MTEM also surpasses the meta-reweighting algorithm, i.e., WIND, by an average of nearly 2 percent on the Amazon dataset and nearly 18 percent on the TWITTER dataset. Thus, the meta-learning algorithm in MTEM, i.e., learning to learn the suitable entropy indexes, is a competitive candidate in the domain adaptation scenario.

\subsection{Ablation Study}
\label{subsec:abs}
We separately remove the meta-learning module (\textit{- w/o M}), the temperature scheduler (\textit{- w/o T}), and the sampling mechanism (\textit{- w/o S}) to observe the adaptation performance across domains on the benchmark datasets. 
\textit{- w/o M} means all instances in the target domain will be allocated the same entropy index (determined with manually attempt). \textit{- w/o T} means removing the temperature scheduler, and the temperature $\kappa$ is fixed to be 1.0.  \textit{- w/o S} means to remove the sampling mechanism, i.e., generates pseudo labels with greedy strategies as previous self-training approaches. The experiments are conducted under the unsupervised domain adaptation scenarios. We validate the effectiveness with F1 score on TWITTER, and use the accuracy score on Amazon.
The experiment results are listed in Tab.~\ref{tab:abl_TWITTER} and Tab.~\ref{tab:abl_Amazon}.

From Tab.~\ref{tab:abl_TWITTER} and Tab.~\ref{tab:abl_Amazon}, we can find that all variants perform worse than MTEM on two benchmark datasets: (i) MTEM surpasses MTEM \textit{- w/o M} on the Amazon dataset with an average of 2 percent, and on the TWITTER dataset with an average of 7 percent. Thus, allocating an instance adaptative entropy index for every unlabeled instance in the target domain is superior to allocating the same entropy index. Furthermore, since the unified entropy index in MTEM \textit{- w/o M} is searched manually, MTEM \textit{- w/o M} should be better than Gibbs Entropy. Otherwise, the entropy index would be determined as 1.0 (Gibbs Entropy). Thus, the instance adaptive Tsallis-entropy in MTEM is better than Gibbs Entropy. (ii) MTEM surpasses MTEM \textit{- w/o S} on the Amazon dataset with an average of 1.4 percent, and on the TWITTER dataset with an average of 1.5 percent, which is attributed to the sampling mechanism can directly correct the model's prediction errors. (iii) MTEM surpasses MTEM \textit{- w/o T} with an average decrease of 0.9 percent on the TWITTER dataset, and with an average of 0.7 percent on the Amazon dataset, which is consistent with our claims that the annealing mechanism is beneficial to align the domains gradually.

\begin{table}[t]
\small
\centering
\begin{tabular}{c|llll|l}
\hline
Domain                               & \multicolumn{1}{c}{Cha.} & \multicolumn{1}{c}{Fer.} & \multicolumn{1}{c}{Ott.} & \multicolumn{1}{c|}{Syd.} & Mean  \\ \hline
MTEM  & \textbf{0.627} & \textbf{0.549} & \textbf{0.728} & \textbf{0.729} &\textbf{0.658} \\ \hline
\textit{- w/o M}    &0.569& 0.452& 0.633& 0.647& 0.575  \\ \hline
\textit{- w/o A}    &0.621  & 0.537& 0.716& 0.722& 0.649\\ 
\textit{- w/o S}    &0.622  & 0.529& 0.707& 0.714& 0.643\\ \hline
\end{tabular}
\caption{Ablation Study on TWITTER}
\label{tab:abl_TWITTER}
\end{table}
\begin{table}[t]
\small
\centering
\setlength{\tabcolsep}{4.pt}
\begin{tabular}{c|cccc|c}
\hline
Domain                               & books & dvd   & electronics & kitchen & Mean  \\ \hline
MTEM                               & \textbf{0.939} & \textbf{0.937} & \textbf{0.935}       & \textbf{0.937}   & \textbf{0.937} \\ \hline
\textit{- w/o M}    &  0.912&  0.917&   0.919&  0.919 & 0.916\\  \hline
\textit{- w/o A}    &0.931& 0.935& 0.927& 0.929& 0.930\\ 
\textit{- w/o S}    &0.929&  0.912&  0.927& 0.922  & 0.923\\ \hline
\end{tabular}
\caption{Ablation Study on the Amazon dataset}
\label{tab:abl_Amazon}
\end{table}

\subsection{Computation Cost}
\begin{figure*}[t]
    \centering
    \subfigure[Time Cost]{
        \includegraphics[height=2.28in]{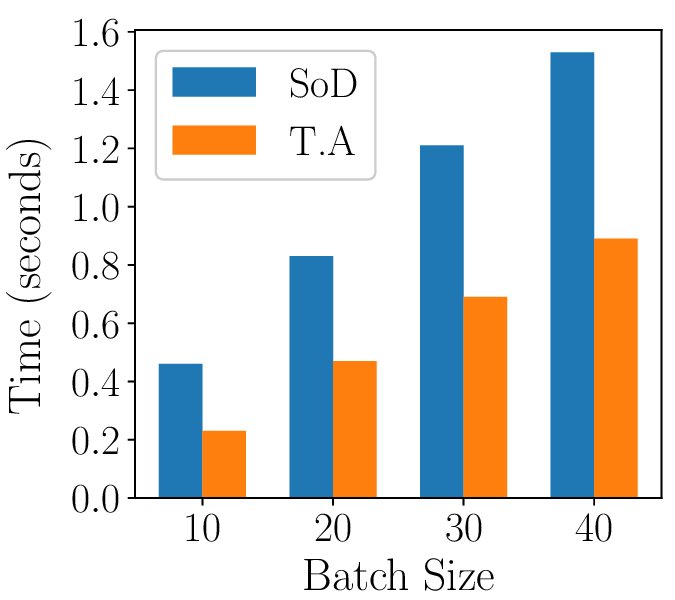}
        \label{subfig:time}
    }
    \subfigure[Memory Cost]{
        \includegraphics[height=2.28in]{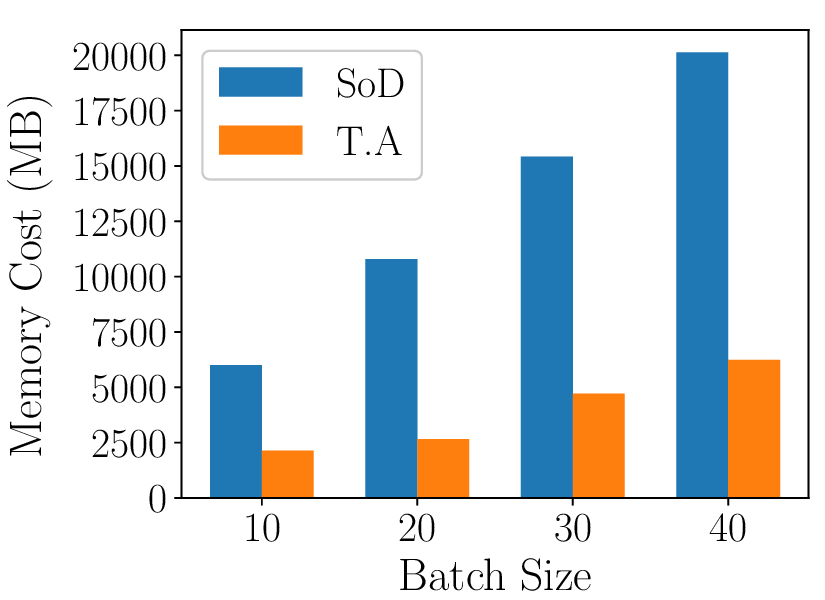}
        \label{subfig:mem}
    }
  \caption{Computation Cost: Taylor Approximation (T.A) v.s. Second-order Derivation (SoD).}
  \label{fig:cost}
\end{figure*}
We conduct experiments on the Amazon dataset to compare the computation cost in the Taylor approximation and the original Second-order derivation. 
We separately count the time and the memory consumed in computing the gradient of the entropy indexes with different batch sizes. Experiments are deployed on an Nvidia Tesla V100 GPU.
From Fig.~\ref{fig:cost}, the time cost in the Second-order derivation is almost two times higher than in the Taylor approximation, and the memory cost in the Second-order derivation is 3-4 times higher than in the Taylor approximation technique. 


We also count the different performances of adapting the BERT model to the `kitchen' domain with respect to different batch sizes. The experiment results are listed in Tab.~\ref{tab:multi2K_bs}, where `/' means the memory cost is out of the device's capacity. From Tab.~\ref{tab:multi2K_bs}, we can observe that the Taylor Approximation technique keeps a similar performance with the Second-order derivation. What's more, the best performance is achieved by using a batch with more than 50 instances (the setting in $\S$~\ref{subsec:general_rst}), which would exceed the memory capacity if we use the Second-order derivation. Thus, the benefit of reducing the computation cost is apparent, as a larger batch size leads to better adaptation performance. 
\begin{table}[t]
\small
\centering
\setlength\tabcolsep{3pt} 
\begin{tabular}{c|cccccc}
\hline
Batch Size & 10 & 20 & 30 & 40 & 50 & 60  \\ \hline
T.A &0.873 & 0.892& 0.914 & 0.924& 0.935&0.935\\ \hline
SoD  &  0.876&  0.897& 0.915 &0.927 &\textit{/}& \textit{/}\\ \hline
\end{tabular}
\caption{Domain Adaptation with Different Batch Size: Taylor Approximation (T.A) v.s. Second-order Derivation (SoD)}
\label{tab:multi2K_bs}
\end{table}

\begin{table}[t]
\small
\begin{tabular}{c|l}
\hline
$\psi \approx$1.0 & \begin{tabular}[c]{@{}l@{}}i bought this at amazon , but it's \textcolor{red}{cheaper} at cutlery \\
and more, \$ 9.95 , so is the \$ 89.00 wusthof santoku \\
7 knife ( \$ 79.00), and they have \textcolor{red}{free shipping} ! check \\
yahoo shopping before amazon ! ! !\end{tabular}                                                                              \\ \hline
$\psi$=5.0 & \begin{tabular}[c]{@{}l@{}}i like and dislike these bowls, what i like about them is \\the shape and size for certain foods and for the dish-\\
-washer. they are too small for cereal if you would like \\
to add fruit to your cereal, perfect for \\ oatmeal or ice cream \textcolor{red}{but too small for soup or stew}.\end{tabular} \\ \hline
\end{tabular}
\caption{Unlabeled instances with different entropy index ($\psi$).}
\label{tab:case_study}
\end{table}

\subsection{Case Study}

In Tab.~\ref{tab:case_study}, we present two cases with different entropy indexes learned in the meta-learning process (more cases are provided in Appendix.~E). Experiments are conducted on sentiment classification tasks, and the settings are the same as `kitchen' in $\S$~\ref{subsec:general_rst}. On the sentences with smaller entropy index (updating the model aggressively), the sentiment words are more transferrable across domains in e-commerce , e.g., `cheaper' and `free shipping'. Otherwise, sentences with larger entropy index contain more domain discriminative words, e.g., the n-gram `but too small for soup or stew' in the kitchen domain are less relevant to the other domains (electronics, books, dvd). In this case, MTEM uses a large entropy index to update the model more cautiously.

\section{Related Work}
\subsection{Domain Adaptation} 
To adapt a model to a new domain, 
feature-alignment approaches~\cite{ganin2016domain,saito2019semi,saito2019semi} focus on explicitly aligning the feature space across domains. For example, DANN~\cite{ganin2016domain} proposes to align the feature space by min-max the domain classification loss. With similar efforts, MME~\cite{saito2019semi} min-max the conditional entropy on the unlabeled data. BiAT~\cite{jiang2020bidirectional} proposes to decouple the min-max optimization process in DANN, i.e., firstly maximize the risk loss to obtain a gradient-based perturbation on the input space and then minimize the objective on the perturbed input cases.
On the other hand, data-centric approaches use the unlabeled data in the target domain or the labeled data from the source domain to align the feature space implicitly. 
To select labeled data from the source domain, researchers~\cite{DBLP:conf/acl/MooreL10} design a technique based on
topic models for measuring the domain similarity,  while~\cite{chen2021wind} takes a meta-learning algorithm to implicitly measure the domain similarity.
To exploit the unlabeled data from the target domain, pseudo labeling approaches, including self-training~\cite{zou2019confidence}, co-training~\cite{chen2011co}, and tri-training~\cite{saito2017asymmetric}, are widely applied and become an important direction. The difference lies in that self-training~\cite{zou2018unsupervised,zou2019confidence,liu2021cycle} uses the model’s prediction to improve the model, while co-training~\cite{chen2011co} and tri-training~\cite{saito2017asymmetric} involves more models which learn the task information from each other. In the research of self-training for domain adaptation, many efforts tried to use prediction confidence to reduce the label noise of pseudo instances~\cite{zou2018unsupervised,zou2019confidence,liu2021cycle}, i.e., they preserve only the easy examples that have high prediction confidences while discarding the hard examples that have low prediction confidences. However, fitting the model on easy pseudo instances cannot effectively improve the performance, as the model is already confident about its prediction. 
\subsection{Meta-Learning}
Meta-learning is an emerging new branch in machine learning that aims to train a model that can adapt to a new task or new domain quickly given a few new samples. For this purpose, previous studies tried to learn better initial model parameters~\cite{finn2017model}, or better learning rates~\cite{DBLP:journals/corr/LiZCL17}. \textit{Learning to Compare} methods, e.g., relation network~\cite{sung2018learning} and prototypical learning~\cite{snell2017prototypical}, are investigated more widely in text
classification tasks\cite{tan2019out,geng2020dynamic}. With the recent success of the pre-trained language model, Network Architect Search (NAS) methods, e.g., DARTs~\cite{liu2018darts}), are also widely studied in Natural Language Processing (NLP) tasks~\cite{xu2021bert,dong2021efficientbert}. Some meta-learning algorithm learns to knowledge distillation~\cite{zhou2022bert}, i.e., increase the number of teacher models to train a meta-teacher model that works better than a single teacher model. Meta reweighting algorithm~\cite{ren2018learning}, which proposes to dynamically reweight the risk on different instances, has also inspired some NLP tasks~\cite{li2020meta,chen2021wind}. 
Our research is similar to the meta reweighting algorithm, i.e., the training objective is instance adaptive, the difference lies in that the entropy index controls the loss function on different instances while the instance weights do not change the loss function. 

\section{Conclusion}
This paper proposes a new meta-learning algorithm for domain adaptation on text classification, namely MTEM. Inheriting the principle of entropy minimization, MTEM imposes an instance adaptative Tsallis entropy minimization process on the target domain, and such a process is formulated as a meta-learning process. To reduce the computation cost, we propose a Taylor approximation technique to compute the gradient of the entropy indexes. Also, we propose an annealing sampling mechanism to generate pseudo labels. In addition, we analyze the proposed MTEM theoretically, i.e., we prove the convergence of the meta-learning algorithm in optimizing the instance-adaptative entropy and provide insights for understanding why MTEM is effective in achieving domain adaptation. Extensive experiments on two popular models, BiGCN and BERT, verify the effectiveness of MTEM.
\section*{Acknowledgements} 
This work is supported by the following foundations: the National Natural Science Foundation of China under Grant No. 62025208, the Xiangjiang Laboratory Foundation under Grant No. 22XJ01012, 2022 International Postdoctoral Exchange Fellowship Program (Talent-Introduction Program) under Grant No. YJ20220260.

\section*{Contribution Statement}
Menglong Lu and Zhen Huang contributed equally to this work. Zhiliang Tian and  Yunxiang Zhao are the corresponding authors.

\newpage
\bibliographystyle{named}
\bibliography{ijcai22}

\begin{thebibliography}{}

\bibitem[\protect\citeauthoryear{Benson}{1995}]{benson1995concave}
Harold~P Benson.
\newblock Concave minimization: theory, applications and algorithms.
\newblock In {\em Handbook of Global Optimization}, pages 43--148. Springer,
  1995.

\bibitem[\protect\citeauthoryear{Bian \bgroup \em et al.\egroup
  }{2020}]{bian2020rumor}
Tian Bian, Xi~Xiao, Tingyang Xu, Peilin Zhao, Wenbing Huang, Yu~Rong, and
  Junzhou Huang.
\newblock Rumor detection on social media with bi-directional graph
  convolutional networks.
\newblock In {\em Proceedings of the AAAI Conference on Artificial
  Intelligence}, pages 549--556, 2020.

\bibitem[\protect\citeauthoryear{Blitzer \bgroup \em et al.\egroup
  }{2007}]{blitzer2007biographies}
John Blitzer, Mark Dredze, and Fernando Pereira.
\newblock Biographies, bollywood, boom-boxes and blenders: Domain adaptation
  for sentiment classification.
\newblock In {\em Proceedings of the Annual Meeting of the Association of
  Computational Linguistics}, pages 440--447, 2007.

\bibitem[\protect\citeauthoryear{Chen \bgroup \em et al.\egroup
  }{2011}]{chen2011co}
Minmin Chen, Kilian~Q Weinberger, and John~C Blitzer.
\newblock Co-training for domain adaptation.
\newblock In {\em Proceedings of the International Conference on Neural
  Information Processing Systems}, pages 2456--2464, 2011.

\bibitem[\protect\citeauthoryear{Chen \bgroup \em et al.\egroup
  }{2021}]{chen2021wind}
Xiang Chen, Yue Cao, and Xiaojun Wan.
\newblock Wind: Weighting instances differentially for model-agnostic domain
  adaptation.
\newblock In {\em Findings of the Annual Meeting of the Association for
  Computational Linguistics}, pages 2366--2376, 2021.

\bibitem[\protect\citeauthoryear{Devlin \bgroup \em et al.\egroup
  }{2019}]{devlin2019bert}
Jacob Devlin, Ming-Wei Chang, Kenton Lee, and Kristina Toutanova.
\newblock Bert: Pre-training of deep bidirectional transformers for language
  understanding.
\newblock In {\em Proceedings of the North American Chapter of the Association
  for Computational Linguistics}, pages 4171--4186, 2019.

\bibitem[\protect\citeauthoryear{Dong \bgroup \em et al.\egroup
  }{2021}]{dong2021efficientbert}
Chenhe Dong, Guangrun Wang, Hang Xu, Jiefeng Peng, Xiaozhe Ren, and Xiaodan
  Liang.
\newblock Efficientbert: Progressively searching multilayer perceptron via
  warm-up knowledge distillation.
\newblock In {\em Findings of the Association for Computational Linguistics},
  pages 1424--1437, 2021.

\bibitem[\protect\citeauthoryear{Du \bgroup \em et al.\egroup
  }{2020}]{du2020adversarial}
Chunning Du, Haifeng Sun, Jingyu Wang, Qi~Qi, and Jianxin Liao.
\newblock Adversarial and domain-aware bert for cross-domain sentiment
  analysis.
\newblock In {\em Proceedings of the Annual Meeting of the Association for
  Computational Linguistics}, pages 4019--4028, 2020.

\bibitem[\protect\citeauthoryear{Finn \bgroup \em et al.\egroup
  }{2017}]{finn2017model}
Chelsea Finn, Pieter Abbeel, and Sergey Levine.
\newblock Model-agnostic meta-learning for fast adaptation of deep networks.
\newblock In {\em International Conference on Machine Learning}, pages
  1126--1135, 2017.

\bibitem[\protect\citeauthoryear{Foret \bgroup \em et al.\egroup
  }{2020}]{foret2020sharpness}
Pierre Foret, Ariel Kleiner, Hossein Mobahi, and Behnam Neyshabur.
\newblock Sharpness-aware minimization for efficiently improving
  generalization.
\newblock In {\em International Conference on Learning Representations}, 2020.

\bibitem[\protect\citeauthoryear{Ganin \bgroup \em et al.\egroup
  }{2016}]{ganin2016domain}
Yaroslav Ganin, Evgeniya Ustinova, Hana Ajakan, Pascal Germain, Hugo
  Larochelle, Fran{\c{c}}ois Laviolette, Mario Marchand, and Victor Lempitsky.
\newblock Domain-adversarial training of neural networks.
\newblock {\em Journal of Machine Learning Research}, 17:2096--2030, 2016.

\bibitem[\protect\citeauthoryear{Geng \bgroup \em et al.\egroup
  }{2020}]{geng2020dynamic}
Ruiying Geng, Binhua Li, Yongbin Li, Jian Sun, and Xiaodan Zhu.
\newblock Dynamic memory induction networks for few-shot text classification.
\newblock In {\em Proceedings of the Annual Meeting of the Association for
  Computational Linguistics}, pages 1087--1094, 2020.

\bibitem[\protect\citeauthoryear{Gnecco \bgroup \em et al.\egroup
  }{2008}]{gnecco2008approximation}
Giorgio Gnecco, Marcello Sanguineti, et~al.
\newblock Approximation error bounds via rademacher complexity.
\newblock {\em Applied Mathematical Sciences}, 2:153--176, 2008.

\bibitem[\protect\citeauthoryear{Jiang \bgroup \em et al.\egroup
  }{2020}]{jiang2020bidirectional}
Pin Jiang, Aming Wu, Yahong Han, Yunfeng Shao, Meiyu Qi, and Bingshuai Li.
\newblock Bidirectional adversarial training for semi-supervised domain
  adaptation.
\newblock In {\em Proceedings of the International Joint Conference on
  Artificial Intelligence}, pages 934--940, 2020.

\bibitem[\protect\citeauthoryear{Kipf and Welling}{2017}]{kipf2016semi}
Thomas~N. Kipf and Max Welling.
\newblock Semi-supervised classification with graph convolutional networks.
\newblock In {\em Proceedings of the International Conference on Learning
  Representations}, pages 1592--1601, 2017.

\bibitem[\protect\citeauthoryear{Kumar \bgroup \em et al.\egroup
  }{2010}]{kumar2010self}
M~Kumar, Benjamin Packer, and Daphne Koller.
\newblock Self-paced learning for latent variable models.
\newblock {\em Advances in Neural Information Processing Systems}, 23, 2010.

\bibitem[\protect\citeauthoryear{Kumar \bgroup \em et al.\egroup
  }{2020}]{kumar2020understanding}
Ananya Kumar, Tengyu Ma, and Percy Liang.
\newblock Understanding self-training for gradual domain adaptation.
\newblock In {\em International Conference on Machine Learning}, pages
  5468--5479, 2020.

\bibitem[\protect\citeauthoryear{Lee and others}{2013}]{lee2013pseudo}
Dong-Hyun Lee et~al.
\newblock Pseudo-label: The simple and efficient semi-supervised learning
  method for deep neural networks.
\newblock In {\em International Conference on Machine Learning}, page 896,
  2013.

\bibitem[\protect\citeauthoryear{Li \bgroup \em et al.\egroup
  }{2017}]{DBLP:journals/corr/LiZCL17}
Zhenguo Li, Fengwei Zhou, Fei Chen, and Hang Li.
\newblock Meta-sgd: Learning to learn quickly for few shot learning.
\newblock {\em CoRR}, abs/1707.09835, 2017.

\bibitem[\protect\citeauthoryear{Li \bgroup \em et al.\egroup
  }{2020}]{li2020meta}
Zhenzhen Li, Jian-Yun Nie, Benyou Wang, Pan Du, Yuhan Zhang, Lixin Zou, and
  Dongsheng Li.
\newblock Meta-learning for neural relation classification with distant
  supervision.
\newblock In {\em Proceedings of the ACM International Conference on
  Information \& Knowledge Management}, pages 815--824, 2020.

\bibitem[\protect\citeauthoryear{Liu \bgroup \em et al.\egroup
  }{2018}]{liu2018darts}
Hanxiao Liu, Karen Simonyan, and Yiming Yang.
\newblock Darts: Differentiable architecture search.
\newblock In {\em Proceedings of the International Conference on Learning
  Representations}, pages 934--940, 2018.

\bibitem[\protect\citeauthoryear{Liu \bgroup \em et al.\egroup
  }{2021}]{liu2021cycle}
Hong Liu, Jianmin Wang, and Mingsheng Long.
\newblock Cycle self-training for domain adaptation.
\newblock {\em Advances in Neural Information Processing Systems},
  34:22968--22981, 2021.

\bibitem[\protect\citeauthoryear{Lu \bgroup \em et al.\egroup
  }{2022}]{lu2022sifter}
Menglong Lu, Zhen Huang, Binyang Li, Yunxiang Zhao, Zheng Qin, and DongSheng
  Li.
\newblock Sifter: A framework for robust rumor detection.
\newblock {\em IEEE/ACM Transactions on Audio, Speech, and Language
  Processing}, 30:429--442, 2022.

\bibitem[\protect\citeauthoryear{{McClosky} \bgroup \em et al.\egroup
  }{2006}]{mcclosky2006reranking}
David {McClosky}, Eugene {Charniak}, and Mark {Johnson}.
\newblock Reranking and self-training for parser adaptation.
\newblock In {\em Proceedings of the Annual Meeting of the Association for
  Computational Linguistics}, pages 337--344, 2006.

\bibitem[\protect\citeauthoryear{Moore and
  Lewis}{2010}]{DBLP:conf/acl/MooreL10}
Robert~C. Moore and William~D. Lewis.
\newblock Intelligent selection of language model training data.
\newblock In {\em Proceedings of the Annual Meeting of the Association for
  Computational Linguistics}, pages 220--224, 2010.

\bibitem[\protect\citeauthoryear{Mukherjee and
  Awadallah}{2020}]{mukherjee2020uncertainty}
Subhabrata Mukherjee and Ahmed Awadallah.
\newblock Uncertainty-aware self-training for few-shot text classification.
\newblock {\em Advances in Neural Information Processing Systems},
  33:21199--21212, 2020.

\bibitem[\protect\citeauthoryear{Plastino and
  Plastino}{1999}]{plastino1999tsallis}
ARPA Plastino and AR~Plastino.
\newblock Tsallis entropy and jaynes' information theory formalism.
\newblock {\em Brazilian Journal of Physics}, 29:50--60, 1999.

\bibitem[\protect\citeauthoryear{{Reichart} and
  {Rappoport}}{2007}]{reichart2007self}
Roi {Reichart} and Ari {Rappoport}.
\newblock Self-training for enhancement and domain adaptation of statistical
  parsers trained on small datasets.
\newblock In {\em Proceedings of the Annual Meeting of the Association of
  Computational Linguistics}, pages 616--623, 2007.

\bibitem[\protect\citeauthoryear{Ren \bgroup \em et al.\egroup
  }{2018}]{ren2018learning}
Mengye Ren, Wenyuan Zeng, Bin Yang, and Raquel Urtasun.
\newblock Learning to reweight examples for robust deep learning.
\newblock In {\em Proceedings of the International Conference on Machine
  Learning}, pages 4334--4343, 2018.

\bibitem[\protect\citeauthoryear{{Rotman} and
  {Reichart}}{2019}]{rotman2019deep}
Guy {Rotman} and Roi {Reichart}.
\newblock Deep contextualized self-training for low resource dependency
  parsing.
\newblock {\em Transactions of the Association for Computational Linguistics},
  7:695--713, 2019.

\bibitem[\protect\citeauthoryear{RoyChowdhury \bgroup \em et al.\egroup
  }{2019}]{roychowdhury2019automatic}
Aruni RoyChowdhury, Prithvijit Chakrabarty, Ashish Singh, SouYoung Jin, Huaizu
  Jiang, Liangliang Cao, and Erik Learned-Miller.
\newblock Automatic adaptation of object detectors to new domains using
  self-training.
\newblock In {\em Proceedings of the IEEE/CVF Conference on Computer Vision and
  Pattern Recognition}, pages 780--790, 2019.

\bibitem[\protect\citeauthoryear{Saito \bgroup \em et al.\egroup
  }{2017}]{saito2017asymmetric}
Kuniaki Saito, Yoshitaka Ushiku, and Tatsuya Harada.
\newblock Asymmetric tri-training for unsupervised domain adaptation.
\newblock In {\em International Conference on Machine Learning}, pages
  2988--2997, 2017.

\bibitem[\protect\citeauthoryear{Saito \bgroup \em et al.\egroup
  }{2019}]{saito2019semi}
Kuniaki Saito, Donghyun Kim, Stan Sclaroff, Trevor Darrell, and Kate Saenko.
\newblock Semi-supervised domain adaptation via minimax entropy.
\newblock In {\em Proceedings of the IEEE/CVF International Conference on
  Computer Vision}, pages 8050--8058, 2019.

\bibitem[\protect\citeauthoryear{Shin \bgroup \em et al.\egroup
  }{2020}]{shin2020two}
Inkyu Shin, Sanghyun Woo, Fei Pan, and In~So Kweon.
\newblock Two-phase pseudo label densification for self-training based domain
  adaptation.
\newblock In {\em European Conference on Computer Vision}, pages 532--548,
  2020.

\bibitem[\protect\citeauthoryear{Shu \bgroup \em et al.\egroup
  }{2019}]{shu2019meta}
Jun Shu, Qi~Xie, Lixuan Yi, Qian Zhao, Sanping Zhou, Zongben Xu, and Deyu Meng.
\newblock Meta-weight-net: learning an explicit mapping for sample weighting.
\newblock In {\em Proceedings of the International Conference on Neural
  Information Processing Systems}, pages 1919--1930, 2019.

\bibitem[\protect\citeauthoryear{Snell \bgroup \em et al.\egroup
  }{2017}]{snell2017prototypical}
Jake Snell, Kevin Swersky, and Richard Zemel.
\newblock Prototypical networks for few-shot learning.
\newblock {\em Advances in Neural Information Processing Systems}, 2017.

\bibitem[\protect\citeauthoryear{Sohn \bgroup \em et al.\egroup
  }{2020}]{sohn2020fixmatch}
Kihyuk Sohn, David Berthelot, Nicholas Carlini, Zizhao Zhang, Han Zhang,
  Colin~A Raffel, Ekin~Dogus Cubuk, Alexey Kurakin, and Chun-Liang Li.
\newblock Fixmatch: Simplifying semi-supervised learning with consistency and
  confidence.
\newblock {\em Advances in Neural Information Processing Systems}, 33:596--608,
  2020.

\bibitem[\protect\citeauthoryear{Sung \bgroup \em et al.\egroup
  }{2018}]{sung2018learning}
Flood Sung, Yongxin Yang, Li~Zhang, Tao Xiang, Philip~HS Torr, and Timothy~M
  Hospedales.
\newblock Learning to compare: Relation network for few-shot learning.
\newblock In {\em Proceedings of the IEEE Conference on Computer Vision and
  Pattern Recognition}, pages 1199--1208, 2018.

\bibitem[\protect\citeauthoryear{Tan \bgroup \em et al.\egroup
  }{2019}]{tan2019out}
Ming Tan, Yang Yu, Haoyu Wang, Dakuo Wang, Saloni Potdar, Shiyu Chang, and
  Mo~Yu.
\newblock Out-of-domain detection for low-resource text classification tasks.
\newblock In {\em Conference on Empirical Methods in Natural Language
  Processing}, 2019.

\bibitem[\protect\citeauthoryear{Wei \bgroup \em et al.\egroup
  }{2021}]{DBLP:conf/iclr/WeiSCM21}
Colin Wei, Kendrick Shen, Yining Chen, and Tengyu Ma.
\newblock Theoretical analysis of self-training with deep networks on unlabeled
  data.
\newblock In {\em International Conference on Learning Representations}, 2021.

\bibitem[\protect\citeauthoryear{Xu \bgroup \em et al.\egroup
  }{2021}]{xu2021bert}
Jin Xu, Xu~Tan, Renqian Luo, Kaitao Song, Jian Li, Tao Qin, and Tie-Yan Liu.
\newblock Nas-bert: task-agnostic and adaptive-size bert compression with
  neural architecture search.
\newblock In {\em Proceedings of the ACM SIGKDD Conference on Knowledge
  Discovery \& Data Mining}, pages 1933--1943, 2021.

\bibitem[\protect\citeauthoryear{Zhou \bgroup \em et al.\egroup
  }{2022}]{zhou2022bert}
Wangchunshu Zhou, Canwen Xu, and Julian McAuley.
\newblock Bert learns to teach: Knowledge distillation with meta learning.
\newblock In {\em Proceedings of the Annual Meeting of the Association for
  Computational Linguistics}, pages 7037--7049, 2022.

\bibitem[\protect\citeauthoryear{Zou \bgroup \em et al.\egroup
  }{2018}]{zou2018unsupervised}
Yang Zou, Zhiding Yu, BVK Kumar, and Jinsong Wang.
\newblock Unsupervised domain adaptation for semantic segmentation via
  class-balanced self-training.
\newblock In {\em Proceedings of the European Conference on Computer Vision},
  pages 289--305, 2018.

\bibitem[\protect\citeauthoryear{Zou \bgroup \em et al.\egroup
  }{2019}]{zou2019confidence}
Yang Zou, Zhiding Yu, Xiaofeng Liu, BVK Kumar, and Jinsong Wang.
\newblock Confidence regularized self-training.
\newblock In {\em Proceedings of the IEEE/CVF International Conference on
  Computer Vision}, pages 5982--5991, 2019.

\bibitem[\protect\citeauthoryear{Zubiaga \bgroup \em et al.\egroup
  }{2016}]{zubiaga2016learning}
Arkaitz Zubiaga, Maria Liakata, and Rob Procter.
\newblock Learning reporting dynamics during breaking news for rumour detection
  in social media.
\newblock {\em CoRR}, abs/1610.07363, 2016.

\end{thebibliography}
\newpage
\textnormal{  }
\newpage
\appendix
\renewcommand\thelemma{\Alph{lemma}}
\section{Proofs}
\label{appendix:A}
\begin{table}[hbt]
\caption{Mathematic Symbol List.}
\label{tab:symbol_list}
\scriptsize
\begin{tabular}{c|l}
\hline
$\alpha$ & static entropy index \\ 
$\psi$ & dynamic entropy indexes  \\ 
$\psi_{[i]}$ & dynamic entropy index on instance $i$ \\ 
$e_{\psi_{[i]}}$ & Tsallis entropy on instance $i$ with $\psi_{[i]}$ as the entropy index \\ 
$\ell_{\psi_{[i]}}$ & Tsallis loss on instance $i$ with $\psi_{[i]}$ as the entropy index \\ \hline
$\mathbb{D}_{T}$ & target domain \\
$\mathbb{D}_{S}$ & source domain \\
$D_{T}^{u}$ & unlabeled instances accessed from the target domain\\
$D_{S}$ & labeled instances collected in the source domain\\ 
$K$ & number of classes \\ \hline
$x$ & the input of an instance \\
$y$ & the label of an instance \\
$z$ & the index of the non-zero element of the one-hot label $y$ \\ \hline
$\mathcal{L}_{S}$ & training loss on labeled data from the source domain \\ 
$\mathcal{L}_{T}$ & training loss on unlabeled data from the target domain \\
$\epsilon_{D}$ & the error rate on dataset $D$ \\ \hline
$f$ & model\\
$\theta$ & model's parameters\\
$f(x;\theta)$ & model's prediction probability on instance $x$ \\
$h_{\theta}$ & hypothesis with $\theta$ as the parameters \\
$\kappa$ & temperature controlling the smoothness of prediction\\\hline
$\eta$ & learning rate for updating the model's parameters \\
$\beta$ & learning rate to update the entropy indexes \\ \hline
\end{tabular}
\end{table}

\setcounter{assumption}{0}
\begin{assumption}
\label{assmp:model_Assmp}
For a model with parameters $\theta$ (i.e., $f(x;\theta)$), we assume that: (i) the model's prediction probability on every dimension is larger than $0$; (ii) the model's gradient back-propagated from its prediction is bounded by $\mathfrak{f}$, i.e., $||\bigtriangledown_{\mathbf{\theta}}f(x_i; \mathbf{\theta})||_2 \leq \mathfrak{f}$, where $\mathfrak{f}$ is a finit constant. (iii) its second-order derivation is bounded by $\mathfrak{b}$, i.e., $\bigtriangledown^2_{\mathbf{\theta}}f(x_i; \mathbf{\theta}) \leq \mathfrak{b}$, where $\mathfrak{b}$ is also a finit constant. 
\end{assumption}

Assumption~\ref{assmp:model_Assmp} is easy to be satisfied. Condition (i) and (ii) can be met technically, e.g., clipping the values that are too small or too large in model's prediction and gradients. Condition (iii) requires all operations involved in the model should be smooth. To our knowledge, existing pretrained language models satisfy such requirements.

\begin{assumption}
\label{assmp:2}
The learning rate $\eta_t$ (line 10 of Algorithm~1) satisfies $\eta_t = min\{ 1,\frac{k_1}{t}\}$ for some $k_1 > 0$, where $\frac{k_1}{t} < 1$.  In addition, The learning rate $\beta_t$ (line 8 of Algorithm~1) is a monotone descent sequence and $\beta_t = min\{\frac{1}{L},{\frac{k_2}{\sqrt[3]{t^2}}}\}$ for some $k_2 > 0$, where $L=\max\lbrace L_1, L_2 \rbrace$ and ${\frac{\sqrt[3]{t^2}}{k_2} \geq L}$.
\end{assumption}

\label{appendix:proofs}
\subsection{Proof for Theorem 1}
\label{appendix:proof_1}

\setcounter{lemma}{0}
\begin{lemma}
\label{lemma:collary_func}
For $\forall o \in [\wr, 1]^K$ where $\wr$ is a constant scalar greater than zero ($\wr > 0$), $|o^{\mathbf{\alpha}-2}log^m(o)|$ is bounded for $\mathbf{\alpha} > 1$, where $K \in \lbrace 2, 3, \ldots\rbrace$ and $m \in \lbrace 1, 2, \ldots\rbrace$.
\end{lemma}
\begin{proof}
Let $g(o) = o^{\mathbf{\alpha}-2}log^m(o)$. Then, we can derive the gradient of $o$ on $g(o)$ as below:
\begin{tiny}
\begin{eqnarray}
    &&\bigtriangledown_{o}g(o) \nonumber\\
    &=& (\bigtriangledown_{o}o^{\mathbf{\alpha}-2})\cdot log^{m}(o) + (\bigtriangledown_{o} log^m(o))\cdot o^{\mathbf{\alpha}-2} \\
    &=& ((\mathbf{\alpha}-2)o^{\mathbf{\alpha}-3})\cdot log^{m}(o) + (m\cdot log^{m-1}(o)\cdot\frac{1}{o})\cdot o^{\mathbf{\alpha}-2} \\
    &=& o^{\mathbf{\alpha}-3}log^{m-1}(o)[(\mathbf{\alpha}-2)log(o) + m] \mathcal{\epsilon}
\end{eqnarray}
\end{tiny}
For $o \in [\wr, 1]$, $o^{\mathbf{\alpha}-3}log^{m-1}(o) < 0$ is always true. For $o \leq e^{\frac{m}{2-\mathbf{\alpha}}}$, we have $((\mathbf{\alpha}-2)log(o) + m) \leq 0$. Thus, $g(o)$ monotonically decreases in the region $(0, e^\frac{m}{2-\mathbf{\alpha}}]$, and monotonically increases in the region $[e^\frac{m}{2-\mathbf{\alpha}}, \infty)$.

If $\wr \leq e^\frac{m}{2-\mathbf{\alpha}} \leq 1$, we can conclude:
\begin{tiny}
    \begin{eqnarray}
        |g(o)| \leq \max\lbrace |g(\wr)|, |g(1)|, |g(e^\frac{m}{2-\mathbf{\alpha}})|\rbrace \nonumber
    \end{eqnarray}
\end{tiny}
Thus, $|g(o)|$ is bounded.

If $e^\frac{m}{2-\mathbf{\alpha}} < \wr$ or $1 < e^\frac{m}{2-\mathbf{\alpha}}$, g(o) is monotonical in the region $[\wr, 1]$. Thus, 
\begin{tiny}
    \begin{eqnarray}
        |g(o)| \leq \max\lbrace |g(\wr)|, |g(1)|\rbrace \nonumber
    \end{eqnarray}
\end{tiny}
In this case, $|g(o)|$ is also bounded in the region$[\wr, 1]$.
\end{proof}

\begin{lemma}
\label{lemma:bound_func}
For $\forall \psi_{[i]}>1, \forall x_{i} \in D_T^u$, $\ell_{\psi_{[i]}}(f(x_i;\theta), \tilde{y}_i)$, $\frac{\partial^2 \mathcal{L}_T(\theta, \psi | D_T^u)}{\partial\mathbf{\theta} \partial\mathbf{\psi}}$, and $\frac{\partial^3\mathcal{L}_T(\theta, \psi | D_T^u)}{\partial\mathbf{\theta}\partial^2\mathbf{\psi}}$ are all bounded terms.
\end{lemma}
\begin{proof}
According to the definition of $\ell_{\psi_{[i]}}$ in Eq.~(3),
\begin{tiny}
\begin{eqnarray}
&&||\ell_{\psi_{[i]}}(f(x_i;\theta), \tilde{y}_i)||_2\nonumber \\
&=&||\frac{1}{\psi_{[i]} - 1} \cdot (1 - \tilde{y}_i^T\cdot f^{\psi_{[i]}-1}(x_i;\theta))||_2\\
&\leq& ||\frac{1}{\psi_{[i]} - 1}||_2\times ||(1 - \tilde{y}_i^T\cdot f^{\psi_{[i]}-1}(x_i;\theta))||_2\\
&\leq& ||\frac{1}{\psi_{[i]} - 1}||_2\times[ 1 + ||\tilde{y}_i^T\cdot f^{\psi_{[i]}-1}(x_i;\theta)||_2]
\end{eqnarray}
\end{tiny}

Since $\psi_{[i]} > 1$, $|\frac{1}{\psi_{[i]} - 1}| < \infty$ holds. What's more, 

\begin{tiny}
\begin{eqnarray}
||\tilde{y}_i^T\cdot f^{\psi_{[i]}-1}(x_i;\theta)||_2 &\leq& ||\tilde{y}_i^T ||_2\cdot ||f^{\psi_{[i]}-1}(x_i;\theta)||_2 \\
&\leq& ||\tilde{y}_i^T ||_2\cdot ||\mathbf{1}||_2 \\
&<& \infty
\end{eqnarray}
\end{tiny}

Thus, $||\ell_{\psi_{[i]}}(f(x_i;\theta), \tilde{y}_i)||_2 < \infty$.

We obtain the gradient of $\mathbf{\theta}$ with respect to $\mathcal{L}_T(\theta, \psi | D_T^u)$ (abbreviated as $\mathcal{L}_{T}$) as below:
\begin{tiny}
\begin{eqnarray}
\bigtriangledown_{\mathbf{\theta}}\mathcal{L}_{T} &=& -\frac{1}{n}\sum_{i=0}^{|D_u|} \bigtriangledown_{\mathbf{\theta}}f(x_i;{\mathbf{\theta}})^T\cdot (\tilde{y}_i \odot f^{\mathbf{\psi}^t_{[i]} -2}(x_i;{\mathbf{\theta}}))
\end{eqnarray}
\end{tiny}

Then, we derive the gradient of $\mathbf{\psi}_{[i]}$ with respect to $\bigtriangledown_{\mathbf{\psi}_{[i]}}\mathcal{L}_{T}$ and obtain Eq.~\eqref{eq:t1_1}. When we further derive the gradient of $\mathbf{\psi}_{[i]}$ with respect to $\frac{\partial^2 \mathcal{L}_T}{\partial\mathbf{\theta}\partial\mathbf{\psi}_{[i]}}$, we can obtain Eq.~\eqref{eq:t1_2}.

\begin{tiny}
\begin{eqnarray}
\frac{\partial^2 \mathcal{L}_T}{\partial\mathbf{\theta}\partial\mathbf{\psi}_{[i]}} &=& \bigtriangledown_{\mathbf{\theta}}f(x_i;{\mathbf{\theta}})^T\cdot[\tilde{y}_i\odot log(f(x_i;{\mathbf{\theta}})) \odot f^{\mathbf{\psi}_{[i]} -2 }(x_i;{\mathbf{\theta}})] \nonumber \\ \label{eq:t1_1} \\
\frac{\partial^3 \mathcal{L}_T}{\partial\mathbf{\theta}\partial^2\mathbf{\psi}_{[i]}} &=& \bigtriangledown_{\mathbf{\theta}}f(x_i;{\mathbf{\theta}})^T\cdot[\tilde{y}_i\odot log^2(f(x_i;{\mathbf{\theta}})) \odot f^{\mathbf{\psi}_{[i]} -2 }(x_i;{\mathbf{\theta}})] \nonumber\\ \label{eq:t1_2}
\end{eqnarray}
\end{tiny}

According to the Assumption~\ref{assmp:model_Assmp}, $\bigtriangledown_{\mathbf{\theta}}f(x_i;{\mathbf{\theta}})$ is bounded by $\mathfrak{f}$. Then, Eq.~\eqref{ineq:t1_3} and Eq.~\eqref{eq:t1_4} hold.
\begin{tiny}
\begin{eqnarray}
||\frac{\partial^2 \mathcal{L}_T}{\partial\mathbf{\theta}\partial\mathbf{\psi}_{[i]}}||_2 &=& ||\bigtriangledown_{\mathbf{\theta}}f(x_i;{\mathbf{\theta}})^T\cdot[\tilde{y}_i\odot log(f(x_i;{\mathbf{\theta}})) \odot f^{\mathbf{\psi}_{[i]} -2 }(x_i;{\mathbf{\theta}})]||_2 \nonumber \\
& \leq & ||\bigtriangledown_{\mathbf{\theta}}f(x_i;{\mathbf{\theta}})||_2||[\tilde{y}_i\odot log(f(x_i;{\mathbf{\theta}})) \odot f^{\mathbf{\psi}_{[i]} -2 }(x_i;{\mathbf{\theta}})]||_2 \nonumber \\
& \leq & \mathfrak{f}||[\tilde{y}_i\odot log(f(x_i;{\mathbf{\theta}})) \odot f^{\mathbf{\psi}_{[i]} -2 }(x_i;{\mathbf{\theta}})]||_2 \nonumber \\\label{ineq:t1_3} \\
||\frac{\partial^3 \mathcal{L}_T}{\partial\mathbf{\theta}\partial\mathbf{\psi}_{[i]}^2}||_2 &=& ||\bigtriangledown_{\mathbf{\theta}}f(x_i;{\mathbf{\theta}})^T\cdot [\tilde{y}_i\odot log^2(f(x_i;{\mathbf{\theta}})) \odot f^{\mathbf{\psi}_{[i]} -2 }(x_i;{\mathbf{\theta}})]||_2\nonumber\\
&\leq& ||\bigtriangledown_{\mathbf{\theta}}f(x_i;{\mathbf{\theta}})||_2||_2 [\tilde{y}_i\odot log^2(f(x_i;{\mathbf{\theta}})) \odot f^{\mathbf{\psi}_{[i]} -2 }(x_i;{\mathbf{\theta}})]||_2\nonumber\\
&\leq& \mathfrak{f}||_2 [\tilde{y}_i\odot log^2(f(x_i;{\mathbf{\theta}})) \odot f^{\mathbf{\psi}_{[i]} -2 }(x_i;{\mathbf{\theta}})]||_2\nonumber\\\label{eq:t1_4}
\end{eqnarray}
\end{tiny}

As $\tilde{y}_i$ is a one-hot vector, we only need to analyze the non-zero element in $\tilde{y}_i\odot log^2(f(x_i;{\mathbf{\theta}})) \odot f^{\mathbf{\psi}_{[i]} -2 }(x_i;{\mathbf{\theta}})$. Let the index of the non-zero element be $j$, then Eq.~\eqref{ineq:t1_5} holds.

\begin{tiny}
    \begin{eqnarray}
        &&||_2 [\tilde{y}_i\odot log^2(f(x_i;{\mathbf{\theta}})) \odot f^{\mathbf{\psi}_{[i]} -2 }(x_i;{\mathbf{\theta}})]||_2\nonumber\\
        &=&||[\tilde{y}_{i[j]}\times log^2(f_{[j]}(x_i;{\mathbf{\theta}}) \times f_{[j]}^{\mathbf{\psi}_{[i]} -2 }(x_i;{\mathbf{\theta}})]||_2\nonumber\\
        &=&|[log^2(f_{[j]}(x_i;{\mathbf{\theta}})) \times f_{[j]}^{\mathbf{\psi}_{[i]} -2 }(x_i;{\mathbf{\theta}})]|\label{ineq:t1_5}
    \end{eqnarray}
\end{tiny}

According to Lemma~\ref{lemma:collary_func}, Eq.~\eqref{ineq:t1_5} is bounded, thus \begin{tiny}$||\frac{\partial^3 \mathcal{L}_T}{\partial\mathbf{\theta}\partial\mathbf{\psi}_{[i]}^2}||$\end{tiny} is a bounded term. Similar analysis can be conducted on Eq.~\eqref{ineq:t1_3}, and the result is that \begin{tiny}$||\frac{\partial^2 \mathcal{L}_T}{\partial\mathbf{\theta}\partial\mathbf{\psi}_{[i]}}||$\end{tiny} is also bounded. Since entropy index is instance adaptive, $\psi_{[i]}$ is independent to $\psi_{[j\neq i]}$. Thus, \begin{tiny}$||\frac{\partial^2 \mathcal{L}_T}{\partial\mathbf{\theta}\partial\mathbf{\psi}}||$\end{tiny} and \begin{tiny}$||\frac{\partial^3 \mathcal{L}_T}{\partial\mathbf{\theta}\partial\mathbf{\psi}^2}||$\end{tiny} are bounded
\end{proof}

\renewcommand\thelemma{\arabic{lemma}}
\setcounter{lemma}{0}
\begin{lemma}
\label{lemma:1}
 Suppose the operations in the base model is Lipschitz smooth, then $\ell_{\mathbf{\psi}_{[i]}}(f(x_i, \mathbf{\theta}), \tilde{y}_i)$ is Lipschitz smooth with respect to $\mathbf{\theta}$ for $\forall \psi_{[i]} > 1$ and $\forall x_{i} \in D_{S}\bigcup D^u_{T}$, i.e., there exists a finite constant $\rho_{1}$ and a finite constant $L_{1}$ that satisfy: 

\begin{tiny}
\begin{eqnarray}
||\frac{\partial \ell_{\psi_{[i]}}(f(x_i, \mathbf{\theta}), \tilde{y}_i)}{\partial \mathbf{\theta}}||_2 \leq \rho_1, \quad ||\frac{\partial^2 l_{\psi_{[i]}}(f(x_i, \mathbf{\theta}), \tilde{y}_i)}{\partial \mathbf{\theta}^2}||_2 &\leq& L_1   \label{eq:smooth}
\end{eqnarray}
\end{tiny}

Also, for $\forall \psi_{[i]} > 1$ and $\forall x_{i} \in D^u_{T}$, $\ell_{\mathbf{\psi}_{[i]}}(f(x_i, \mathbf{\theta}), \tilde{y}_i)$ is Lipschitz smooth with respect to $\psi_{[i]}$, i.e., there exists a finite constant $\rho_{2}$ and a finite constant $L_{2}$ that satisfy:

\begin{tiny}
\begin{eqnarray}
||\frac{\partial \ell_{\psi_{[i]}}(f(x_i, \mathbf{\theta}), \tilde{y}_i)}{\partial \mathbf{\psi}_{[i]}}||_2 \leq \rho_2, \quad
||\frac{\partial^2 \ell_{\psi_{[i]}}(f(x_i, \mathbf{\theta}), \tilde{y}_i)}{\partial \mathbf{\psi}_{[i]}^2}||_2 \leq L_2 \label{eq:psi_smooth}
\end{eqnarray}
\end{tiny}
\end{lemma}
\begin{proof}
With the definition of Tsallis loss in Eq.~(3), we have:

\begin{tiny}
\begin{eqnarray}
||\frac{\partial l_{\psi}(f(x_i, \mathbf{\theta}), \tilde{y}_i)}{\partial \mathbf{\theta}}||_2 &=& ||\bigtriangledown_{\mathbf{\theta}}f(x_i;{\mathbf{\theta}})^T\cdot (\tilde{y}_i \odot f^{\mathbf{\psi}^t_{[i]} -2}(x_i;{\mathbf{\theta}})))||_2 \\
&\leq& ||\bigtriangledown_{\mathbf{\theta}}f(x_i;{\mathbf{\theta}})||_2\cdot ||(\tilde{y}_i \odot f^{\mathbf{\psi}^t_{[i]} -2}(x_i;{\mathbf{\theta}})))||_2 \nonumber\\
&& \\
&\leq& \mathfrak{f}\cdot ||(\tilde{y}_i \odot f^{\mathbf{\psi}^t_{[i]} -2}(x_i;{\mathbf{\theta}})))||_2 \\
&=& \mathfrak{f}\cdot ||f_{[z]}^{\mathbf{\psi}^t_{[i]} -2}(x_i;{\mathbf{\theta}})))||_2
\end{eqnarray}
\end{tiny}

\noindent where $z$ is the index of the non-zero element in one-hot $\tilde{y}_i$. According to Assumption~\ref{assmp:model_Assmp}, $f_{[z]}(x_i;{\mathbf{\theta}}))$ is constrained in $(0, 1]$, $f_{[z]}^{\mathbf{\psi}^t_{[i]} -2}(x_i;{\mathbf{\theta}}))$ is thus bounded. Therefore, there exists a finit constant $\rho_{1}$ such that \begin{tiny}$||\frac{\partial l_{\psi_{[i]}}(f(x_i, \mathbf{\theta}), \tilde{y}_i)}{\partial \mathbf{\theta}}||_2 < \rho_{1}$\end{tiny}.

Furthermore, we have:

\begin{tiny}
\begin{eqnarray}
&&||\frac{\partial^2 l_{\psi_{[i]}}(f(x_i, \mathbf{\theta}), \tilde{y}_i)}{\partial \mathbf{\theta}^2}||_2 \nonumber \\
&=& ||\bigtriangledown^2_{\mathbf{\theta}}f(x_i;{\mathbf{\theta}})^T\cdot (\tilde{y}_i \odot f^{\mathbf{\psi}^t_{[i]} -2}(x_i;{\mathbf{\theta}}))) \nonumber \\
&& + \bigtriangledown_{\mathbf{\theta}}f(x_i;{\mathbf{\theta}})^T\cdot (\tilde{y}_i \odot f^{\mathbf{\psi}^t_{[i]} -3}(x_i;{\mathbf{\theta}})))\cdot \bigtriangledown_{\mathbf{\theta}}f(x_i;{\mathbf{\theta}})^T||_2 \\
&\leq& ||\bigtriangledown^2_{\mathbf{\theta}}f(x_i;{\mathbf{\theta}})^T\cdot (\tilde{y}_i \odot f^{\mathbf{\psi}^t_{[i]} -2}(x_i;{\mathbf{\theta}}))) ||_2\nonumber \\
&& + ||\bigtriangledown_{\mathbf{\theta}}f(x_i;{\mathbf{\theta}})^T\cdot (\tilde{y}_i \odot f^{\mathbf{\psi}^t_{[i]} -3}(x_i;{\mathbf{\theta}})))\cdot \bigtriangledown_{\mathbf{\theta}}f(x_i;{\mathbf{\theta}})^T||_2 \\
&\leq& \mathfrak{b}\cdot ||(\tilde{y}_i \odot f^{\mathbf{\psi}^t_{[i]} -2}(x_i;{\mathbf{\theta}}))) ||_2\nonumber \\
&& + \mathfrak{f}^2\cdot||(\tilde{y}_i \odot f^{\mathbf{\psi}^t_{[i]} -3}(x_i;{\mathbf{\theta}})))||_2 
\end{eqnarray}
\end{tiny}

Since $f(x_i;{\mathbf{\theta}}))$ is bounded, $f^{\mathbf{\psi}^t_{[i]} -2}(x_i;{\mathbf{\theta}}))$ is bounded. Thus, \begin{tiny}$||\frac{\partial^2 l_{\psi}(f(x_i, \mathbf{\theta}), \tilde{y}_i)}{\partial \mathbf{\theta}^2}||$\end{tiny} is also bounded, which means that there exists a finite constant $L_{1}$ that satisfies \begin{tiny}$||\frac{\partial^2 l_{\psi}(f(x_i, \mathbf{\theta}), \tilde{y}_i)}{\partial \mathbf{\theta}^2}||_2 < L_{1}$\end{tiny}

We write the gradient \begin{tiny}$\nabla_{\psi_{[i]}}\ell_{\psi_{[i]}}(f(x_i;\theta), \tilde{y}_i)$\end{tiny} as below:
\begin{tiny}
\begin{eqnarray}
\frac{\partial l_{\psi_{[i]}}(f(x_i, \mathbf{\theta}), \tilde{y}_i)}{\partial \mathbf{\psi}_{[i]}} &=& \frac{1}{\psi_{[i]}-1} [\ell_{1}(f(x_i;\mathbf{\theta}), \tilde{y}_i)-\ell_{\psi_{[i]}}(f(x_i;\mathbf{\theta}), \tilde{y}_i)] \nonumber \\
&&- \ell_{1}(f(x_i;\mathbf{\theta}), \tilde{y}_i)\times \ell_{\psi_{[i]}}(f(x_i;\mathbf{\theta}), \tilde{y}_i) \label{eq:A_1_1}
\end{eqnarray}
\end{tiny}

Since $\psi_{[i]} > 1$, $||\frac{1}{\psi_{[i]}-1}||<\infty$ holds. According to Lemma~\ref{lemma:bound_func}, $\ell_{1}(f(x_i;\mathbf{\theta}), \tilde{y}_i)$ and $\ell_{\psi_{[i]}}(f(x_i;\mathbf{\theta}), \tilde{y}_i)$ are bounded. Thus, 

\begin{tiny}
\begin{eqnarray}
&&||\nabla_{\psi_{[i]}}\ell_{\psi_{[i]}}(f(x_i;\theta), \tilde{y}_i)||_2 \nonumber \\
&=& ||\frac{1}{\psi_{[i]}-1} [\ell_{1}(f(x_i;\mathbf{\theta}), \tilde{y}_i)-\ell_{\psi_{[i]}}(f(x_i;\mathbf{\theta}), \tilde{y}_i)] \nonumber \\
&&- \ell_{1}(f(x_i;\mathbf{\theta}), \tilde{y}_i)\times \ell_{\psi_{[i]}}(f(x_i;\mathbf{\theta}), \tilde{y}_i)||_2 \\
&\leq& ||\frac{1}{\psi_{[i]}-1} [\ell_{1}(f(x_i;\mathbf{\theta}), \tilde{y}_i)-\ell_{\psi_{[i]}}(f(x_i;\mathbf{\theta}), \tilde{y}_i)]||_2 \nonumber \\
&& + ||\ell_{1}(f(x_i;\mathbf{\theta}), \tilde{y}_i)\times \ell_{\psi_{[i]}}(f(x_i;\mathbf{\theta}), \tilde{y}_i)||_2 \\
&\leq& ||\frac{1}{\psi_{[i]}-1} ||_2\times||[\ell_{1}(f(x_i;\mathbf{\theta}), \tilde{y}_i)-\ell_{\psi_{[i]}}(f(x_i;\mathbf{\theta}), \tilde{y}_i)]||_2 \nonumber \\
&& + ||\ell_{1}(f(x_i;\mathbf{\theta}), \tilde{y}_i)\times \ell_{\psi_{[i]}}(f(x_i;\mathbf{\theta}), \tilde{y}_i)||_2 \\
&<& \infty \label{eq:psi_grad_bound}
\end{eqnarray}
\end{tiny}
which implies that there exists a finit constant $\rho_{2}$ that satisfies  \begin{tiny}$||\nabla_{\psi_{[i]}}\ell_{\psi_{[i]}}(f(x_i;\theta), \tilde{y}_i)||\leq \rho_{2}$\end{tiny}

With similar efforts, we write \begin{tiny}$\nabla^2_{\psi_{[i]}}\ell_{\psi_{[i]}}(f(x_i;\theta), \tilde{y}_i)$\end{tiny} as:

\begin{tiny}
\begin{eqnarray}
&&\frac{\partial^2 l_{\psi_{[i]}}(f(x_i, \mathbf{\theta}), \tilde{y}_i)}{\partial \mathbf{\psi}^2_{[i]}} \nonumber \\
&=& (\nabla_{\psi_{[i]}}\frac{1}{\psi_{[i]}-1})\cdot[\ell_{1}(f(x_i;\mathbf{\theta}), \tilde{y}_i)-\ell_{\psi_{[i]}}(f(x_i;\mathbf{\theta}), \tilde{y}_i)] \nonumber \\
&& - (\frac{1}{\psi_{[i]}-1}) \cdot [\nabla_{\psi_{[i]}}\ell_{\psi_{[i]}}(f(x_i;\mathbf{\theta}), \tilde{y}_i)] \nonumber \\
&&- \ell_{1}(f(x_i;\mathbf{\theta}), \tilde{y}_i)\times \nabla_{\psi_{[i]}}\ell_{\psi_{[i]}}(f(x_i;\mathbf{\theta}), \tilde{y}_i) \label{eq:A_1_1} \\
&=& (\frac{1}{\psi_{[i]}-1})^2\cdot[\ell_{\psi_{[i]}}(f(x_i;\mathbf{\theta}), \tilde{y}_i)-\ell_{1}(f(x_i;\mathbf{\theta}), \tilde{y}_i)] \nonumber \\
&& - (\frac{1}{\psi_{[i]}-1}) \cdot [\nabla_{\psi_{[i]}}\ell_{\psi_{[i]}}(f(x_i;\mathbf{\theta}), \tilde{y}_i)] \nonumber \\
&&- \ell_{1}(f(x_i;\mathbf{\theta}), \tilde{y}_i)\times \nabla_{\psi_{[i]}}\ell_{\psi_{[i]}}(f(x_i;\mathbf{\theta}), \tilde{y}_i) \label{eq:A_1_1}
\end{eqnarray}
\end{tiny}

As illustrated above, $|\frac{1}{\psi_{[i]}-1}|<\infty$, and $\ell_{1}(f(x_i;\mathbf{\theta}), \tilde{y}_i)$, $\ell_{\psi_{[i]}}(f(x_i;\mathbf{\theta}), \tilde{y}_i)$ are bounded terms. From Eq.~\eqref{eq:psi_grad_bound}, \begin{tiny}$||\nabla_{\psi_{[i]}}\ell_{\psi_{[i]}}(f(x_i;\theta), \tilde{y}_i)||\leq \infty$\end{tiny} holds. 
With a similar analysis in Eq.~\eqref{eq:psi_grad_bound}, we conclude:

\begin{tiny}
\begin{eqnarray}
&&||\frac{\partial^2 l_{\psi_{[i]}}(f(x_i, \mathbf{\theta}), \tilde{y}_i)}{\partial \mathbf{\psi}^2_{[i]}}||_2 \nonumber \\
&\leq& ||\frac{1}{\psi_{[i]}-1}||_2^2\times||\ell_{\psi_{[i]}}(f(x_i;\mathbf{\theta}), \tilde{y}_i)-\ell_{1}(f(x_i;\mathbf{\theta}), \tilde{y}_i)||_2 \nonumber \\
&& + ||\frac{1}{\psi_{[i]}-1}||_2 \times ||\nabla_{\psi_{[i]}}\ell_{\psi_{[i]}}(f(x_i;\mathbf{\theta}), \tilde{y}_i)||_2 \nonumber \\
&& + ||\ell_{1}(f(x_i;\mathbf{\theta}), \tilde{y}_i) ||_2\times ||\nabla_{\psi_{[i]}}\ell_{\psi_{[i]}}(f(x_i;\mathbf{\theta}), \tilde{y}_i)||_2 \\
&<& \infty
\end{eqnarray}
\end{tiny}

Thus, there exists a finit constant $L_{2}$ that satisfies \begin{tiny}$||\nabla^2_{\psi_{[i]}}\ell_{\psi_{[i]}}(f(x_i;\theta), \tilde{y}_i)||\leq L_{2}$\end{tiny}.

\end{proof}

\renewcommand\thelemma{\Alph{lemma}}
\setcounter{lemma}{2}
\begin{lemma}
The entropy indexes $\psi$ is Lipschitz continuous with constant $\rho_v$, and Lipschitz smooth with constant $L_v$ to the loss $\mathcal{L}_S(\hat{\theta}(\psi)| D_S)$. Formally, 
\begin{tiny}
\begin{eqnarray}
||\frac{\partial \mathcal{L}_S(\hat{\theta}(\psi)|D_S)}{\partial \mathbf{\psi}}||_2 \leq \rho_v, \quad
||\frac{\partial^2 \mathcal{L}_S(\hat{\theta}(\psi)|D_S)}{\partial \mathbf{\psi}^2}||_2 \leq L_v \label{eq:psi_smooth}
\end{eqnarray}
\end{tiny}
\end{lemma}

\begin{proof}
The gradients of $\mathbf{\psi}$ with respect to meta loss are written as:

\begin{tiny}
\begin{eqnarray}
\frac{\partial \mathcal{L}_S}{\partial \mathbf{\psi}} &=& \frac{\partial \mathcal{L}_S}{\partial \hat{\mathbf{\theta}}(\psi)}\cdot \frac{\partial \hat{\mathbf{\theta}}(\psi)}{\partial \mathbf{\psi}} \nonumber \\
 &=& (\frac{\partial\mathcal{L}_S}{\partial\hat{\theta}})^T\cdot (\frac{\partial^2 \mathcal{L}_T}{\partial\theta \partial\mathbf{\psi}} ) \label{eq:partial_alpha}
\end{eqnarray}
\end{tiny}

According to Lemma~\ref{lemma:bound_func}, $\frac{\partial^2 \mathcal{L}_T}{\partial\mathbf{\theta}\partial\mathbf{\psi}_{[i]}}$ and $\frac{\partial^3 \mathcal{L}_T}{\partial\mathbf{\theta}\partial\mathbf{\psi}^{2}_i}$ are bounded. Here, we let $\frac{\partial^2 \mathcal{L}_T}{\partial\mathbf{\psi}\partial\mathbf{\psi}_{[i]}}$ be bounded by $\varrho$ and $\frac{\partial^2 \mathcal{L}_T}{\partial\mathbf{\psi}\partial^2\mathbf{\psi}_{[i]}}$ be bounded by $\mathfrak{B}$. Thus, we have the following inequality:

\begin{tiny}
\begin{eqnarray}
||\frac{\partial \mathcal{L}_S}{\partial \mathbf{\psi}}||_2 &=& ||(\frac{\partial\mathcal{L}_S}{\partial\hat{\theta}})^T\cdot (\frac{\partial^2 \mathcal{L}_T}{\partial\theta \partial\mathbf{\psi}}) ||_2 \label{eq:partial_alpha} \\
&\leq& ||(\frac{\partial\mathcal{L}_S}{\partial\hat{\theta}})||_2\times||(\frac{\partial^2 \mathcal{L}_T}{\partial\theta \partial\mathbf{\psi}}) ||_2 \label{eq:partial_alpha} \\
&=& \rho_1 \times \varrho
\end{eqnarray}
\end{tiny}

As $\rho_{1}$ and $\varrho$ are finite constants, we know that there exists a finit constant $\rho_{v}$ that satisfies $||\frac{\partial \mathcal{L}_S(\hat{\theta}(\psi)|D_S)}{\partial \mathbf{\psi}}||_2 \leq \rho_v$.

Further, we observe that:

\begin{tiny}
\begin{eqnarray}
\frac{\partial^2 \mathcal{L}_S}{\partial \mathbf{\psi}^2} &=& [\nabla_{\psi}(\frac{\partial\mathcal{L}_S}{\partial\hat{\theta}})^T] \cdot (\frac{\partial^2 \mathcal{L}_T}{\partial\theta \partial\mathbf{\psi}} ) + (\frac{\partial\mathcal{L}_S}{\partial\hat{\theta}})^T \cdot [\nabla_{\psi}(\frac{\partial^2 \mathcal{L}_T}{\partial\theta \partial\mathbf{\psi}} )] \label{eq:alpha_hessian_1} \\
&=& [\nabla_{\psi}(\frac{\partial\mathcal{L}_S}{\partial\hat{\theta}})^T] \cdot (\frac{\partial^2 \mathcal{L}_T}{\partial\theta \partial\mathbf{\psi}} ) + (\frac{\partial\mathcal{L}_S}{\partial\hat{\theta}})^T \cdot [(\frac{\partial^2 \mathcal{L}_T}{\partial\theta \partial\mathbf{\psi}^2} )] \label{eq:alpha_hessian_2} \\
&=& [\frac{\partial^2 \mathcal{L}_T}{\partial\theta \partial\mathbf{\psi}}]^T\cdot (\frac{\partial^2\mathcal{L}_S}{\partial\hat{\theta}^2})\cdot (\frac{\partial^2 \mathcal{L}_T}{\partial\theta \partial\mathbf{\psi}} ) + (\frac{\partial\mathcal{L}_S}{\partial\hat{\theta}})^T \cdot [(\frac{\partial^2 \mathcal{L}_T}{\partial\theta \partial\mathbf{\psi}^2} )] \nonumber\\
&& \label{eq:alpha_hessian_3} \\
&=& (\frac{\partial^2\mathcal{L}_S}{\partial\hat{\theta}^2})^T \cdot (\frac{\partial^2 \mathcal{L}_T}{\partial\theta \partial\mathbf{\psi}})^2 + (\frac{\partial\mathcal{L}_S}{\partial\hat{\theta}})^T \cdot [(\frac{\partial^2 \mathcal{L}_T}{\partial\theta \partial\mathbf{\psi}^2} )] \label{eq:alpha_hessian_3} 
\end{eqnarray}
\end{tiny}

Thus, we have:

\begin{tiny}
\begin{eqnarray}
||\frac{\partial^2 \mathcal{L}_S}{\partial \mathbf{\psi}^2}||_2 &=& ||(\frac{\partial^2\mathcal{L}_S}{\partial\hat{\theta}^2})^T \cdot (\frac{\partial^2 \mathcal{L}_T}{\partial\theta \partial\mathbf{\psi}})^2 + (\frac{\partial\mathcal{L}_S}{\partial\hat{\theta}})^T \cdot (\frac{\partial^2 \mathcal{L}_T}{\partial\theta \partial\mathbf{\psi}^2} )||_2 \label{eq:alpha_hessian_3} \\
&\leq& ||(\frac{\partial^2\mathcal{L}_S}{\partial\hat{\theta}^2})^T \cdot (\frac{\partial^2 \mathcal{L}_T}{\partial\theta \partial\mathbf{\psi}})^2||_2 + ||(\frac{\partial\mathcal{L}_S}{\partial\hat{\theta}})^T \cdot (\frac{\partial^2 \mathcal{L}_T}{\partial\theta \partial\mathbf{\psi}^2} )||_2 \nonumber \\
&&\label{eq:alpha_hessian_3} \\
&\leq& L_1\cdot\varrho^2 + \rho\cdot\mathfrak{B}
\end{eqnarray}
\end{tiny}

As $L_{1}$, $\rho_{1}$, $\mathcal{B}$ and $\varrho$ are finite constants, there exists a finit constant $L_{v}$ that satisfies $||\frac{\partial^2 \mathcal{L}_S(\hat{\theta}(\psi)|D_S)}{\partial \mathbf{\psi}^2}||_2 \leq L_v$.

\end{proof}

\setcounter{theorem}{0}
\begin{theorem}
\label{theo:1}
The training process in MTEM can achieve \begin{tiny} $\mathbb{E}[\|\nabla_{\psi} \mathcal{L}_{S}(\hat{\mathbf{\theta}}_{t}(\mathbf{\psi}_{t})|D_S)\|_2^2] \leq \epsilon $\end{tiny} in \begin{tiny}$\mathcal{O}(\frac{1}{\epsilon^3})$\end{tiny} steps:

\begin{tiny}
\begin{eqnarray}
   \min _{0 \leq t \leq T} \mathbb{E}[\|\nabla_{\mathbf{\psi}} \mathcal{L}_{S}(\hat{\mathbf{\theta}}_{t}(\mathbf{\psi}_{t})|D_S)\|_2^2] \leq \mathcal{O}(\frac{C}{\sqrt[3]{T}}) ,
\end{eqnarray}
\end{tiny}

\noindent where $C$ is an independent constant.
\end{theorem}
\begin{proof}
The update of $\mathbf{\psi}$ in each iteration is as follows:
\end{proof}
\begin{tiny}
\begin{eqnarray}
\mathbf{\psi}_{t+1}=\mathbf{\psi}_{t}- \beta \frac{1}{m} \sum_{i=1}^m \nabla_{\mathbf{\psi}} \mathcal{L}_S(\hat{\mathbf{\theta}}_{t}(\mathbf{\psi})| D_S)|_{\mathbf{\psi}_{t}} \label{proof_theo1:1}
\end{eqnarray}
\end{tiny}

In our implementation, we sample validation batch $\mathcal{V}$ from $D_{S}$ and replace Eq.~\eqref{proof_theo1:1} with Eq.~\eqref{proof_theo1:2}, as shown in Algorithm~1. 
\begin{tiny}
\begin{eqnarray}
\mathbf{\psi}_{t+1}=\mathbf{\psi}_{t}-\beta_t \nabla_{\psi} \mathcal{L}_{S}(\hat{\mathbf{\theta}}_{t}(\mathbf{\psi}_{t}) | \mathcal{V} ) \label{proof_theo1:2}
\end{eqnarray}
\end{tiny}

In the following proof, we abbreviate $\mathcal{L}_{S}(\hat{\mathbf{\theta}}_{t}(\mathbf{\psi}_{t}) | D_S)$ as $\mathcal{L}_{S}(\hat{\mathbf{\theta}}_{t}(\mathbf{\psi}_{t}))$, and abbreviate $\mathcal{L}_{S}(\hat{\mathbf{\theta}}_{t}(\mathbf{\psi}_{t}) | \mathcal{V})$ as $\tilde{\mathcal{L}}_{S}(\hat{\mathbf{\theta}}_{t}(\mathbf{\psi}_{t}))$. Since the validation batch $\mathcal{V}$ is uniformly from the entire data set $D_{S}$, we rewrite the update as:
\begin{tiny}
\begin{eqnarray}
\mathbf{\psi}_{t+1}=\mathbf{\psi}_{t}-\beta_t[\nabla_{\psi} \mathcal{L}_{S}(\hat{\mathbf{\theta}}_{t}(\mathbf{\psi}_{t}))+\xi_{t}]
\end{eqnarray}
\end{tiny}

\noindent where $\xi_{t} = \nabla_{\psi} \tilde{\mathcal{L}}_S(\hat{\mathbf{\theta}}_{t}(\mathbf{\psi}_{t})) - \nabla_{\psi} \mathcal{L}_{S}(\hat{\mathbf{\theta}}_{t}(\mathbf{\psi}_{t}))$ are \textit{i.i.d} random
variable with finite variance $\sigma_{S}$. Furthermore, $\mathbb{E}[\xi_{t}] =0$, as $\mathcal{V}$ are drawn uniformly at random. 

\noindent Observe that
\begin{tiny}
\begin{eqnarray}
&&\mathcal{L}_S(\hat{\mathbf{\theta}}_{t+1}(\mathbf{\psi}_{t+1}))-\mathcal{L}_S(\hat{\mathbf{\theta}}_{t}(\mathbf{\psi}_{t})) \label{eq:meta_decrease_1}\\
&=&\{\mathcal{L}_S(\hat{\mathbf{\theta}}_{t+1}(\mathbf{\psi}_{t+1}))-\mathcal{L}_S(\hat{\mathbf{\theta}}_{t}(\mathbf{\psi}_{t+1}))\} \nonumber \\
&& +\{\mathcal{L}_{S}(\hat{\mathbf{\theta}}_{t}(\mathbf{\psi}_{t+1}))-\mathcal{L}_{S}(\hat{\mathbf{\theta}}_{t}(\mathbf{\psi}_{t}))\}
\end{eqnarray}
\end{tiny}
By Lipschitz smoothness of $\theta$ to $\mathcal{L}_{S}(\theta|D_S)$, we have

\begin{tiny}
\begin{eqnarray}
&&\mathcal{L}_S(\hat{\mathbf{\theta}}_{t+1}(\mathbf{\psi}_{t+1}))-\mathcal{L}_S(\hat{\mathbf{\theta}}_{t}(\mathbf{\psi}_{t+1})) \nonumber\\
&\leq&\langle\nabla_{\theta} \mathcal{L}_S(\hat{\mathbf{\theta}}_{t}(\mathbf{\psi}_{t+1})), \hat{\mathbf{\theta}}_{t+1}(\mathbf{\psi}_{t+1})-\hat{\mathbf{\theta}}_{t}(\mathbf{\psi}_{t+1})\rangle \nonumber \\
&& +\frac{L_1}{2}\|\hat{\mathbf{\theta}}_{t+1}(\mathbf{\psi}_{t+1})-\hat{\mathbf{\theta}}_{t}(\mathbf{\psi}_{t+1})\|_2^2 \label{ineq:term_1_1}\\
&=&\langle\nabla_{\theta}\mathcal{L}_S(\hat{\mathbf{\theta}}_{t}(\mathbf{\psi}_{t+1})), -\eta_t\cdot \nabla_{\theta}\mathcal{L}_T(\mathbf{\theta}, \mathbf{\psi}^{t+1}| \mathcal{B})\rangle \nonumber \\
&& +\frac{L_1}{2}\|-\eta_t\cdot \nabla_{\theta}\mathcal{L}_T(\mathbf{\theta}, \mathbf{\psi}^{t+1}| \mathcal{B})\|_2^2 \label{ineq:term_1_2}\\
&\leq& -\eta_t \rho_1^2+\frac{L_1}{2}\cdot\eta_t^2\cdot\rho_1^2 \label{ineq:LS_dec_1_3}
\end{eqnarray}
\end{tiny}

\noindent Eq.~\eqref{ineq:term_1_2} is obtained according to line 8 in Algorithm~1, Eq.~\eqref{ineq:LS_dec_1_3} is due to Lemma~\ref{lemma:1}. 

Due to the Lipschitz continuity of $\mathcal{L}_S(\hat{\mathbf{\theta}}_{t}(\mathbf{\psi}))$ (mentioned in Lemma~\ref{lemma:1}), we can obtain the following:

\begin{tiny}
\begin{eqnarray}
&&\mathcal{L}_S(\hat{\mathbf{\theta}}_{t}(\mathbf{\psi}_{t+1}))-\mathcal{L}_S(\hat{\mathbf{\theta}}_{t}(\mathbf{\psi}_{t})) \nonumber \\
&\leq& \langle\nabla_{\psi} \mathcal{L}_S(\hat{\mathbf{\theta}}_{t}(\mathbf{\psi}_{t})), \mathbf{\psi}_{t+1}-\mathbf{\psi}_{t}\rangle +\frac{L_v}{2}\|\mathbf{\psi}_{t+1}-\mathbf{\psi}_{t}\|_2^2 \label{ineq:term_2_3}\\
&=&\langle\nabla_{\psi} \mathcal{L}_S(\hat{\mathbf{\theta}}_{t}(\mathbf{\psi}_{t})),-\beta_t[\nabla_{\psi} \mathcal{L}_S(\hat{\mathbf{\theta}}_{t}(\mathbf{\psi}_{t}))+\xi_{t}]\rangle \nonumber \\
&& +\frac{L_v \beta_t^2}{2}\|\nabla_{\psi} \mathcal{L}_S(\hat{\mathbf{\theta}}_{t}(\mathbf{\psi}_{t}))+\xi_{t}\|_2^2 \label{ineq:term_1_3}\\
&=&-(\beta_t-\frac{L_v \beta_t^2}{2})\|\nabla_{\psi} \mathcal{L}_S(\hat{\mathbf{\theta}}_{t}(\mathbf{\psi}_{t}))\|_2^2 \nonumber \\
&& +\frac{L_v \beta_t^2}{2}\|\xi_{t}\|_2^2-(\beta_t-L_v \beta_t^2)\langle\nabla_{\psi} \mathcal{L}_S(\hat{\mathbf{\theta}}_{t}(\mathbf{\psi}_{t})), \xi_{t}\rangle \label{ineq:term_1_3}
\end{eqnarray}
\end{tiny}

Thus, Eq.~\eqref{eq:meta_decrease_1} satisfies:

\begin{tiny}
\begin{eqnarray}
&&\mathcal{L}_S(\hat{\mathbf{\theta}}_{t+1}(\mathbf{\psi}_{t+1}))-\mathcal{L}_S(\hat{\mathbf{\theta}}_{t}(\mathbf{\psi}_{t})) \nonumber \\
&\leq& \eta_t \rho_1^2(\frac{\eta_t L_1}{2} - 1) \nonumber\\
&&-(\beta_t-\frac{L_v \beta_t^2}{2})\|\nabla \mathcal{L}_S(\hat{\mathbf{\theta}}_{t}(\mathbf{\psi}_{t}))\|_2^2+\frac{L_v \beta_t^2}{2}\|\xi_{t}\|_2^2 \nonumber\\
&&-(\beta_t-L_v \beta_t^2)\langle\nabla \mathcal{L}_S(\hat{\mathbf{\theta}}_{t}(\mathbf{\psi}_{t})), \xi_{t}\rangle
\end{eqnarray}
\end{tiny}

Rearranging the terms, we can obtain:

\begin{tiny}
\begin{eqnarray}
&&(\beta_t-\frac{L_v \beta_t^2}{2})\|\nabla \mathcal{L}_S(\hat{\mathbf{\theta}}_{t}(\mathbf{\psi}_{t}))\|_2^2 \nonumber\\
&\leq& \eta_t \rho_1^2(\frac{\eta_t L_1}{2} - 1)+\mathcal{L}_S(\hat{\mathbf{\theta}}_{t}(\mathbf{\psi}_{t}))-\mathcal{L}_S(\hat{\mathbf{\theta}}_{t+1}(\mathbf{\psi}_{t+1})) \nonumber \\
&&+\frac{L_v \beta_t^2}{2}\|\xi_{t}\|_2^2-(\beta_t-L_v \beta_t^2)\langle\nabla \mathcal{L}_S(\hat{\mathbf{\theta}}_{t}(\mathbf{\psi}_{t})), \xi_{t}\rangle
\end{eqnarray}
\end{tiny}

Summing up the above inequalities and rearranging the terms, we can obtain:

\begin{tiny}
\begin{eqnarray}
&&\sum_{t=1}^T(\beta_t-\frac{L_v \beta_t^2}{2})\|\nabla \mathcal{L}_S(\hat{\mathbf{\theta}}_{t}(\mathbf{\psi}_{t}))\|_2^2 \nonumber \\
&\leq& \mathcal{L}_S(\hat{\mathbf{\theta}}_{1})(\mathbf{\psi}_{1})-\mathcal{L}_S(\hat{\mathbf{\theta}}_{T+1}(\mathbf{\psi}_{T+1})) + \sum_{t=1}^T \eta_t \rho_1^2(\frac{\eta_t L_1}{2} -1) \nonumber \\
&&-\sum_{t=1}^T(\beta_t-L_v \beta_t^2)\langle\nabla \mathcal{L}_S(\hat{\mathbf{\theta}}_{t}(\mathbf{\psi}_{t})), \xi_{t}\rangle + \frac{L_v}{2} \sum_{t=1}^T \beta_t^2\|\xi_{t}\|_2^2 \label{ineq:theo1_1}\\
&\leq& \mathcal{L}_S(\hat{\mathbf{\theta}}_{1}(\mathbf{\psi}_{1}))+\sum_{t=1}^T \mathbf{\psi}_t \rho_1^2(\frac{\mathbf{\psi}_t L_1}{2} - 1) \nonumber \\
&&-\sum_{t=1}^T(\beta_t-L_v \beta_t^2)(\nabla \mathcal{L}_S(\hat{\mathbf{\theta}}_{t}(\mathbf{\psi}_{t})), \xi_{t}\rangle+\frac{L_v}{2} \sum_{t=1}^T \beta_t^2\|\xi_{t}\|_2^2 \label{ineq:theo1_random_2}
\end{eqnarray}
\end{tiny}

Since $\xi_{t}$ are \textit{i.i.d} random
variable with finite variance $\sigma_{S}$ and $\mathbb{E}[\xi_{t}] =0$, we have $\mathbb{E}_{\xi_{t}}\langle\nabla \mathcal{L}_S({\mathbf{\psi}}_{t}),\xi_{t}\rangle = 0$ $\mathbb{E}[\|\xi_{t}\|_2^2] \leq \sigma^2$.
Thus, by taking expectations with respect to $\xi_{t}$  on both sides, we obtain:

\begin{tiny}
\begin{eqnarray}
&&\min _t \mathbb{E}[\|\nabla \mathcal{L}_S(\hat{\mathbf{\theta}}_{t}(\mathbf{\psi}_{t}))\|_2^2] \nonumber \\
&\leq& \frac{\sum_{t=1}^T(\beta_t-\frac{L_v \beta_t^2}{2}) \mathbb{E}_{\xi_{t}}\|\nabla \mathcal{L}_S(\hat{\mathbf{\theta}}_{t}(\mathbf{\psi}_{t}))\|_2^2}{\sum_{t=1}^T(\beta_t-\frac{L_v \beta_t^2}{2})}\\
&\leq& \frac{2 \mathcal{L}_S(\hat{\mathbf{\theta}}_{1}(\mathbf{\psi}_{1}))+\sum_{t=1}^T \eta_t \rho_1^2(\eta_t L_1 - 2)+L_v \sigma_S^2 \sum_{t=1}^T \beta_t^2}{\sum_{t=1}^T(2 \beta_t-L_v \beta_t^2)}\\
&\leq& \frac{2 \mathcal{L}_S(\hat{\mathbf{\theta}}_{1}(\mathbf{\psi}_{1}))+\sum_{t=1}^T \eta_t \rho_1^2(\eta_t L_1 - 2)}{\sum_{t=1}^T(2 \beta_t-L_v \beta_t^2)} +L_v \sigma_S^2\\
&\leq& \frac{2 \mathcal{L}_S(\hat{\mathbf{\theta}}_{1}(\mathbf{\psi}_{1}))+\sum_{t=1}^T \eta_t \rho_1^2(\eta_t L_1 - 2)}{\sum_{t=1}^T(\beta_t)} +L_v \sigma_S^2\\
&\leq& \frac{1}{T \beta_t}[2 \mathcal{L}_S(\hat{\mathbf{\theta}}_{1}(\mathbf{\psi}_{1}))+\eta_1 \rho_1^2 T(L_1 - 2)] + L_v \sigma_S^2\\
&=&\frac{2 \mathcal{L}_S(\hat{\mathbf{\theta}}_{1}(\mathbf{\psi}_{1}))}{T} \frac{1}{\beta_t}+\frac{2 \eta_1 \rho_1^2(L_1 - 2)}{\beta_t}+ L_v \sigma_S^2\\
&\leq& \frac{2 \mathcal{L}_S(\hat{\mathbf{\theta}}_{1}(\mathbf{\psi}_{1}))}{T} \frac{1}{\beta_t}+\frac{2 \eta_1 \rho_1^2(L_1 - 2)}{\beta_t}+L_v \sigma_S^2\\
&=&\frac{\mathcal{L}_S(\hat{\mathbf{\theta}}_{1}(\mathbf{\psi}_{1}))}{T} \max \{L, \frac{ \sqrt[3]{T^2}}{k_2}\} \nonumber\\
&&+\min \{1, \frac{k_1}{T}\} \max \{L, \frac{\sqrt[3]{T^2}}{k_2}\} \rho_1^2(L_1 - 2) \\
&&+L_v \sigma_S^2 \\
&\leq& \frac{\mathcal{L}_S(\hat{\mathbf{\theta}}_{1}(\mathbf{\psi}_{1})}{k_2 \sqrt[3]{T}}+\frac{k_1 \rho_1^2(L_1 - 2)}{k_2 \sqrt[3]{T}}+L \sigma_S^2  \\
&=&\mathcal{O}(\frac{1}{\sqrt[3]{T}})
\end{eqnarray}
\end{tiny}

The third inequlity holds for \begin{tiny}$\sum_{t=1}^T(2 \beta_t-L \beta_t^2) \geq {\sum_{t=1}^T\beta_t^2}$\end{tiny}. The fourth inequlity holds for \begin{tiny}$\sum_{t=1}^T(2 \beta_t-L \beta_t^2) \geq \sum_{t=1}^T\beta_t$\end{tiny}. Therefore, we can achieve \begin{tiny}$\min\limits_{0 \leq t \leq T} \mathbb{E}[\|\nabla \mathcal{L}_S(\mathbf{\psi}_{t})\|_2^2] \leq \mathcal{O}(\frac{1}{\sqrt[3]{T}})$ \end{tiny} in T steps, and this finishes our proof of Theorem 1.

\subsection{Proof for Theorem 2}
\label{appendix:proof_2}

\begin{lemma}
\label{lemma:ab_convergence}
(Lemma 2 in~\cite{shu2019meta}) Let $(a_n)_n \leq 1$ , $(b_n)_n\leq1$  be two non-negative real sequences such that the series $\sum_{t=1}^{\infty} a_n$ diverges, the series $\sum_{t=1}^{\infty} a_n b_n$ converges, and there exists $\nu > 0$ such that $|b_{n+1}-b_n| \leq \nu a_n$. Then the sequences $(b_n)_n \leq 1$  converges to 0.
\end{lemma}

\begin{theorem}
\label{theo:2}
With the training process in MTEM, the instance adaptive Tsallis entropy is guaranteed to be converged on unlabeled data. Formally, 
\begin{tiny}
\begin{eqnarray}
\lim _{t \rightarrow \infty} \mathbb{E}[\|\nabla_{\mathbf{\theta}} \mathcal{L}_T(\mathbf{\theta}_{t}, \mathbf{\psi}_{t+1}| D_T^u)\|_2^2]=0
\end{eqnarray}
\end{tiny}
\end{theorem}

\begin{proof}
With the assumption for the learning rate $\eta_{t}$ and $\beta_{t}$, we can conclude that $\eta_{t}$ satisfies $\sum_{t=0}^{\infty} \eta_t = \infty$ and $\sum_{t=0}^{\infty} \eta_t^2 < \infty$,  $\beta_{t}$ satisfies $\sum_{t=0}^{\infty} \beta_t = \infty$ and
$\sum_{t=0}^{\infty} \beta_t^2 < \infty$. We abbreviate $\mathcal{L}_{T}(\theta, \psi|D_T^u)$ as $\mathcal{L}_{T}(\theta, \psi)$, $\mathcal{L}_{T}(\theta, \psi|\mathcal{B})$ as $\tilde{\mathcal{L}}_{T}(\theta, \psi)$, where $\mathcal{B}$ is a training batch sampled uniformly from $D_{T}^{u}$, as shown in Algorithm~1. Then, each update step is written below:
\begin{tiny}
\begin{eqnarray}
\mathbf{\theta}_{t+1}&=&\mathbf{\theta}_{t}-\eta_t\nabla_{\mathbf{\theta}} \tilde{\mathcal{L}}_T(\mathbf{\theta}_t, \psi_{t+1}) \\
&=&\mathbf{\theta}_{t}-\eta_t[\nabla_{\mathbf{\theta}} \mathcal{L}_T(\mathbf{\theta}_t, \psi_{t+1}) + \Upsilon_t]
\end{eqnarray}
\end{tiny}
where $\Upsilon_t = \nabla_{\theta} \tilde{\mathcal{L}}_T(\mathbf{\theta}_t, \psi_{t+1}) - \nabla_{\theta} \mathcal{L}_{T}(\mathbf{\theta}_t, \psi_{t+1})$ are \textit{i.i.d} random
variable with finite variance at most $\sigma_{T}$. Furthermore, $\mathbb{E}[\Upsilon_{t}] =0$, as $\mathcal{B}$ are drawn uniformly at random. 

Observe that
\begin{tiny}
\begin{eqnarray}
&&\mathcal{L}_T(\mathbf{\theta}_{t+1} ; \mathbf{\psi}_{t+2})-\mathcal{L}_T(\mathbf{\theta}_{t} ; \mathbf{\psi}_{t+1}) \nonumber\\
&=&\{\mathcal{L}_T(\mathbf{\theta}_{t+1} ; \mathbf{\psi}_{t+2})-\mathcal{L}_T(\mathbf{\theta}_{t+1} ; \mathbf{\psi}_{t+1})\} \nonumber \\
&& + \lbrace \mathcal{L}_{T}(\mathbf{\theta}_{t+1} ; \mathbf{\psi}_{t+1})-\mathcal{L}_T(\mathbf{\theta}_{t} ; \mathbf{\psi}_{t+1}) \rbrace \nonumber \\
\end{eqnarray}
\end{tiny}
For the first term,
\begin{tiny}
\begin{eqnarray}
& &\mathcal{L}_T(\mathbf{\theta}_{t+1} ; \mathbf{\psi}_{t+2})-\mathcal{L}_T(\mathbf{\theta}_{t+1} ; \mathbf{\psi}_{t+1}) \\
&\leq & \langle \nabla_{\psi} \mathcal{L}_{T}(\mathbf{\theta}_{t+1} ; \mathbf{\psi}_{t+1}), \mathbf{\psi}_{t+1}-\mathbf{\psi}_{t}\rangle+\frac{L_2}{2}\|\mathbf{\psi}_{t+2}-\mathbf{\psi}_{t+1}\|_2^2\} \\
&=& \langle \nabla_{\psi} \mathcal{L}_{T}(\mathbf{\theta}_{t+1} ; \mathbf{\psi}_{t+1}) ,-\beta_t[\nabla_{\psi}\mathcal{L}_S(\hat{\mathbf{\theta}}_{t}(\mathbf{\psi}_{t}))+\xi_{t}]\rangle \nonumber \\
&& + \frac{L_2 \beta_t^2}{2}\|\nabla_{\psi}\mathcal{L}_S(\hat{\mathbf{\theta}}_{t}(\mathbf{\psi}_{t}))+\xi_{t}\|_2^2\} \\
&=& \langle \nabla_{\psi} \mathcal{L}_{T}(\mathbf{\theta}_{t+1} ; \mathbf{\psi}_{t+1}) ,-\beta_t[\nabla_{\psi}\mathcal{L}_S(\hat{\mathbf{\theta}}_{t}(\mathbf{\psi}_{t}))+\xi_{t}]\rangle \nonumber\\
&& + \frac{L_2 \beta_t^2}{2} \{\|\nabla_{\psi}\mathcal{L}_S(\hat{\mathbf{\theta}}_{t}(\mathbf{\psi}_{t}))||_2^2+2\langle \nabla_{\psi}\mathcal{L}_S(\hat{\mathbf{\theta}}_{t}(\mathbf{\psi}_{t})), \xi_{t}\rangle + ||\xi_{t}\|_2^2\}  \nonumber \\
\end{eqnarray}
\end{tiny}
For the second term,
\begin{tiny}
\begin{eqnarray}
&&\mathcal{L}_{T}(\mathbf{\theta}_{t+1} ; \mathbf{\psi}_{t+1})-\mathcal{L}_{T}(\mathbf{\theta}_{t} ; \mathbf{\psi}_{t+1}) \nonumber\\
&\leq&\langle\nabla_{\theta}\mathcal{L}_{T}(\mathbf{\theta}_{t} ; \mathbf{\psi}_{t+1}), \mathbf{\theta}_{t+1}-\mathbf{\theta}_{t}\rangle+\frac{L_1}{2}\|\mathbf{\theta}_{t+1}-\mathbf{\theta}_{t}\|_2^2 \nonumber\\
&=&\langle\nabla_{\theta}\mathcal{L}_{T}(\mathbf{\theta}_{t} ; \mathbf{\psi}_{t+1}),-\eta_t[\nabla_{\theta}\mathcal{L}_{T}(\mathbf{\theta}_{t} ; \mathbf{\psi}_{t+1})+\Upsilon_{t}]\rangle \nonumber\\
&&+\frac{L_1 \eta_t^2}{2}\|\nabla_{\theta}\mathcal{L}_T(\mathbf{\theta}_{t} ; \mathbf{\psi}_{t+1})+\Upsilon_{t}\|_2^2 \nonumber\\
&=&-(\eta_t-\frac{L_1 \eta_t^2}{2})\|\nabla_{\theta}\mathcal{L}_{T}(\mathbf{\theta}_{t} ; \mathbf{\psi}_{t+1})\|_2^2+\frac{L_1 \eta_t^2}{2}\|\Upsilon_{t}\|_2^2 \nonumber\\
&& -(\eta_t-L_1 \eta_t^2)\langle\nabla_{\theta}\mathcal{L}_{T}(\mathbf{\theta}_{t} ; \mathbf{\psi}_{t+1}), \Upsilon_{t}\rangle \nonumber \\
\end{eqnarray}
\end{tiny}
Therefore, we have:
\begin{tiny}
\begin{eqnarray}
&&\mathcal{L}_T(\mathbf{\theta}_{t+1} ; \mathbf{\psi}_{t+2})-\mathcal{L}_{T}(\mathbf{\theta}_{t} ; \mathbf{\psi}_{t+1}) \nonumber\\
&\leq& \langle \nabla_{\psi} \mathcal{L}_{T}(\mathbf{\theta}_{t+1} ; \mathbf{\psi}_{t+1}) ,-\beta_t[\nabla_{\psi}\mathcal{L}_S(\hat{\mathbf{\theta}}_{t}(\mathbf{\psi}_{t}))+\xi_{t}]\rangle \nonumber\\
&& + \frac{L_2 \beta_t^2}{2} \{\|\nabla_{\psi}\mathcal{L}_S(\hat{\mathbf{\theta}}_{t}(\mathbf{\psi}_{t}))||_2^2+2\langle \nabla_{\psi}\mathcal{L}_S(\hat{\mathbf{\theta}}_{t}(\mathbf{\psi}_{t})), \xi_{t}\rangle + ||\xi_{t}\|_2^2\}  \nonumber \\
&& -(\eta_t-\frac{L_1 \eta_t^2}{2})\|\nabla_{\theta}\mathcal{L}_{T}(\mathbf{\theta}_{t} ; \mathbf{\psi}_{t+1})\|_2^2+\frac{L_1 \eta_t^2}{2}\|\Upsilon_{t}\|_2^2 \nonumber\\
&& -(\eta_t-L_1 \eta_t^2)\langle\nabla_{\theta}\mathcal{L}_{T}(\mathbf{\theta}_{t} ; \mathbf{\psi}_{t+1}), \Upsilon_{t}\rangle \label{ineq:LT_dec} \\
\end{eqnarray}
\end{tiny}

Taking expectation of both sides of Eq.~\eqref{ineq:LT_dec} and since $\mathbb{E}[\xi_{t}]=0$,$\mathbb{E}[\Upsilon_{t}]=0$, we have:

\begin{tiny}
\begin{eqnarray}
&&\mathbb{E}[\mathcal{L}_T(\mathbf{\theta}_{t+1} ; \mathbf{\psi}_{t+2})]-\mathbb{E}[\mathcal{L}_T(\mathbf{\theta}_{t} ; \mathbf{\psi}_{t+1})] \nonumber \\
&\leq& \langle \nabla_{\psi} \mathcal{L}_{T}(\mathbf{\theta}_{t+1} ; \mathbf{\psi}_{t+1}) ,-\beta_t\cdot \nabla \mathcal{L}_S(\hat{\mathbf{\theta}}_{t}(\mathbf{\psi}_{t}))\rangle \nonumber\\
&& + \frac{L_2 \beta_t^2}{2} \{\|\nabla \mathcal{L}_S(\hat{\mathbf{\theta}}_{t}(\mathbf{\psi}_{t}))||_2^2 + \mathbb{E}[||\xi_{t}\|_2^2]\}  \nonumber\\
&& -(\eta_t-\frac{L_1 \eta_t^2}{2})\|\nabla \mathcal{L}_{T}(\mathbf{\theta}_{t} ; \mathbf{\psi}_{t+1})\|_2^2+\frac{L_1 \eta_t^2}{2}\mathbb{E}[\|\Upsilon_{t}\|_2^2] 
\end{eqnarray}
\end{tiny}

Summing up the above inequalities over $ t = 1$, ..., $\infty$ in both sides, we obtain:

\begin{tiny}
\begin{eqnarray}
&& \sum\limits_{t=1}^{\infty}\eta_t\|\nabla_\theta \mathcal{L}_{T}(\mathbf{\theta}_{t} ; \mathbf{\psi}_{t+1})\|_2^2 \nonumber\\
&&+\sum\limits_{t=1}^{\infty}\beta_t ||_2 \nabla_{\psi} \mathcal{L}_{T}(\mathbf{\theta}_{t+1} ; \mathbf{\psi}_{t+1})||_2\cdot||\nabla_{\psi} \mathcal{L}_S(\hat{\mathbf{\theta}}_{t}(\mathbf{\psi}_{t})) ||_2 \nonumber \\
&\leq& \sum\limits_{t=1}^{\infty}\frac{L_2 \beta_t^2}{2} \{\|\nabla_{\psi} \mathcal{L}_S(\hat{\mathbf{\theta}}_{t}(\mathbf{\psi}_{t}))||_2^2 + \mathbb{E}[||\xi_{t}\|_2^2]\} \nonumber\\
&& +\sum\limits_{t=1}^{\infty}\frac{L_1 \eta_t^2}{2}\{ \|\nabla_\theta \mathcal{L}_{T}(\mathbf{\theta}_{t} ; \mathbf{\psi}_{t+1})\|_2^2 + \mathbb{E}[\|\Upsilon_{t}\|_2^2]\} \nonumber\\
&& +\mathbb{E}[\mathcal{L}_T(\mathbf{\theta}_{1} ; \mathbf{\psi}_{2})]-\mathbb{E}_{t\rightarrow \infty}[\mathcal{L}_T(\mathbf{\theta}_{t+1} ; \mathbf{\psi}_{t+2})] \nonumber \\ 
&&\label{eq:train_loss_dec_1} \\
&\leq& \sum\limits_{t=1}^{\infty}\frac{L_2 \beta_t^2}{2} \{\rho_v^2 + \sigma_S^2\} +\sum\limits_{t=1}^{\infty}\frac{L_1 \eta_t^2}{2}(\rho_1^2 + \sigma_T^2) \nonumber\\
&& +\mathbb{E}[\mathcal{L}_T(\mathbf{\theta}_{1} ; \mathbf{\psi}_{2})]-\mathbb{E}_{t\rightarrow \infty}[\mathcal{L}_T(\mathbf{\theta}_{t+1} ; \mathbf{\psi}_{t+2})] \label{eq:train_loss_dec_2} \\
&\leq& \frac{L_2 }{2} \{\rho_v^2 + \sigma_S^2\}\sum\limits_{t=1}^{\infty}\beta_t^2 + \frac{L_1 (\sigma_T^2 + \rho_1^2)}{2}\sum\limits_{t=1}^{\infty}\eta_t^2 \nonumber\\
&& +\mathbb{E}[\mathcal{L}_T(\mathbf{\theta}_{1} ; \mathbf{\psi}_{2})]-\mathbb{E}_{t\rightarrow \infty}[\mathcal{L}_T(\mathbf{\theta}_{t+1} ; \mathbf{\psi}_{t+2})] \label{eq:train_loss_dec_3} \\
&<& \infty \label{eq:train_loss_dec_4}
\end{eqnarray}
\end{tiny}

Eq.~\eqref{eq:train_loss_dec_4} is obtained since $\sum_{t=0}^{\infty} \eta_t^2 < \infty$, 
$\sum_{t=0}^{\infty} \beta_t^2 < \infty$, and $\mathcal{L}_T(\mathbf{\theta}_{1} ; \mathbf{\psi}_{2})$ is bounded. Thus we have:

\begin{tiny}
\begin{eqnarray}
&& \sum\limits_{t=1}^{\infty}\eta_t\|\nabla_\theta \mathcal{L}_{T}(\mathbf{\theta}_{t} ; \mathbf{\psi}_{t+1})\|_2^2 \nonumber\\
&&+\sum\limits_{t=1}^{\infty}\beta_t ||_2 \nabla_{\psi} \mathcal{L}_{T}(\mathbf{\theta}_{t+1} ; \mathbf{\psi}_{t+1})||_2\cdot||\nabla_{\psi} \mathcal{L}_S(\hat{\mathbf{\theta}}_{t}(\mathbf{\psi}_{t})) ||_2 < \infty
\end{eqnarray}
\end{tiny}

Since \begin{tiny} $\sum\limits_{t=1}^{\infty}\beta_t < \infty$ \end{tiny}, we have:

\begin{tiny}
\begin{eqnarray}
&&\sum\limits_{t=1}^{\infty}\beta_t ||_2 \nabla_{\psi} \mathcal{L}_{T}(\mathbf{\theta}_{t+1} ; \mathbf{\psi}_{t+1})||_2\cdot||\nabla_{\psi} \mathcal{L}_S(\hat{\mathbf{\theta}}_{t}(\mathbf{\psi}_{t})) ||_2 \nonumber \\
&\leq& \sum\limits_{t=1}^{\infty}\beta_t\cdot\rho_2\cdot\rho_v\leq \rho_2\cdot\rho_v\cdot\sum\limits_{t=1}^{\infty}\beta_t < \infty
\end{eqnarray}
\end{tiny}

\noindent which implies that \begin{tiny}$\sum_{t=1}^{\infty} \eta_t \mathbb{E}[\|\nabla_{\theta}\mathcal{L}_T(\mathbf{\theta}_{t} ; \mathbf{\psi}_{t+1})\|_2^2]<\infty
$\end{tiny}.

With Lemma~\ref{lemma:ab_convergence} and $\sum_{t=0}^{\infty} \eta_t = \infty$, it is easy to deduce that $\lim\limits_{t \rightarrow \infty} \mathbb{E}[\nabla_{\theta}\mathcal{L}_T(\mathbf{\theta}_{t} ; \mathbf{\psi}_{t+1}) \|_2^2]=0$ holds when:
\begin{tiny}
\begin{eqnarray}
|\mathbb{E}[\nabla_{\theta}\mathcal{L}_{T}(\mathbf{\theta}_{t+1} ; \mathbf{\psi}_{t+2}) \|_2^2]-\mathbb{E}[\nabla_{\theta}\mathcal{L}_{T}(\mathbf{\theta}_{t} ; \mathbf{\psi}_{t+1}) \|_2^2]| \leq \nu \eta_t
\end{eqnarray}
\end{tiny}

\noindent for some constant $\nu$. Due to the Cauchy inequality:

\begin{tiny}
\begin{eqnarray}
|(\|a\|+\|b\|)(\|a\|-\|b\|)| = |(a+b)\cdot(a-b)| \leq\|a+b\|\|a-b\|
\end{eqnarray}
\end{tiny}

We then have:

\begin{tiny}
\begin{eqnarray}
&&|\mathbb{E}[\|\nabla_{\theta}\mathcal{L}_T(\mathbf{\theta}_{t+1} ; \mathbf{\psi}_{t+2})\|_2^2]-\mathbb{E}[\|\nabla_{\theta}\mathcal{L}_T(\mathbf{\theta}_{t} ; \mathbf{\psi}_{t+1})\|_2^2]|\\
&=&|\mathbb{E}[(\|\nabla_{\theta}\mathcal{L}_T(\mathbf{\theta}_{t+1} ; \mathbf{\psi}_{t+2})\|_2+\|\nabla_{\theta}\mathcal{L}_T(\mathbf{\theta}_{t} ; \mathbf{\psi}_{t+1})\|_2) \nonumber\\
&&\quad(\|\nabla_{\theta}\mathcal{L}_T(\mathbf{\theta}_{t+1} ; \mathbf{\psi}_{t+2})\|_2-\|\nabla_{\theta}\mathcal{L}_T(\mathbf{\theta}_{t} ; \mathbf{\psi}_{t+1})\|_2)]|\\
&\leq&\mathbb{E}[|(\|\nabla_{\theta}\mathcal{L}_T(\mathbf{\theta}_{t+1} ; \mathbf{\psi}_{t+2})\|_2+\|\nabla_{\theta}\mathcal{L}_T(\mathbf{\theta}_{t} ; \mathbf{\psi}_{t+1})\|_2)| \nonumber\\
&&\quad|(\|\nabla_{\theta}\mathcal{L}_T(\mathbf{\theta}_{t+1} ; \mathbf{\psi}_{t+2})\|_2-\|\nabla_{\theta}\mathcal{L}_T(\mathbf{\theta}_{t} ; \mathbf{\psi}_{t+1})\|_2)|]\\
&\leq& \mathbb{E}[\|\nabla_{\theta}\mathcal{L}_T(\mathbf{\theta}_{t+1} ; \mathbf{\psi}_{t+2})+\nabla_{\theta}\mathcal{L}_T(\mathbf{\theta}_{t} ; \mathbf{\psi}_{t+1})\|_2 \nonumber \\
&& \|\nabla_{\theta}\mathcal{L}_T(\mathbf{\theta}_{t+1} ; \mathbf{\psi}_{t+2})-\nabla_{\theta}\mathcal{L}_{T}(\mathbf{\theta}_{t} ; \mathbf{\psi}_{t+1})\|_2]\\
&\leq& \mathbb{E}[(\|\nabla_{\theta}\mathcal{L}_{T}(\mathbf{\theta}_{t+1} ; \mathbf{\psi}_{t+2})\|_2+\|\nabla_{\theta}\mathcal{L}_{T}(\mathbf{\theta}_{t} ; \mathbf{\psi}_{t+1})\|_2)\nonumber \\
&&\|\nabla_{\theta}\mathcal{L}_T(\mathbf{\theta}_{t+1} ; \mathbf{\psi}_{t+2})-\nabla_{\theta}\mathcal{L}_{T}(\mathbf{\theta}_{t} ; \mathbf{\psi}_{t+1})\|_2]\\
&\leq& \mathbb{E}[2\rho_1\|\nabla_{\theta}\mathcal{L}_T(\mathbf{\theta}_{t+1} ; \mathbf{\psi}_{t+2})-\nabla_{\theta}\mathcal{L}_{T}(\mathbf{\theta}_{t} ; \mathbf{\psi}_{t+1})\|_2]
\end{eqnarray}
\end{tiny}

Observe that:

\begin{tiny}
\begin{eqnarray}
&&\mathbb{E}[\|(\mathbf{\theta}_{t+1}, \mathbf{\psi}_{t+2})-(\mathbf{\theta}_{t}, \mathbf{\psi}_{t+1})\|_2]\\
&\leq&  \eta_t \beta_t \mathbb{E}[\|(\nabla_{\theta}\mathcal{L}_T(\mathbf{\theta}_{t} ; \mathbf{\psi}_{t+1})+\Upsilon_{t}, \nabla_{\theta}\mathcal{L}_S(\mathbf{\psi}_{t+1})+\xi_{t+1})\|_2]\\
&\leq&  \eta_t \beta_t \mathbb{E}[\sqrt{\|\nabla_{\theta}\mathcal{L}_T(\mathbf{\theta}_{t} ; \mathbf{\psi}_{t+1})+\Upsilon_{t}\|_2^2}+\sqrt{\|\nabla_{\theta}\mathcal{L}_S(\mathbf{\psi}_{t+1})+\xi_{t+1}\|_2^2}]\nonumber\\
&& \label{eq:expec_1}\\
&\leq&  \eta_t \beta_t \sqrt{\mathbb{E}[\|\nabla_{\theta}\mathcal{L}_{T}(\mathbf{\theta}_{t} ; \mathbf{\psi}_{t+1})+\psi_{t}\|_2^2]+\mathbb{E}[\|\nabla_{\theta}\mathcal{L}_S(\mathbf{\psi}_{t+1})+\xi_{t+1}\|_2^2]}\nonumber\\
&& \label{eq:expec_2}\\
&\leq& \eta_t \beta_t \sqrt{2 \sigma^2+2 \rho_1^2}\\
&\leq& \beta_1 \eta_t \sqrt{2(\sigma^2+\rho^2)}
\end{eqnarray}
\end{tiny}

Thus, we have:

\begin{tiny}
\begin{eqnarray}
&&|\mathbb{E}[\|\nabla_{\theta}\mathcal{L}_T(\mathbf{\theta}_{t+1} ; \mathbf{\psi}_{t+2})\|_2^2]-\mathbb{E}[\|\nabla_{\theta}\mathcal{L}_T(\mathbf{\theta}_{t} ; \mathbf{\psi}_{t+1})\|_2^2]| \nonumber\\
&\leq& 2 \rho_1\beta_1\eta_t \sqrt{2(\sigma^2+\rho_1^2)} 
\end{eqnarray}
\end{tiny}

According to Lemma~\ref{lemma:ab_convergence}, we can achieve:
\begin{tiny}
\begin{eqnarray}
\lim _{t \rightarrow \infty} \mathbb{E}[\|\nabla \mathcal{L}_T(\mathbf{\theta}_{t} ; \mathbf{\psi}_{t+1})\|_2^2]=0
\end{eqnarray}
\end{tiny}
\end{proof}

\subsection{Proof for Theorem 3}
\label{appendix:proof_3}

We use hypothesis $h:\mathcal{X} \rightarrow \Delta^{K-1}$ to analyze the effectiveness of MTEM in achieving domain adaptation. Formally, $h_{\mathbf{\theta}}(x_i) = \arg\max\limits_{k} f_{[k]}(x_i;\mathbf{\theta})$. We let $R_D(h) = \mathbb{E}_{x \sim D}[\mathbf{1}(h_{\mathbf{\theta}}(x)\neq h_{\mathbf{\theta}}(x'))| \forall x' \in \mathcal{N}(x)]$ denote the model's robustness to the perturbations on dataset $D$, where $\mathcal{N}(x) = \lbrace x'||x-x'| \leq \xi \rbrace $ means the neighbour set of $x$ with a distance smaller than $\xi$. 
Further, we let $\hat{\mathcal{R}}(\mathcal{H}|_D)=\frac{1}{|D|}\mathbb{E}_{\sigma}(\mathop{sup}\limits_{h\in \mathcal{H}}\sum_{i=1}^{|D|}\sigma_i h_{\mathbf{\theta}}(x_i))$ denote the empirical Rademacher complexity~\cite{gnecco2008approximation} of function class $\mathcal{H}$ ($h \in \mathcal{H}$) on dataset $D$, where $\sigma_{i}$ are independent random noise drawn from the Rademacher distribution i.e. $Pr(\sigma_{i}=+1)=Pr(\sigma_{i}=-1)=1/2$. Then, we deduce Theorem~\ref{theo:3}.

\begin{definition}
(q, c)-\textbf{constant expansion} Let $Prob(D)$ denote the distribution of the dataset $D$, $Prob_i(D)$ denote the conditional distribution given label $i$ . For some constant $q, c \in (0, 1)$, if for any set $D \in \mathbb{D}_{S}\cup \mathbb{D}_{T}$ and $\forall i \in [K]$ with $\frac{1}{2}>Prob_i(D) > q$, we have $Prob_i(\mathcal{N}(D) /\ D) > \min\{c, Prob_i(D)\}$
\end{definition}

\begin{theorem}
\label{theo:3}
Suppose $D_{S}$ and $D_{T}^{u}$ satisfy \textbf{$(q, c)-$ constant expansion}~\cite{DBLP:conf/iclr/WeiSCM21} for some constant $q, c \in (0, 1)$. With the probability at least $1 - \delta$ over the drawing of $D_T^{u}$ from $\mathbb{D}_{T}$, the error rates of the model $h_{\mathbf{\theta}}$ ($h\in \mathcal{H}$) on the target domain (i.e., $\epsilon_{\mathbb{D}_T}(h_{\mathbf{\theta}})$) is bounded by:

\begin{tiny}
\begin{eqnarray}
\epsilon_{\mathbb{D}_T}(h_{\mathbf{\theta}}) &\leq& \mathcal{L}_S({\mathbf{\theta}}|D_S) + \mathcal{L}_T(\mathbf{\theta}, \psi|D_T^u) + 2q + 2K\cdot\hat{\mathcal{R}}(\mathcal{H}|_{D_S}) \nonumber\\
&&+ 4K\cdot\hat{\mathcal{R}}(\tilde{\mathcal{H}}\times\mathcal{H}|_{D_T^u}) +\frac{R_{D_S \cup D_T^u}(h)}{\min \lbrace c, q\rbrace} +\zeta \label{eq:theo3}
\end{eqnarray}
\end{tiny}

\noindent where \begin{tiny}$\zeta = \mathcal{O}(\sqrt{\frac{-log(\delta)}{|D_S|}} + \sqrt{\frac{-log(\delta)}{|D_T^u|}})$\end{tiny} is a low-order term. $\tilde{\mathcal{H}}\times\mathcal{H}$ refers to the function class $\lbrace x \rightarrow h(x)_{[h'(x)]}: h, h' \in \mathcal{H} \rbrace$.
\end{theorem}

\begin{proof}
Define a ramp function $\Gamma(o)$ as :

\begin{tiny}
\begin{equation}
\Gamma(o) = \left \{ \begin{aligned}
&1, \quad o\leq 0 \nonumber \\
&1-o, \quad 0 < o \leq 1 \nonumber \\
&0, \quad 1<o 
\end{aligned}
\right.
\end{equation}
\end{tiny}

\noindent where $o$ is a scalar input. Define the margin function as :

\begin{tiny}
\begin{equation}
\mathcal{M}(f(x;\theta), z) = f_{[z]}(x;\theta) - \max\limits_{k\neq z} f_{[k]}(x;\theta)
\end{equation}
\end{tiny}

According to Theorem 2 in~\cite{liu2021cycle}, we have:

\begin{tiny}
\begin{eqnarray}
\epsilon_{\mathbb{D}_T}(h_{\mathbf{\theta}}) &\leq& \frac{1}{|D_S|}\sum\limits_{(x, y) \in D_S} \Gamma(\mathcal{M}(f(x;\theta), z)) \nonumber\\
&& + \frac{1}{|D_T^u|}\sum\limits_{x \in D_T^u} \Gamma(\mathcal{M}(f(x;\theta), \tilde{z}))  \nonumber\\
&&+ 2q + 2K\cdot\hat{\mathcal{R}}(\mathcal{H}|_{D_S}) \nonumber\\
&&+ 4K\cdot\hat{\mathcal{R}}(\tilde{\mathcal{H}}\times\mathcal{H}|_{D_T^u}) +\frac{R_{D_S \cup D_T^u}(h)}{\min \lbrace c, q\rbrace} +\zeta \label{eq:theo3}
\end{eqnarray}
\end{tiny}

\noindent where $z = \arg\max_k \{y\}$ and $\tilde{z} = \arg\max_k \{\tilde{y}\}$ are the index of the non-zero element in one-hot label vectors $y$ and $\tilde{y}$.

Considering that :

\begin{tiny}
\begin{eqnarray}
\Gamma(\mathcal{M}(f(x;\theta), z)) &=& 1 - (f_{[z]}(x;\theta) - \max\limits_{k \neq z}f_{[k]}(x;\theta)) \\
& =& 1 - f_{[z]}(x;\theta) + \max\limits_{k \neq z}f_{[k]}(x;\theta) \\
&\leq& 1 - f_{[z]}(x;\theta) + 1 - f_{[z]}(x;\theta) \\
&=& 2\times(1 - f_{[z]}(x;\theta)) \\
&=& 2\times \ell_2 (f(x;\theta), y) 
\end{eqnarray}
\end{tiny}

\noindent it is natural to conclude that:

\begin{tiny}
\begin{eqnarray}
\frac{1}{|D_S|}\sum\limits_{(x, y) \in D_S} \Gamma(\mathcal{M}(f(x;\theta), z)) &\leq& \frac{1}{|D_S|}\sum\limits_{(x, y) \in D_S} 2\times \ell_1(f(x;\theta), y) \nonumber \\
&=& \mathcal{L}_S(\theta | D_S)
\end{eqnarray}
\end{tiny}

Also, we conclude that:

\begin{tiny}
\begin{eqnarray}
\frac{1}{|D_T^u|}\sum\limits_{x \in D_T^u} \Gamma(\mathcal{M}(f(x;\theta), z)) &\leq& \frac{1}{|D_T^u|}\sum\limits_{i=1}^{|D_T^u|} 2\times \ell_{\psi_{[i]}}(f(x_i;\theta), \tilde{y}_i) \nonumber \\
&=& \mathcal{L}_T(\theta, \psi | D_S)
\end{eqnarray}
\end{tiny}

Thus,

\begin{tiny}
\begin{eqnarray}
\epsilon_{\mathbb{D}_T}(h_{\mathbf{\theta}}) &\leq& \mathcal{L}_S({\mathbf{\theta}}|D_S) + \mathcal{L}_T(\mathbf{\theta}, \psi|D_T^u) + 2q + 2K\cdot\hat{\mathcal{R}}(\mathcal{H}|_{D_S}) \nonumber\\
&&+ 4K\cdot\hat{\mathcal{R}}(\tilde{\mathcal{H}}\times\mathcal{H}|_{D_T^u}) +\frac{R_{D_S \cup D_T^u}(h)}{\min \lbrace c, q\rbrace} +\zeta \label{eq:theo3}
\end{eqnarray}
\end{tiny}
holds.

\end{proof}

\section{More Details about Tsallis Entropy}

\subsection{Variants of Tsallis Entropy and Tsallis Loss}
\label{appendix:tsallis_variants}
By adjusting the entropy index, we can obtain different kind of entropy types. For example, making the entropy index approaching to 1, we can obtain the Gibbs entropy. And $\alpha = 2$ recovers the Gini impurity.

\begin{tiny}
\begin{eqnarray}
\textnormal{\textbf{Gibbs entropy} : }\quad e_{\mathbf{\alpha}\rightarrow 1}(p_i) &=& \frac{\lim_{\mathbf{\alpha} \rightarrow 1}1 - \sum_{j=1}^K p^{\mathbf{\alpha}}_{i[j]}}{\lim_{\mathbf{\alpha} \rightarrow 1}\mathbf{\alpha} - 1} \nonumber\\
&=& \sum_{j=1}^K - p_{i[j]}log(p_{i[j]}) \label{eq:gibbs_entropy} \\
\textnormal{\textbf{Gini impurity} : }\quad e_{\mathbf{\alpha}=2}(p_i) &=& 1 - \sum_{j=1}^K p^{2}_{i[j]} \label{eq:gini_impurity}
\end{eqnarray}
\end{tiny}

Also, adjust entropy index can generate different Tsallis losses. Especially, making the entropy index approach 1.0 recovers the Cross-entropy, as shown below:

\begin{tiny}
\begin{eqnarray}
\textbf{Cross-Entropy : } \quad \ell_{\mathbf{\alpha}\rightarrow 1}(p_i, y_i) &=& \frac{\lim\limits_{\mathbf{\alpha} \rightarrow 1}1 - \sum_{j=1}^K y_{i[j]}\cdot p^{\mathbf{\alpha}-1}_{i[j]}}{\lim\limits_{\mathbf{\alpha} \rightarrow 1}\mathbf{\alpha} - 1} \nonumber \\
&=& \sum_{j=1}^K - y_{i[j]}log(p_{i[j]}) \label{eq:cross_entropy}
\end{eqnarray}
\end{tiny}

\subsection{Deduction of Eq.~(9)}
\label{appeidx:deduc_loss_new}
We denote $f(x_i;\mathbf{\theta})$ as $\mathbf{f}_{i}$, whose $k$-th element is denoted as $\mathbf{f}_{i[k]}$. Also, we denote $\tilde{y}^T_{i[k]}$ as an one-hot vector whose non-zero element is with the index of $k$. Thus, we have:

\begin{tiny}
\begin{eqnarray}
e_{\psi_{[i]}}(f(x_i;\mathbf{\theta}))&=& \frac{1 - \mathbf{1}^T\cdot\mathbf{f}_i^{\psi_{[i]}}}{(\psi_{[i]} -1 )} \label{eq:e_define} \\
&=& \frac{1 - \mathbf{f}_i^T\cdot\mathbf{f}_i^{\psi_{[i]}-1}}{(\psi_{[i]} -1 )} \label{eq:e_define_1} \\
&=& \frac{\sum\limits_{k=1}^{K}\mathbf{f}_{i[k]} - \sum\limits_{k=1}^{K}\mathbf{f}_{i[k]}\times (\tilde{y}_{i[k]}^T\cdot\mathbf{f}_i^{\psi_{[i]}-1})}{(\psi_{[i]} -1 )} \label{eq:e_define_2} \\
&=& \sum\limits_{k=1}^{K}\mathbf{f}_{i[k]}\times \frac{(1 - \tilde{y}_{i[k]}^T\cdot\mathbf{f}_i^{\psi_{[i]}-1})}{(\psi_{[i]} -1)} \label{eq:e_define_2} \\
&=& \sum\limits_{k=1}^{K}\mathbf{f}_{i[k]}\times \ell_{\psi_{[i]}}(f(x_i;\mathbf{\theta}_t), \tilde{y}_i) \label{eq:e_define_2} \\
&=& \mathop{\mathbb{E}}\limits_{\tilde{y}_i \sim f(x_i;\mathbf{\theta}_t)}\ell_{\psi_{[i]}}(f(x_i;\mathbf{\theta}), \tilde{y}_i) \label{eq:e_define_3}
\end{eqnarray}
\end{tiny}

\subsection{Deduction of Eq.~(14)}
\label{appeidx:psi_grad}
We denote $f(x_i;\mathbf{\theta})$ as $\mathbf{f}_{i}$. Considering $\tilde{y}^T_i$ is an one-hot vector, we denote the elements with index $k$ as $\tilde{y}^T_{i[k]}$. Further, we denote the index of the non-zero element as $z$, then $\tilde{y}^T_{i[z]}=1$ and $\tilde{y}^T_{i[k \neq z]} = 0$. With these denotions, we have:
\begin{tiny}
\begin{eqnarray}
&& \frac{\tilde{y}^T_i\cdot \lbrace log(\mathbf{f}_i)\odot [\mathbf{f}_i^{\mathbf{\psi}_{[i]}-1} - \mathbf{1}]\rbrace}{\mathbf{\psi}_{[i]} -1} \nonumber\\
&=& \frac{1}{\mathbf{\psi}_{[i]} -1}\times\sum\limits_{k=1}^{K} \tilde{y}_{i[k]}\times log(\mathbf{f}_{i[k]})\times[\mathbf{f}_{i[k]}^{\mathbf{\psi}_{[i]}-1} - 1] \label{eq:psi_grad_0_2}\\
&=& \frac{1}{\mathbf{\psi}_{[i]} -1}\times \tilde{y}^T_{i[z]}\times log(\mathbf{f}_{i[z]})\times[\mathbf{f}_{i[z]}^{\mathbf{\psi}_{[i]}-1} - 1] \label{eq:psi_grad_0_3}\\
&=& \frac{1}{\mathbf{\psi}_{[i]} -1}\times 1.0 \times log(\mathbf{f}_{i[z]})\times[\mathbf{f}_{i[z]}^{\mathbf{\psi}_{[i]}-1} - 1] \label{eq:psi_grad_0_4}\\
&=& \frac{1}{\mathbf{\psi}_{[i]} -1}\times [1.0 \times log(\mathbf{f}_{i[z]})]\times [1.0 \times \mathbf{f}_{i[z]}^{\mathbf{\psi}_{[i]}-1} - 1] \label{eq:psi_grad_0_5}\\
&=& \frac{1}{\mathbf{\psi}_{[i]} -1}\times [\tilde{y}_{i[z]} \times log(\mathbf{f}_{i[z]})]\times [\tilde{y}_{i[z]}\times\mathbf{f}_{i[z]}^{\mathbf{\psi}_{[i]}-1} - 1] \label{eq:psi_grad_0_6}\\
&=& [\tilde{y}^T_{i} \cdot log(\mathbf{f}_{i})]\times \frac{[\tilde{y}^T_{i}\cdot \mathbf{f}_{i}^{\mathbf{\psi}_{[i]}-1} - 1]}{\mathbf{\psi}_{[i]} -1} \label{eq:psi_grad_0_7}\\
&=& \ell_{1}(f(x_i;\mathbf{\theta}), \tilde{y}_i)\times \ell_{\psi_{[i]}}(f(x_i;\mathbf{\theta}), \tilde{y}_i)   \label{eq:psi_grad_0_8} 
\end{eqnarray}
\end{tiny}

Grounded on Eq.~\eqref{eq:psi_grad_0_8}, we deduce the gradient of $\psi_{[i]}$ with respect to the unsupervised loss $\mathcal{L}_{T}(\mathbf{\theta}, \psi|D_T^u)$ as:

\begin{tiny}
\begin{eqnarray}
&&\bigtriangledown_{\psi_{[i]}}\mathcal{L}_T(\mathbf{\theta}, \psi|D_T^u) \nonumber \\
&=&\bigtriangledown_{\psi_{[i]}}\ell_{\psi_{[i]}}(x_i, \tilde{y}_i) \\
&=& \frac{\tilde{y}_i^T \cdot \mathbf{f}_i^{\mathbf{\psi}_{[i]}-1} - 1}{(\mathbf{\psi}_{[i]} -1)^2} - \frac{\tilde{y}^T_i\cdot[ log(\mathbf{f}_i)\odot \mathbf{f}_i^{\mathbf{\psi}_{[i]}-1}]}{\mathbf{\psi}_{[i]} -1} \label{eq:T_2} \\
&=& \frac{\tilde{y}_i^T \cdot \mathbf{f}_i^{\mathbf{\psi}_{[i]}-1} - 1}{(\mathbf{\psi}_{[i]} -1)^2} - \frac{\tilde{y}^T_i\cdot[ log(\mathbf{f}_i)\odot \mathbf{f}_i^{\mathbf{\psi}_{[i]}-1}]}{\mathbf{\psi}_{[i]} -1} \label{eq:T_2} \\
&=& -\frac{1}{\psi_{[i]}-1} \ell_{\psi_{[i]}}(f(x_i;\mathbf{\theta}), \tilde{y}_i) -  \frac{\tilde{y}^T_i\cdot \lbrace log(\mathbf{f}_i)\odot [\mathbf{f}_i^{\mathbf{\psi}_{[i]}-1} - \mathbf{1} + \mathbf{1}]\rbrace}{\mathbf{\psi}_{[i]} -1} \label{eq:T_2} \\
&=& -\frac{1}{\psi_{[i]}-1} \ell_{\psi_{[i]}}(f(x_i;\mathbf{\theta}), \tilde{y}_i) -  \frac{\tilde{y}^T_i\cdot \lbrace log(\mathbf{f}_i)\odot [\mathbf{f}_i^{\mathbf{\psi}_{[i]}-1} - \mathbf{1}] + log(\mathbf{f}_i)\rbrace}{\mathbf{\psi}_{[i]} -1} \nonumber\\ \label{eq:T_2} \\
&=& -\frac{1}{\psi_{[i]}-1} \ell_{\psi_{[i]}}(f(x_i;\mathbf{\theta}), \tilde{y}_i) -  \frac{\tilde{y}^T_i\cdot \lbrace log(\mathbf{f}_i)\odot [\mathbf{f}_i^{\mathbf{\psi}_{[i]}-1} - \mathbf{1}]\rbrace}{\mathbf{\psi}_{[i]} -1} \nonumber\\
& &- \frac{\tilde{y}^T_i\cdot log(\mathbf{f}_i)}{\mathbf{\psi}_{[i]} -1}  \label{eq:T_2} \\
&=& -\frac{1}{\psi_{[i]}-1} \ell_{\psi_{[i]}}(f(x_i;\mathbf{\theta}), \tilde{y}_i) -  \frac{\tilde{y}^T_i\cdot \lbrace log(\mathbf{f}_i) \odot [\mathbf{f}_i^{\mathbf{\psi}_{[i]}-1} - \mathbf{1}]\rbrace}{\mathbf{\psi}_{[i]} -1} \nonumber\\
& & + \frac{1}{\mathbf{\psi}_{[i]} -1}\ell_1(f(x_i;\mathbf{\theta}))  \label{eq:T_2} \\
&=& \frac{1}{\psi_{[i]}-1} [\ell_{1}(f(x_i;\mathbf{\theta}), \tilde{y}_i)-\ell_{\psi_{[i]}}(f(x_i;\mathbf{\theta}), \tilde{y}_i)] - \nonumber \\
&& \frac{\tilde{y}^T_i\cdot \lbrace log(\mathbf{f}_i)\odot [\mathbf{f}_i^{\mathbf{\psi}_{[i]}-1} - \mathbf{1}]\rbrace}{\mathbf{\psi}_{[i]} -1}   \label{eq:T_2_6} \\
&=& \frac{1}{\psi_{[i]}-1} [\ell_{1}(f(x_i;\mathbf{\theta}), \tilde{y}_i)-\ell_{\psi_{[i]}}(f(x_i;\mathbf{\theta}), \tilde{y}_i)] \nonumber \\
&&- \ell_{1}(f(x_i;\mathbf{\theta}), \tilde{y}_i)\times \ell_{\psi_{[i]}}(f(x_i;\mathbf{\theta}), \tilde{y}_i)   \label{eq:T_2_7} 
\end{eqnarray}
\end{tiny}

\noindent where $\odot$ denotes the element-wise multiplication. Eq.~\eqref{eq:T_2_7} is obtained by substituting Eq.~\eqref{eq:psi_grad_0_8} into Eq.~\eqref{eq:T_2_6}.

\section{Implementation Details}
\label{sec:appendix_C}

For the symbols in Algorithm~1, we set $\eta_t$ and $\beta_{t}$ with respect to Assumption~\ref{assmp:2}. We set $\eta_{t}$ in Algorithm~1 as $5e-5$ for the BERT model, and $5e-3$ for the BiGCN model. In addition, $\mathbf{\psi}$ is initialized with 2.0, and the learning weight to update the entropy indexs, i.e., $\beta_{t}$ in Algorithm.~1, is initialized with 0.1 for both the BERT and the BiGCN model. We conduct all experiments the GeForce RTX 3090 GPU with 24GB memory.



\section{Statistics of the Datasets}
\label{sec:dataset}

TWITTER dataset is provided in the \href{https://figshare.com/ndownloader/articles/6392078/versions/1}{site}\footnote{https://figshare.com/ndownloader/articles/6392078/} under a CC-BY license. Amazon dataset is accessed from \href{https://github.com/ruidan/DAS}{https://github.com/ruidan/DAS}. The statistics of the TWITTER dataset and the Amazon dataset is listed in Table~\ref{tab:statisticTWITTER} and Table~\ref{tab:Amazon_sts}.

\begin{table}[tbh]
\small
\centering
\caption{Statistics of the TWITTER dataset.}
\setlength{\tabcolsep}{2.pt}
\begin{tabular}{l|lll}
\hline
Domain            & Rumours        & Non-Rumours    & Total \\ \hline
Charlie Hebdo\#   & 458 (22\%)     & 1,621 (78\%)   & 2,079 \\
Ferguson\#        & 284 (24.8\%)   & 859 (75.2\%)   & 1,143 \\
Germanwings Crash & 238 (50.7\%)   & 231 (49.3\%)   & 469   \\
Ottawa Shooting   & 470 (52.8\%)   & 420 (47.2\%)   & 890   \\
Sydney Siege      & 522 (42.8\%)   & 699 (57.2\%)   & 1,221 \\ \hline
Total             & 1,921 (34.0\%) & 3,830 (66.0\%) & 5,802 \\ \hline
\end{tabular}
\label{tab:statisticTWITTER}
\end{table}

\begin{table}[htb]
\caption{Statistics of the Amazon dataset}
\small
\centering
\begin{tabular}{l|ll|l}
\hline
Domains     & positive    & negative    & unlabeled \\ \hline
books       & 1000 (50\%) & 1000(50\%)  & 6001      \\
dvd         & 1000 (50\%) & 1000 (50\%) & 34,742    \\
electronics & 1000 (50\%) & 1000 (50\%) & 13,154    \\
kitchen     & 1000 (50\%) & 1000 (50\%) & 16,786    \\ \hline
\end{tabular}
\label{tab:Amazon_sts}
\end{table}

\section{Cases with Different Entropy Indexes}
\label{appedix:case_study}

Cases are drawn from the Amazon dataset. In each sentences, we highlight the sentiment words with red colors, and highlight their label with blue color.

\subsection{sentences with small entropy index ($\psi\approx1.0$)}
\begin{enumerate}[(i)]
\item \textcolor{blue}{[negative]} i bought 3 of these to monitor my invalid mom and they do n[pad]t work well . \textcolor{red}{the manual is not thorough} , but it does tell you after you bought the product that there is a one-second delay unless you buy an ac adapter . i just bought 12 batteries for 3 units and was n[pad]t about to buy an ac adapter to see if that fixed it . it cut off my mom [pad]s voice and i rarely heard her message . the units also have annoying beeps that ca n[pad]t be turned off . no real volume control either . a very overpriced and poorly designed unit . also , the switches to adjust for conference and vox mode are small and need a small object smaller than a pen to change them . the sound is clear if it does n[pad]t cut you off , but the units are far from user-friendly .
 \item \textcolor{blue}{[negative]} i bought this at amazon , \textcolor{red}{but it [pad]s cheaper at} www.cutleryandmore.com \$ 9.95 , so is the \$ 89.00 wusthof santoku 7 [pad][pad] knife ( \$ 79.00 ) , and they have free shipping ! check yahoo shopping be fore amazon ! ! !
 \item \textcolor{blue}{[positive]} i ordered 2 of these fans from seller `` kramnedlog [pad][pad] and received them yesterday . they were the model 1054 with legs and a/c adapters included and the seller upgraded my regular shipping to priority at no extra charge because i ordered more than one . my husband tested them both with batteries and with the adapter and they have very good air output in both modes . we live in florida and these fans will be a lifesaver if we are ever without power . i was very impressed with the seller . i sent him an e-mail through amazon before i ordered as i had some questions . he promptly e-mailed answering my questions and explaining his shipping upgrade . the fans arrived two days after i placed the order , were nicely packed and arrived in perfect condition . they were the exact model as stated with adapters . i would highly recommend this seller ; he was prompt and courteous . do n[pad]t hesitate to contact a seller by e-mail if you have questions or need additional information before you order . amazon will put you in touch with the seller so you [pad]ll know exactly what you [pad]re getting -- no surprises . \textcolor{red}{five stars for a great product and a positive seller experience}
 \item \textcolor{blue}{[negative]} after 6 months of use , the led switch popped out of the coffeemaker [pad]s base , which turned the wired plug into an on/off device . i found that black and decker no longer services these products and \textcolor{red}{the firm that does does n[pad]t respond to e-mail}
 \item \textcolor{blue}{[negative]} \textcolor{red}{due to the very high number of complaints about the problems people were having with heavy duty tasks ( which is what the mixer should have been made for ! ) due to cheap plastic housing holding the heavy duty metal gears} , i contact kitchenaid to find out if this design flaw has been fixed . it has been and here is their response ( note that they are saying that if you receive one of the models with the plastic housing they will replace it under warranty with the new metal housing . it is also important to note that the people who [pad]ve had the problem have almost all stated that kitchenaid had great customer service and replaced their mixers with no ones with no problems . response from kitchenaid : thank you for visiting the kitchenaid website ! i would like to reassure you that our engineering and product teams have addressed the gear box issue . in fact , we have already implemented a change in the manufacturing of the pro 600 stand mixers to a metal gear box . although these will be changed out on additional 5 qt bowl-lift models that include all metal gearing , the time line on that is not known at this time . although a transparent change , this will insure the long-term durability we expect . i would like to emphasize to you that should you purchase a stand mixer ( with all metal gearing ) which has not been assembled with the upgraded metal gear box , and you should have a problem with your mixer ( as a result of the plastic gear box ( failure ) , we will most certainly extend your warranty to resolve the issue properly and effectively . if you have additional questions , feel free to reply back to me . or , you may contact either our kitchenaid customer satisfaction center at 1-800-541-6390 , or you may visit our secure kitchenaid live chat .
 \item \textcolor{blue}{[positive]} \textcolor{red}{this is a great product} , and you can get it , along with any other products on amazon up to \$ 500 free ! participate in this special promotion and get a free \$ 500 amazon gift card at this web site : stuffnocost.com/amazo
 \item \textcolor{blue}{[positive]} \textcolor{red}{this is a great product} , and you can get it , along with any other products on amazon up to \$ 500 free ! participate in a special promotion and get a free \$ 500 amazon gift card at this web site : ilikethis.info/amazo
\item \textcolor{blue}{[positive]} \textcolor{red}{this is a very good product} , and you can get it , along with any other products on amazon up to \$ 500 free ! participate in this special promotion and get a free \$ 500 amazon gift card at this web site : awesomestufffree.com/amazo
\item \textcolor{blue}{[negative]} i bought this product after reading some of these reviews , and i am sorry that i did n[pad]t buy a higher end model \textcolor{red}{because i do n[pad]t like it }. here [pad]s why : pros - it heats up very quickly . in my small room i only used the low setting , usually at about a 3 . -it [pad]s quieter than my other fan , but it [pad]s still loud to someone who likes to sleep with silence . cons - the heater would n[pad]t even start up until i turned the dial to 3 . -if i turned the dial anywhere past 3 it got way too hot in my room . -the auto shutoff on this totally sucked . it only shut itself off when it was exactly on the 3 line . -speaking of lines , i followed another reviewer [pad]s advise and used a sharpie to draw a line where the numbers were because it was so hard to see them on the knob . -while the heater gets hotter the higher up you turn it , the fan mode does n[pad]t get any cooler . in fan mode everything from 3 - 6 was the same speed and really did n[pad]t help cool things off much . i [pad]m very disappointed in this product , as well as the other reviewers because amazon reviews have never let me down before
\item \textcolor{blue}{[negative]} \textcolor{red}{fantastic product , but way overpriced in europe}. the company website lets you order online , but has separate websites for both usa and europe . you can only order from usa website if you live in usa . we europeans are forced to pay highly inflated prices of more than 120 \% markup ( i.e . more than double ) that of the prices on us website , and if you search around to shop in europe you will find prices much the same . this company is controlling market prices and screwing us europeans with huge markups . it can get away with this because it is treating all european countries the same . please do n[pad]t buy this product if being treated fairly matters to you . it [pad]s the only way to get them to listen'
\end{enumerate}

\subsection{sentences with large entropy index ($\psi=5.0$)}
\begin{enumerate}[(i)]
\item \textcolor{blue}{[positive]} these are great for a wide variety of uses and in this great red color they are very cute too ! cute , practical and useful - how good can they get ! they are perfect to use for this recipe : molten chocolate cakes for best results , use a dark baking chocolate with high cocoa butter content ( about 30 percent ) 12 teaspoons plus 5 tablespoons sugar 8 ounces bittersweet ( not unsweetened ) or semisweet chocolate , chopped 3/4 cup ( 1 1/2 sticks ) unsalted butter 3 large eggs 3 large egg yolks 1 tablespoon all purpose flour {1 quart purchased vanilla bean ice cream} generously butter eight 3/4-cup ramekin or custard dish . sprinkle inside of each dish with 1 1/2 teaspoons sugar . stir chocolate and butter in heavy medium saucepan over low heat until smooth . remove from heat . using electric mixer , beat eggs , egg yolks , and remaining 5 tablespoons sugar in large bowl until thick and pale yellow , about 8 minutes . fold 1/3 of warm chocolate mixture into egg mixture , then fold in remaining chocolate . fold in flour . divide batter among ramekins or dishes . ( can be made 1 day ahead . cover with plastic ; chill . bring to room temperature before continuing . ) preheat oven to 425f . place ramekins/dishes on baking sheet . bake cakes uncovered until edges are puffed and slightly cracked but center 1 inch of each moves slightly when dishes are shaken gently , about 13 minutes . \textcolor{red}{top each cake with scoop of vanilla bean ice cream and serve immediately}. makes 8 servings . enjoy'\\
\item \textcolor{blue}{[negative]} my \textcolor{red}{oven has felt neglected for years} . between {my crock pot , microwave oven and my toaster oven , my real oven has become a storage area for junk} . this crockpot is perfect for one person or a family . it makes even the toughest and cheapest cuts of meat tender and moist . i simply toss in the meat , veggies and potatoes and then leave the house . when i come home , supper is ready . i like the two settings - low and high . the high setting really gets the cooking process moving . cleaning is a snap . if i do n[pad]t wash it right away i simply fill it with hot water and let it soak . it cleans right up without scrubbing , even if there [pad]s some burned food on the bottom . last night i tossed in some chicken , sliced onions and a can of mushroom soup . does it get any easier than that ? it took five minutes of preparation to make a tasty meal fit for a king\\
\item \textcolor{blue}{[positive]} i am a regular bundt baker. i have always used just the basic joe-schmo bundt pan ... no biggie . with this \textcolor{red}{nordic ware rose pan} , oh my gosh ! it just fell out with no effort on my part . with my other pan , i had to heavily grease and flour ... and just pray it did n[pad]t stick too much . with the nordic ware , i lightly brushed with oil , and like pure heaven , my bundt came out so perfect ! it cooked so evenly , and the inside so moist ... such a huge difference , especially when it [pad]s with the same recipe that i have always used . i just love it ! last night , i bought 3 more nordic ware pans ... the poinsettia , ( for the holidays ) , the heart , ( valentines ) , and the basket weave with the fruit imprints on top . i know these pans are not cheap , but the quality of these pans , for me , makes all the difference in the world\\
\item \textcolor{blue}{[positive]} i had wanted one of these baking dishes ever since i read about them in carol field [pad]s book the italian baker . my wife and i received one as a wedding present and i have now baked bread in it multiple times . \textcolor{red}{it does indeed make a difference in the texture of the crust} . after years of experimenting i had never been able to achieve that especially crisp crust that you get from a bakery , until now . another plus is that the dough [pad]s final rise inside the covered dish obviates the need for plastic wrap . i lately have purchased the rectangular `` baguette [pad][pad] version of la cloche at an outlet store , and have found it to be a more practical shape for day to day baking . the original bell-shaped la cloche is great , however , for making large , round hearth loaves or a cluster or dinner rolls .\\
\item \textcolor{blue}{[positive]} this le creuset small spatula is \textcolor{red}{so useful -- in cooking , with small pans , and in scraping jars and bowls} . the colors are bright and pretty . you can keep the whole set of different size spatulas in a matching le creuset poterie right next to your cooktop . i like that i can remove the head of the spatula , put it in the top level of my dishwasher , and sanitize it , especially after i have used it to make something with raw eggs . highly recommended\\
\item \textcolor{blue}{[positive]} chicago metallic commericial \textcolor{red}{bakeware is excellent} . these 8-inch round cake pans are very heavyweight and give even baking results . i rub a bit of vegetable shortening on these pans and then line the bottom with a circle of waxed paper . one can dust some flour over the shortening , too . for an 8-inch layer cake , i like to divide the cake batter between three 8-inch round cake pans . make sure they are staggered on the oven racks for the best baking results\\
\item \textcolor{blue}{[negative]} i like and dislike these bowls . what i like about them is the shape and size for certain foods and for the dishwasher . they are actually too small for cereal if you would like to add fruit to your cereal . they are perfect for oatmeal or ice cream \textcolor{red}{but too small for soup or stew} . they are curved perfectly to drain and rinse well in the dishwasher . the depth of color is very attractive on the table\\
\item \textcolor{blue}{[positive]} \textcolor{red}{this bowl is perfect to bake} in the oven and then to serve right from the oven to the table . it is casually pretty and can go with a casual or a dressy setting . we made a mouth watering meatloaf in it and it was very easy to clean . it was a good buy\\
\item \textcolor{blue}{[positive]} two , \textcolor{red}{large turkey wings fit perfectly in the bottom of the 4-quart romertopf} ( `` 113 [pad][pad] stamped on the bottom ) . in the 3 years that i [pad]ve had mine , turkey wings are the only thing i [pad]ve cooked in it although a duck is slated to be cooked in it tonight . as with the wings , this is the right size for anyone wanting to cook a whole duck . turkey wings tend to get a bit dry if roasted in a traditional , covered , roasting pan . in a romertopf , however , the meat is incredibly moist . all i do is season the wings , put them in the bottom of the soaked ( in cold water , both top and bottom ) romertopf , stick it in a cold oven , turn the oven temperature to 425 degrees , then forget about it until the oven timer goes off 50 minutes later . most times the wings are browned as desired . the few times they have not been , i [pad]ve removed the top and returned the uncovered wings to the oven for an additional 10 minutes . although people writing reviews for some other size romertopfs say they stick it in their dishwasher , mine came with instructions to only wash it by hand and never use soap . for removing grease , the use of baking soda is suggested . i have found the easiest way to clean a romertopf is to first remove as much as possible with slightly warm water and a nylon brush , then soak it over night . in the morning the baked-on dripping are much easier to remove . after removing them i do a final cleaning with baking soda and warm water . one thing people should know is that romertopfs , no matter what size , can not be cleaned so they obtain a new , never used look . being made of porous clay , the bottom and inner top get stained and spotted ( from grease )\\
\item \textcolor{blue}{[positive]} these mugs are a beautiful rich red and are large enough to use as an everyday drinking glass . usually mugs in sets are small and useless , but i have found these to \textcolor{red}{be ideal for water , milk , coffee , tea , and hot chocolate}
\end{enumerate}

\end{document}